\documentclass{article} % For LaTeX2e
\usepackage{iclr2027_conference,times}

\usepackage{amsmath,amsfonts,bm}

\def\eqref#1{equation~\ref{#1}}
\def\1{\bm{1}}

\DeclareMathAlphabet{\mathsfit}{\encodingdefault}{\sfdefault}{m}{sl}
\SetMathAlphabet{\mathsfit}{bold}{\encodingdefault}{\sfdefault}{bx}{n}

\usepackage{hyperref}
\usepackage{url}
\usepackage{fontawesome5}
\usepackage{xspace}
\usepackage{cleveref}
\usepackage{graphicx}
\usepackage{booktabs}
\usepackage{multirow}
\usepackage{subcaption}
\usepackage{graphicx}
\usepackage{float}
\usepackage{wrapfig}
\usepackage{needspace}
\usepackage{etoc}
\renewenvironment{abstract}{%
  \vskip.075in\centerline{\large\sc Abstract}\vspace{0.5ex}%
  \begin{list}{}{\setlength{\leftmargin}{2.75em}\setlength{\rightmargin}{2.75em}%
    \setlength{\topsep}{0pt}\setlength{\partopsep}{0pt}\setlength{\parsep}{0pt}}%
  \item\relax
}{\end{list}\vskip 1ex}
\title{Generative Neural Retargeting for Human-to-Robot Dexterous Manipulation}

\newcommand{\authsym}[1]{\textsuperscript{\normalsize$#1$}}
\newcommand{\authorlink}[2]{{\hypersetup{hidelinks}\href{#1}{#2}}}
\author{\parbox{\dimexpr\textwidth-2\tabcolsep\relax}{\vskip 10pt\centering
\textbf{\authorlink{https://gaodechen.github.io/}{Dechen~Gao}$^{1,2}$\authsym{\ast}\authsym{\dagger} \quad \authorlink{https://yy-gx.github.io/}{Yue~Yang}$^{1,3}$\authsym{\ast}\authsym{\dagger} \quad \authorlink{https://babbatem.github.io/}{Ben~Abbatematteo}$^{1}$ \quad \authorlink{https://github.com/nathanrgodwin}{Nathan~Godwin}$^{1}$ \quad \authorlink{https://outsider86.github.io/}{Pengcheng~Wang}$^{1,4}$\authsym{\dagger} \quad \authorlink{https://ahlab.org/people/roger-boldu/}{Roger~Boldu}$^{1}$ \quad \authorlink{https://stevenwman.github.io/}{Steven~Man}$^{1,6}$\authsym{\dagger} \quad \authorlink{https://people.csail.mit.edu/frankzydou/}{Zhiyang~Dou}$^{1,5}$\authsym{\dagger} \quad \authorlink{https://www.linkedin.com/in/chuanqin/}{Chuan~Qin}$^{1}$ \quad \authorlink{https://www.linkedin.com/in/shonakagome/}{Sho~Nakagome}$^{1}$ \quad \authorlink{https://www.ericwhitmire.com/}{Eric~Whitmire}$^{1}$}\\[4pt]
{\normalfont $^{1}$Meta Reality Labs Research \quad $^{2}$UC Davis \quad $^{3}$UNC Chapel Hill}\\
{\normalfont $^{4}$UC Berkeley \quad $^{5}$MIT \quad $^{6}$CMU}\\[2pt]
{\normalfont\small \authsym{\ast}Equal contribution. \quad \authsym{\dagger}Work done at Meta.}\\[2pt]
{\normalfont\small \href{https://yy-gx.github.io/GNR/}{\faGlobe~Project Page} \quad\textbar\quad \faEnvelope~Correspondence: \texttt{dcgao@ucdavis.edu}, \texttt{yygx@cs.unc.edu}}
}}

\newcommand{\new}[1]{#1}
\newcommand{\ABBR}{\texttt{GNR}\xspace}

\iclrfinalcopy
\begin{document}

\maketitle
% arXiv version: remove the ICLR running header and its rule
\lhead{}
\renewcommand{\headrulewidth}{0pt}
% arXiv version: the ICLR style sets \flushbottom, which stretches a short page; let short pages end naturally instead
\raggedbottom
% arXiv version: tighten the gap between the author block and the abstract so Figure 1 stays on page 1
\vspace{-14pt}
\etocdepthtag.toc{mainbody}

\begin{abstract}

Human demonstrations are a scalable data source for learning dexterous manipulation, but the embodiment gap prevents human motion from being executed directly on robots. Inverse kinematics (IK) retargets human motion to robots efficiently but ignores dynamics, often producing infeasible motions. Reinforcement learning (RL) and sampling-based model predictive control (MPC) are commonly employed to yield dynamically feasible motions, but both are sample-inefficient and sensitive to hyperparameters. RL suffers from costly and unstable training and tedious reward engineering; MPC avoids policy optimization, yet retargets each trajectory in isolation, and solving one does not make the next easier. Sampling cost grows rapidly with dataset size and task difficulty. We hypothesize that dynamically feasible trajectories concentrate near a low-dimensional manifold shared across demonstrations, so that retargeting can be reduced to sampling from that manifold, conditioned on human motion, rather than solving a fresh optimization problem for every demonstration. We propose \textbf{Generative Neural Retargeting} (\ABBR), which uses a flow matching model to sample feasible trajectories. \ABBR{} outperforms MPC with only $8.5\%$ of the samples required by MPC, achieving a success rate of $56.10\%$ compared to $28.13\%$ for MPC. \ABBR{} can be used for scalable and efficient retargeting of large-scale, long-horizon, and millimeter-precision human demonstrations: by applying \ABBR{} within a real-to-sim data engine, we produce a dexterous manipulation dataset spanning $223$k demonstrations and $3.3$k object geometries. Using \ABBR{}-generated data, we train visuomotor policies and demonstrate successful zero-shot sim-to-real transfer.
\end{abstract}

\begin{figure}[H]
\centering
\includegraphics[width=\linewidth]{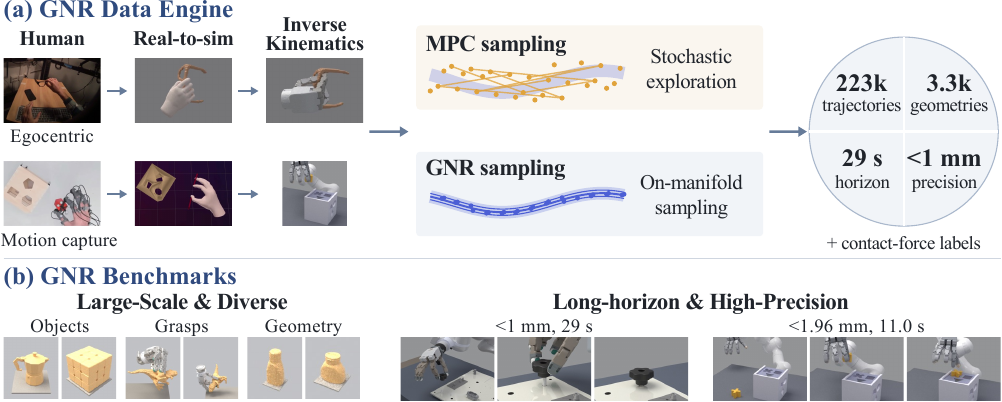}
\caption{We propose \textbf{Generative Neural Retargeting} (\ABBR{}), a scalable and generalizable method that retargets large-scale, long-horizon and high-precision human demonstrations for five-fingered dexterous manipulation. We apply \ABBR{} with a real-to-sim data engine that reconstructs human motion from egocentric videos or a wearable exoskeleton with motion capture to scale up both diverse and difficult high-precision tasks.}
\label{fig:header}
\end{figure}

\section{Introduction}
\label{sec:intro}

Progress in dexterous manipulation has been driven largely by the growing scale of demonstration data~\citep{o2024open, jiang2025dexmimicgen,hoque2026egodex,goel2023humans}. Collecting such data directly on robots, e.g., through teleoperation, is prohibitively expensive: it demands extensive human labor and is constrained by hardware availability and embodiment-specific interfaces~\citep{handa2020dexpilot, qin2023anyteleop}. A promising and scalable alternative is to record human hand manipulation from videos or wearable devices and transfer it to robots~\citep{xu2025dexumi, hot3d, zheng2026egoscale, ma2026humanscale}. However, a fundamental challenge in utilizing such data is that human demonstrations do not provide robot action labels, while the \emph{embodiment gap} prevents human motions from being transferred directly to robots~\citep{spider}. 

Prior work retargets human motions to robots through \emph{kinematic} or \emph{physics-based retargeting}~\citep{spider,dexmachina,dexplore,xu2025dexcanvas,allshire2025visual-videomimic,maniptrans,qiu2026enforcing-amp}. Kinematic methods, e.g., fingertip IK, align human and robot poses geometrically. They are fast but ignore dynamics and contact, so the resulting trajectories are often dynamically infeasible~\citep{handa2020dexpilot,qin2023anyteleop}. \emph{Physics-based retargeting} instead optimizes for dynamically feasible motion through reinforcement learning (RL)~\citep{qiu2026enforcing-amp,maniptrans,allshire2025visual-videomimic,liao2026beyondmimic,dextrack,adadextrack,dexmachina, tang2025deep} or sampling-based optimization methods such as model predictive control (MPC)~\citep{spider,williams2017information-mppi,xue2025full}. The retargeting objective is highly non-convex and discontinuous, which makes each trajectory expensive to solve. RL suffers from unstable and expensive training and tedious reward engineering~\citep{dextrack,chord,dexmachina}. Sampling-based MPC avoids costly policy optimization and directly optimizes the control sequence, but it can be sample-inefficient and sensitive to hyperparameters as every trajectory is solved with highly stochastic sampling and hand-designed heuristics~\citep{spider, paliwal2026dexterous,dexmachina}. The sample inefficiency and engineering burden create a scaling bottleneck for robot learning with human data, which motivates our research question: \emph{how can we generate dynamically feasible robot trajectories from human demonstrations in a scalable and efficient way?}

A key observation is that conventional sampling-based retargeting repeatedly solves a non-convex optimization problem for every trajectory in isolation, and solving one does not make the next cheaper~\citep{dexmachina,dextrack}. However, solved trajectories can provide useful knowledge about feasible motion and contact when retargeting other demonstrations. Thus inspired, we propose \textbf{Generative Neural Retargeting} (\ABBR{}), a conditional flow matching model that learns from a small seed dataset generated by MPC to guide sampling toward dynamically feasible trajectories, conditioned on human motion. Scaling the training dataset can improve model performance~\citep{yang2023diffusion}, but generating the seed dataset can itself be expensive. \ABBR{} complements data scaling with \emph{test-time scaling}: increasing the sampling budget improves retargeting success while reducing reliance on costly seed data. \ABBR{} generalizes to unseen motions, geometries, and objects, and even solves demonstrations on which MPC fails under the sampling budget used to generate \ABBR{}'s training data.

We apply \ABBR{} within a real-to-sim data engine to address the scaling bottleneck of retargeting along two axes, diversity and difficulty, for dexterous manipulation. Along the \textbf{diversity} axis, the goal is a dataset spanning many distinct geometries and motions, where total sampling cost grows linearly with the number of human trajectories, each of which can take hours to optimize~\citep{spider,dexmachina,chord}. Therefore, state-of-the-art methods often demonstrate a few thousand trajectories derived from sources such as egocentric videos~\citep{dextrack,adadextrack,chord,paliwal2026dexterous}, far short of the scale needed for large-scale, diverse dataset generation. Along the \textbf{difficulty} axis, the goal is solving long-horizon and high-precision dexterous manipulation. Errors compound over time, and millimeter-level precision leaves little tolerance for error, so the number of samples needed to discover a feasible trajectory grows rapidly~\citep{ross2011reduction, yang2025boss}. For example, a complex long-horizon industrial assembly task inspired by \texttt{NIST}~\citep{lian2021benchmarking} that requires $1$\,mm precision to screw a knob onto its seat (see \Cref{fig:appendix-nist-demo} for a demonstration) requires 221 CPU core-h to optimize a trajectory averaging 29 seconds. Along the difficulty axis, existing works primarily retarget human demonstrations on tasks with 10 to 150 millimeter precision requirements~\citep{spider,dextrack,dexmachina,chord} due to both the lack of high-precision data sources and the massive sampling required for retargeting. We further discuss existing retargeting datasets in App.~\ref{sec:app:comparing-retargeting-datasets}.
 
As illustrated in~\Cref{fig:header}, the data engine takes human demonstrations from large-scale egocentric video and high-precision motion capture to scale along both axes. A real-to-sim pipeline transfers them into SuperDex~\citep{superdex2026} and applies domain randomization to object pose, physical properties, and geometry. Along the \textbf{diversity} axis, \ABBR{} outperforms state-of-the-art sampling-based methods such as MPC~\citep{xue2025full,pinneri2021sample-icem} with about $8.5\%$ of their sampling budget on \texttt{HOT3D} after learning from a small seed dataset generated by MPC. On out-of-distribution (OOD) objects, motions, and geometries, it reaches $56.10\%$ success versus around $30\%$ for sampling-based methods, and solves $48.01\%$ of demonstrations on which MPC fails. \ABBR{} reduces CPU cost per trajectory by $7.6\times$ relative to MPC. Along the \textbf{difficulty} axis, \ABBR{} solves long-horizon, millimeter-precision tasks open-loop or by supplying MPC with on-manifold initializations, and reduces sampling cost by $3.0\times$--$14.9\times$. We demonstrate that \ABBR{}'s test-time scaling is a strong complement to data scaling, and is particularly useful when \ABBR{} needs to learn in low-data regimes. For example, with only 50 training demonstrations, \ABBR{} success rises from $11.5\%$ to $99.5\%$ as the sample budget grows to 256. Our contributions are:

\textbf{Retargeting as generative modeling with test-time scaling.} We recast physics-based retargeting from per-trajectory sampling-based optimization to generative modeling, \ABBR{}, which allows for efficient and generalizable robot action sampling conditioned on human motion. \ABBR{} demonstrates test-time scaling and generalizes to unseen geometries and motions, making it a scalable and cost-efficient method for retargeting large-scale, long-horizon, and high-precision human demonstrations to feasible robot trajectories.

\textbf{A scalable and generalizable data engine.} We develop a real-to-sim pipeline that extracts and retargets human demonstrations from egocentric videos and motion capture at scale. From \texttt{HOT3D}, it produces 223k paired human and feasible robot trajectories across 3.3k object geometries with dense contact-force labels. We also produce 3.4k retargeted trajectories for two long-horizon tasks requiring millimeter precision. \new{When applied to policy learning, \ABBR{} enables training a visuomotor policy that achieves successful sim-to-real transfer using a five-fingered dexterous hand.}

\section{Related Work}

Dexterous retargeting converts human demonstrations into robot trajectories~\citep{spider,dexmachina,dextrack,chord,paliwal2026dexterous,handa2020dexpilot,qin2023anyteleop,bahety2024screwmimic,bahety2025safemimic,abbatematteo2019learning,abdul2022learning,wu2026grounding-galatea}. Kinematic retargeting aligns human and robot keypoints~\citep{dariush2008online,rohling1993optimized,antotsiou2018task,dariush2008whole,yin2025geometric,wu2026toporetarget}, but does not enforce dynamic feasibility. Physics-based retargeting instead aims at generating feasible trajectories through reinforcement learning (RL)~\citep{dextrack,chord,dexmachina} or sampling-based optimization~\citep{spider,paliwal2026dexterous,xue2025full}. RL suffers from tedious reward engineering and unstable, costly policy optimization. Sampling-based methods such as MPC avoid costly policy training but optimize each trajectory in isolation and fail to utilize knowledge in solved trajectories. Additional related work is discussed in App.~\ref{sec:app:extended-related-work}.

Real-to-sim methods reconstruct simulation environments from real data~\citep{torne2024reconciling,jiang2022ditto,dai2024automated,chen2024urdformer,fang2025rebot}, and have been utilized for robot data generation from human demonstrations~\citep{zhang2026robowheel,qin2022dexmv}. 
These datasets typically omit contact-force labels, with DexCanvas~\citep{xu2025dexcanvas} as an exception, but its labels describe interactions with an actuated MANO human-hand model rather than a robot hand.
Our data engine labels contact forces on the robot hand and also uses \ABBR{} for large-scale and high-precision retargeting, scaling to 223k physically feasible trajectories across 3.3k geometries.

\section{Generative Neural Retargeting}
\label{sec:method}
\subsection{Retargeting as a Generative Problem}
\label{sec:retargeting_as_generation}

\textbf{Physics-based retargeting problem.} Retargeting takes a human demonstration, represented either as human hand motion or as a robot kinematic reference obtained by IK, and asks for a dynamically feasible control sequence $\bm{u}_{0:T-1}$ over $T$ timesteps that reproduces the demonstrated behavior under physics. We denote the reference motion as $\bar{\bm{x}}_{0:T} = (\bar{\bm{x}}_0,\dots,\bar{\bm{x}}_T)$, where each $\bar{\bm{x}}_t$ contains the robot and object states at timestep $0 \le t \le T$. We follow the physics-based retargeting formulation that casts retargeting as a constrained optimization problem~\citep{spider}, $\min_{\bm{u}_{0:T-1}} J(\bm{u}_{0:T-1})$, subject to $\bm{x}_{t+1}=f(\bm{x}_t,\bm{u}_t)$ for $0\le t<T$, $\bm{x}_{0:T}\in\mathcal{X}$, and $\bm{u}_{0:T-1}\in\mathcal{U}$, where $\bm{x}_t$ is the simulated robot and object state, $f$ is the dynamics, $\mathcal{X}$ and $\mathcal{U}$ are the sets of admissible state and control sequences, and $J$ is the retargeting cost (e.g., object tracking errors and force regularization).

\textbf{The scaling bottleneck of sampling-based optimization.} The retargeting problem is highly non-convex and discontinuous due to the contact-rich nature of dexterous manipulation~\citep{spider}. Sampling-based optimizers solve it by numerous stochastic rollouts~\citep{spider, dexmachina, xue2025full} but exploring with task-agnostic noise makes it sample-inefficient to discover feasible trajectories. Thus, several methods mitigate the issue with hand-designed heuristics such as contact guidance and annealed noise schedules~\citep{spider}%, and virtual spring-damper constraints~\citep{dexmachina}.
Sample inefficiency and heuristics limit scalability. Furthermore, each trajectory is optimized independently, which fails to transfer knowledge across demonstrations.

\textbf{Generative modeling with test-time scaling.} We posit that feasible trajectories concentrate on a low-dimensional manifold that can be learned by a model that samples from it. Given a set $\mathcal{D}$ of human demonstrations, \ABBR{} learns from a seed dataset of $N$ demonstrations retargeted by sampling-based methods such as DIAL-MPC~\citep{xue2025full}, then retargets the remaining $|\mathcal{D}|-N$ demonstrations. Increasing $N$ can improve model performance but requires costly seed generation. Test-time scaling complements data scaling by improving success through sampling, reducing reliance on seed datasets. Notably, \ABBR{} can outperform the seed generator and solve demonstrations on which it fails, expanding dataset coverage while reducing generation cost.

% yue's claude update: E2: residual-over-IK was never motivated in the main text; the appendix ablation supports it
% \textbf{Flow-matching~\citep{lipman2022flow}.} We model \ABBR{} as a conditional flow matching generator $G_\theta(\cdot\mid\bm{c})$ of residual corrections $\bm{\Delta}=\bm{u}_{0:T-1}-\bar{\bm{u}}_{0:T-1}$ to the IK reference controls $\bar{\bm{u}}_{0:T-1}$, conditioned on $\bm{c}=(\bar{\bm{x}}_{0:T},\bm{o})$, where $\bm{o}$ encodes object geometry or pose, depending on the task. For training pairs $(\bm{\Delta},\bm{c})$ from the optimizer-generated data, we independently sample standard Gaussian noise $\bm{z}_0\sim\mathcal{N}(\bm{0},\mathbf{I})$ and flow time $\tau\sim\mathcal{U}[0,1]$, set $\bm{z}_\tau=(1-\tau)\bm{z}_0+\tau\bm{\Delta}$, and train the velocity field $\bm{v}_\theta$ with the flow matching loss $\mathcal{L}_{\mathrm{FM}}(\theta)=\mathbb{E}_{(\bm{\Delta},\bm{c}),\bm{z}_0,\tau}\big[\|\bm{v}_\theta(\bm{z}_\tau,\tau,\bm{c})-(\bm{\Delta}-\bm{z}_0)\|_2^2\big]$. At inference, we draw Gaussian noise $\bm{z}_0$ and integrate $d\hat{\bm{z}}_\tau/d\tau=\bm{v}_\theta(\hat{\bm{z}}_\tau,\tau,\bm{c})$ from $\hat{\bm{z}}_0=\bm{z}_0$ to $\tau=1$, yielding the residual $\hat{\bm{\Delta}}=G_\theta(\bm{z}_0\mid\bm{c})=\hat{\bm{z}}_1$ and retargeted trajectory $\bm{u}_{0:T-1}=\bar{\bm{u}}_{0:T-1}+\hat{\bm{\Delta}}$.
\textbf{Flow-matching~\citep{lipman2022flow}.} We model \ABBR{} as a conditional flow matching generator $G_\theta(\cdot\mid\bm{c})$ of residual corrections $\bm{\Delta}=\bm{u}_{0:T-1}-\bar{\bm{u}}_{0:T-1}$ to the IK reference controls $\bar{\bm{u}}_{0:T-1}$, conditioned on $\bm{c}=(\bar{\bm{x}}_{0:T},\bm{o})$, where $\bm{o}$ encodes object geometry or pose, depending on the task. For training pairs $(\bm{\Delta},\bm{c})$ from the optimizer-generated data, we independently sample standard Gaussian noise $\bm{z}_0\sim\mathcal{N}(\bm{0},\mathbf{I})$ and flow time $\tau\sim\mathcal{U}[0,1]$, set $\bm{z}_\tau=(1-\tau)\bm{z}_0+\tau\bm{\Delta}$, and train the velocity field $\bm{v}_\theta$ with the flow matching loss $\mathcal{L}_{\mathrm{FM}}(\theta)=\mathbb{E}_{(\bm{\Delta},\bm{c}),\bm{z}_0,\tau}\big[\|\bm{v}_\theta(\bm{z}_\tau,\tau,\bm{c})-(\bm{\Delta}-\bm{z}_0)\|_2^2\big]$. At inference, we draw Gaussian noise $\bm{z}_0$ and integrate $d\hat{\bm{z}}_\tau/d\tau=\bm{v}_\theta(\hat{\bm{z}}_\tau,\tau,\bm{c})$ from $\hat{\bm{z}}_0=\bm{z}_0$ to $\tau=1$, yielding the residual $\hat{\bm{\Delta}}=G_\theta(\bm{z}_0\mid\bm{c})=\hat{\bm{z}}_1$ and retargeted trajectory $\bm{u}_{0:T-1}=\bar{\bm{u}}_{0:T-1}+\hat{\bm{\Delta}}$. App.~\ref{sec:shapefilter-ablations} explains the design choice of predicting residuals, which outperforms predicting absolute controls.

\subsection{\ABBR{} Deployment}
\label{sec:gnr_deployment}
\label{sec:test-time-scaling}
\label{sec:closed-loop}

% yue's claude update: App.~\ref on a table/figure rendered as "App. N"; changed to \Cref plus the appendix section
\textbf{Open-Loop \ABBR{} Retargeting.} \ABBR{} enables test-time scaling through its generative formulation: for a condition $\bm{c}$, we can draw $S$ latents $\{\bm{z}_0^{(k)}\}_{k=1}^{S}$ and decode $S$ candidate trajectories $\{\bm{u}_{0:T-1}^{(k)}\}_{k=1}^{S}$. Each candidate is replayed in simulation and checked against the task success criterion (see \Cref{tab:success-criteria}). Retargeting succeeds if any of the $S$ candidates passes, which we report as pass@$S$.

\textbf{\ABBR{} for MPC Initialization.} Closed-loop feedback remains valuable for long-horizon and high-precision tasks, where small errors compound over time and lead to catastrophic failure. \ABBR{} integrates naturally with MPC by supplying a high-quality on-manifold initialization, $\bm{u}_{0:T-1}=\bar{\bm{u}}_{0:T-1}+\hat{\bm{\Delta}}$, which MPC then refines.

\section{A Human-to-Robot Data Engine}
\label{sec:r2s}
\begin{figure}[tbp]
\centering
\includegraphics[width=\linewidth]{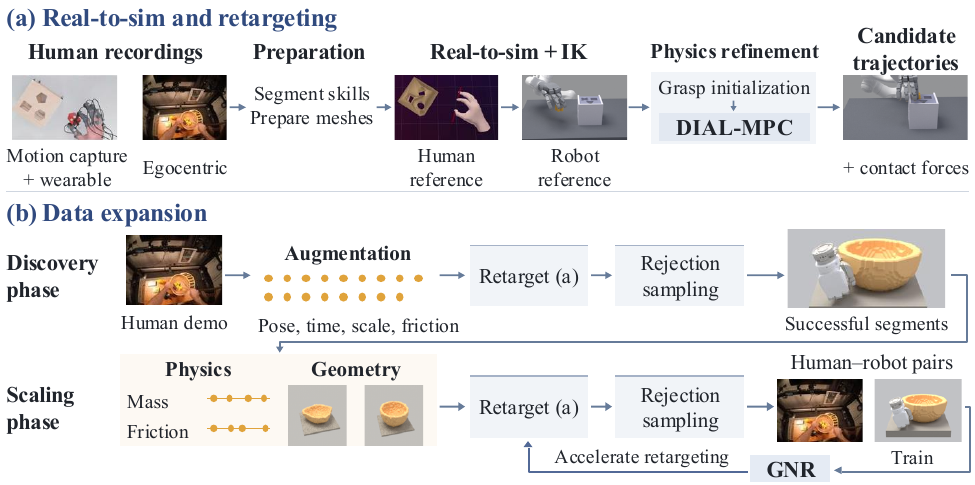}
\caption{\textbf{The human-to-robot data engine.}
(a) A pipeline retargets motion-capture recordings for the difficulty axis and egocentric demonstrations for the diversity axis into candidate robot trajectories with simulation-derived contact-force labels.
(b) The discovery and scaling phases repeatedly apply (a) under varied conditions and filter candidates using the same success criterion $\Gamma$ (\Cref{sec:r2s_diversity}), first identifying retargetable segments and then expanding their physical and geometric coverage.
The resulting pairs train \ABBR{} to accelerate subsequent retargeting.}
\label{fig:r2s_pipeline}
\end{figure}

\ABBR{} accelerates physics-based retargeting and drives a human-to-robot data engine (\Cref{fig:r2s_pipeline}) which produces feasible robot trajectories with simulation-derived contact-force annotations across both difficulty and diversity axes. We describe the retargeting pipeline in \Cref{sec:r2s_engine} and the discovery and scaling phases for expanding dataset diversity in \Cref{sec:r2s_diversity}.

\subsection{Human-to-Robot Retargeting}
\label{sec:r2s_engine}

The data engine takes per-frame human hand motion, object poses, and an object surface mesh $\mathcal{M}$ as input. We segment multi-skill recordings using hand--object contact estimated from finger-to-surface distances, while using single-skill recordings directly. Object meshes are repaired to obtain watertight surfaces, and their signed distance fields $\Phi$ are precomputed for penetration checks. We transfer the captured motion to a simulated human hand, then apply keypoint-based IK to retarget it to the robot. Together with the recorded object poses, the resulting robot configurations form the kinematic reference trajectory $\bar{\bm{x}}_{0:T}$. For segments starting with an object already grasped, we re-establish the grasp by opening the robot hand, closing its fingers around the temporarily fixed object, and then releasing the object to move freely in simulation. To construct the initial training pairs for \ABBR{}, we use DIAL-MPC~\citep{xue2025full} to optimize robot controls that track the reference object motion, recording simulation-derived contact forces during replay. Once trained, \ABBR{} accelerates subsequent retargeting in open loop or by initializing MPC for refinement.
Implementation details are provided in App.~\ref{app:data-engine}. For the difficulty axis, individually recorded motion-capture demonstrations are converted into paired training data through this pipeline without skill segmentation. For the diversity axis, egocentric demonstrations undergo the additional stages below.

\subsection{Two-Phase Data Expansion}
\label{sec:r2s_diversity}

To improve data diversity, we apply a two-phase data expansion pipeline to egocentric demonstrations (\Cref{fig:r2s_pipeline}b). The discovery phase identifies retargetable segments, while the scaling phase expands them across object geometry and physical properties.
\paragraph{Discovery Phase.}
\label{sec:r2s_stage1}
To avoid costly expansion of unsuccessful segments, we first search for feasible retargeting under a small set of candidate conditions. Reconstruction errors and the embodiment gap can hinder retargeting under the recorded conditions. We therefore search for feasible solutions by optimizing each segment under the nominal condition and sixteen single-axis perturbations of object position, orientation, start time, scale, or surface friction. However, low object-tracking error alone does not ensure valid physical interaction. Grasp initialization reduces initial penetration but does not prevent it during subsequent motion, and optimized controls may still produce excessive contact forces. For demonstrations with little object motion, low tracking error can also occur without a sustained grasp.
We therefore retain only simulated trajectories $\xi=(\bm{x}_{0:T},\bm{u}_{0:T-1})$ with $\Gamma(\xi)=1$, where $\Gamma(\xi) = \gamma_{\mathrm{track}}(\xi) \wedge \gamma_{\mathrm{pen}}(\xi) \wedge \gamma_{\mathrm{grasp}}(\xi) \wedge \gamma_{\mathrm{force}}(\xi)$.
Here, $\gamma_{\mathrm{track}}$ bounds object position and orientation errors relative to the reference;
$\gamma_{\mathrm{pen}}$ bounds finger-capsule penetration using $\Phi$;
$\gamma_{\mathrm{grasp}}$ requires a minimum sustained squeeze force; and
$\gamma_{\mathrm{force}}$ limits per-finger contact force. Thresholds are provided in \Cref{tab:success-criteria}.
We denote accepted trajectories as the discovery dataset  $\mathcal{D}_1$. Only segments with at least one accepted trajectory proceed to the scaling phase. % We also use $\Gamma$ to evaluate \ABBR{} trajectories on \texttt{HOT3D}.
\paragraph{Scaling Phase.}
\label{sec:r2s_stage2}
We expand successfully retargeted segments by varying object physics and geometry. For physical variation, we resample surface friction and compute object mass from mesh volume and a density sampled from an object-specific range estimated by a large language model. Geometric variation includes mesh regeneration from the same image with different random seeds, anisotropic deformation, completion from partially masked images, regeneration after changing the object's orientation about its vertical axis, and single-axis stretching. We use SAM3D for image-based mesh generation and apply deformation and stretching directly to the meshes. For each variant, we optimize control sequences under the modified conditions using multiple optimizer seeds and retain trajectories satisfying the same criterion $\Gamma$ used in discovery.
The additional accepted trajectories form $\mathcal{D}_2$, yielding the MPC-generated robot dataset $\mathcal{D}_{\mathrm{robot}}=\mathcal{D}_1\cup\mathcal{D}_2$.

\section{Experiments}
\label{sec:experiments}
We evaluate \ABBR{} along the diversity and difficulty axes. We first describe the experimental settings (\Cref{sec:exp:train-eval}), then evaluate \ABBR{}'s performance and efficiency 
on large-scale, diverse egocentric demonstrations
(\Cref{sec:exp:diversity}) and long-horizon, high-precision demonstrations
(\Cref{sec:exp:difficulty}).\new{~\Cref{sec:exp:bc_s2r} discusses policy learning using \ABBR{}-generated data and reports successful sim-to-real transfer without any real robot demonstrations.}

\subsection{Experimental Setup}
\label{sec:exp:train-eval}

\textbf{Simulator \& Embodiment.}
We use SuperDex Physics~\citep{superdex2026} as the simulator, which supports non-convex collision geometry and spatially distributed 3-D contact forces, enabling dense contact-force annotations. The target embodiment is a five-fingered Tesollo DG-5F hand for \texttt{HOT3D}, and a DG-5F hand mounted on a Franka FR3 arm for \texttt{ShapeFilter} and \texttt{NIST}.

\textbf{\ABBR{} Training and Sampling.} We implement \ABBR{} as a conditional flow matching model, which predicts the residual to correct the IK reference. The backbone is a temporal U-Net~\citep{ronneberger2015unet} with feature-wise linear modulation (FiLM)~\citep{perez2018film} to inject the IK reference and pose or geometry conditioning. We use linear projector to encode the IK and object poses. For geometry-based \ABBR{}, we use a Point Transformer V3~\citep{ptv3} to encode 1,024 object points. We sample from \ABBR{} using an Euler ordinary differential equation (ODE) solver with 10 steps~\citep{lipman2022flow}; model-specific settings are in App.~\ref{sec:optimizer_hyperparameters}. We write $S$ for the number of samples drawn per demonstration. Multiple samples can be drawn concurrently in a batch.

\textbf{MPC Sampling for Seed Dataset Generation.} We denote samples per MPC iteration by $n_s$ and the maximum number of MPC iterations in each optimization by $M$. Following the retargeting workflow in \Cref{sec:r2s_engine}, we use DIAL-MPC~\citep{xue2025full} as both a retargeting baseline and the seed dataset generator for all tasks. We use 300 samples for 10 MPC iterations on \texttt{HOT3D} and 500 samples for 6 iterations on \texttt{ShapeFilter}. For \texttt{NIST}, the three stages use 86, 86, and 172 samples, respectively, for 10 iterations per stage.

\textbf{Baselines.} On \texttt{HOT3D}, we compare against the widely adopted inverse kinematics (IK) baseline~\citep{pieper1969kinematics-ik} and four sampling-based methods: MPPI~\citep{williams2017information-mppi}, DIAL-MPC~\citep{xue2025full}, iCEM~\citep{pinneri2021sample-icem}, and CMA-ES~\citep{hansen2001completely-cma-es}. Sampling-based methods use the cost function $J$ in~\Cref{sec:retargeting_as_generation}. We report the \texttt{HOT3D} results in~\Cref{sec:exp:diversity} and additional \texttt{ShapeFilter} results in App.~\ref{app:shapefilter:baselines}. We compare \ABBR{} against SPIDER~\citep{spider} on its own benchmarks to evaluate multi-embodiment retargeting and \new{applicability in a different simulator, MuJoCo Warp~\citep{zakka2026mjlab} (see App.~\ref{sec:spider-sample-accounting}). We also compare against RL baselines on \texttt{HOT3D} (see App.~\ref{app:cmp-with-rl}).}

\new{\textbf{Metrics.} The primary metric is success rate (SR), measured using task-specific criteria in \Cref{tab:success-criteria}. For \texttt{HOT3D}, success requires all four criteria defined in~\Cref{sec:r2s_stage1} to pass: tracking ($\gamma_{\mathrm{track}}$), penetration ($\gamma_{\mathrm{pen}}$), grasp ($\gamma_{\mathrm{grasp}}$), and contact force ($\gamma_{\mathrm{force}}$). We report the violation rate for each criterion separately. On \texttt{ShapeFilter}, success requires the object centroid to enter the insertion region within the box. On \texttt{NIST}, success requires sufficient screw depth and alignment of the knob. We compare CPU cost per trajectory in core-hours (core-h) or core-seconds (core-s). We additionally report object position error $E_{\mathrm{pos}}$, object orientation error $E_{\mathrm{rot}}$, and finger-joint angle error $E_{\mathrm{finger}}$ relative to the reference trajectory~\citep{dexmachina,dextrack,chord}, as well as ADD-AUC10, the area under the object-pose accuracy curve for ADD thresholds up to 10\,cm~\citep{spider}. These tracking metrics measure how closely the object or fingers follow the reference trajectory but overlook physics such as excessive contact forces and penetration. Therefore, we also report the violation rate of force and penetration.}

\subsection{Scaling to Large-Scale and Diverse Retargeting}
\label{sec:exp:diversity}

This section demonstrates \ABBR{}'s sample efficiency and generalization advantages that enable scalable and efficient retargeting of large, diverse collections of human demonstrations.

\textbf{Collecting demonstrations from egocentric data.} Egocentric data has emerged as a cost-efficient source to expand data diversity and quantity for robot learning~\citep{zheng2026egoscale,ma2026humanscale}. We use human demonstrations from the egocentric \texttt{HOT3D} dataset~\citep{hot3d}, extracting 780 source motions covering 33 objects with an average duration of approximately 2\,s. The data engine in \Cref{sec:r2s} transfers these motions into simulation and constructs 360k augmented human demonstrations by varying object pose, geometry, physical properties, and motion timing.

\textbf{Retargeting coarse motion references.} However, reconstructed human motions from videos are often infeasible due to reconstruction errors. IK exacerbates infeasibility due to the embodiment gap. We therefore use physics-based retargeting. We employ DIAL-MPC~\citep{xue2025full}, which succeeds on 95k of 360k demonstrations (26.47\%) and requires 390k CPU core-h, equivalent to 31.7 days on 512 CPU cores. We refer to the successful and unsuccessful demonstrations as the MPC-success (95k) and MPC-failure (264k) datasets, respectively.

\textbf{\ABBR{} training and evaluation.} The costly MPC generation motivates the key questions: \textit{does \ABBR{} outperform MPC while reducing sampling, and does \ABBR{} generalize to OOD or MPC-failure demonstrations?} To examine this, we randomly sample a small training dataset for \ABBR{} using 28k demonstrations (about 30\% of the 95k MPC-success) covering 23 of 33 objects. Because running sampling-based baselines is prohibitively expensive, we randomly sample 1,000 test demonstrations from those with an unseen object (203), source grasp (338), or geometry variant (459); 272 are MPC successes and 728 are MPC failures.

We compare open-loop \ABBR{} with retargeting baselines in~\Cref{tab:baselines}. Fingertip IK is a deterministic method and is run once to obtain the metrics. IK directly matches the fingertip reference and thus has the lowest $E_{\mathrm{finger}}$, but achieves only $18.30\%$ success. We use IK to initialize \ABBR{} and all the sampling-based baselines, including MPPI, DIAL-MPC, iCEM, and CMA-ES, \new{which are run with three seeds. We report the mean and standard deviation}. Sampling-based methods directly optimize object tracking and generally achieve better tracking metrics. However, optimization can yield trajectories with better object tracking that still violate grasp or contact constraints, so better tracking alone does not ensure task success. These sampling-based methods achieve only $27.53$--$34.93\%$ success with up to 3,000 samples, whereas \textit{open-loop} \ABBR{} reaches $52.90\%$ and $56.10\%$ at $S=128$ and $S=256$, respectively. \ABBR{} yields the highest SR with a smaller sample budget, the lowest finger-tracking error among the non-IK methods, and the lowest grasp, force, and penetration violation rates. Compared with its seed generator DIAL-MPC, \ABBR{} increases SR from $28.13\%$ to $56.10\%$, while using $7.6\times$ less CPU time.

\begin{table*}[htb]
\centering
\caption{Performance comparison on \texttt{HOT3D}.}
\label{tab:baselines}
\scriptsize
\setlength{\tabcolsep}{1.5pt}
\newcommand{\hotSD}[2]{#1$\mathord{\pm}$#2}
\newcommand{\hotIK}[1]{\textcolor{gray}{#1}}
\begin{tabular*}{\textwidth}{@{\extracolsep{\fill}}lrrrrrrrrrrr@{}}
\toprule
Method & $S$ & core-s & SR (\%) & $E_{\mathrm{pos}}$ (mm) & $E_{\mathrm{rot}}$ ($^{\circ}$) & $E_{\mathrm{finger}}$ ($^{\circ}$) & AUC & $\gamma_{\mathrm{track}}$ & $\gamma_{\mathrm{pen}}$ & $\gamma_{\mathrm{grasp}}$ & $\gamma_{\mathrm{force}}$ \\
\midrule
\hotIK{IK} & \hotIK{1} & \hotIK{39.0} & \hotIK{18.30} & \hotIK{46.78} & \hotIK{27.46} & \hotIK{0.12} & \hotIK{46.90} & \hotIK{56.3} & \hotIK{29.4} & \hotIK{37.1} & \hotIK{15.9} \\
\midrule
DIAL-MPC & 3,000 & 2,689.3 & \hotSD{28.13}{1.27} & \hotSD{31.71}{0.24} & \hotSD{15.95}{0.44} & \hotSD{1.40}{0.01} & \hotSD{58.88}{0.52} & \hotSD{33.5}{1.1} & \hotSD{34.5}{0.5} & \hotSD{20.6}{1.0} & \hotSD{25.2}{0.7} \\
MPPI & 3,000 & 1,849.6 & \hotSD{27.53}{0.15} & \hotSD{33.15}{0.52} & \hotSD{16.95}{0.19} & \hotSD{1.85}{0.01} & \hotSD{57.34}{0.27} & \hotSD{35.2}{1.0} & \hotSD{33.3}{1.0} & \hotSD{21.6}{0.5} & \hotSD{24.7}{0.4} \\
iCEM & 3,000 & 3,862.1 & \hotSD{34.93}{0.49} & \hotSD{\textbf{25.14}}{0.33} & \hotSD{\textbf{11.63}}{0.17} & \hotSD{2.06}{0.02} & \hotSD{\textbf{61.83}}{0.31} & \hotSD{\textbf{20.8}}{0.2} & \hotSD{37.9}{1.2} & \hotSD{13.2}{0.6} & \hotSD{30.9}{0.6} \\
CMA-ES & 3,000 & 2,912.7 & \hotSD{32.13}{0.23} & \hotSD{\underline{29.55}}{0.59} & \hotSD{\underline{12.45}}{0.26} & \hotSD{1.07}{0.00} & \hotSD{\underline{60.45}}{0.11} & \hotSD{25.7}{0.5} & \hotSD{34.4}{0.9} & \hotSD{19.5}{0.1} & \hotSD{23.4}{1.0} \\
\midrule
\ABBR{} & 1 & \textbf{48.9} & \hotSD{18.80}{0.62} & \hotSD{44.85}{0.62} & \hotSD{25.26}{0.77} & \hotSD{1.00}{0.21} & \hotSD{48.28}{0.41} & \hotSD{55.3}{1.2} & \hotSD{31.3}{1.0} & \hotSD{34.8}{0.4} & \hotSD{18.6}{0.7} \\
\ABBR{} & 64 & \underline{90.9} & \hotSD{50.00}{0.53} & \hotSD{34.54}{0.50} & \hotSD{16.89}{0.25} & \hotSD{\textbf{0.97}}{0.04} & \hotSD{57.63}{0.13} & \hotSD{26.5}{0.2} & \hotSD{19.6}{0.6} & \hotSD{13.6}{0.2} & \hotSD{10.1}{0.5} \\
\ABBR{} & 128 & 178.6 & \hotSD{\underline{52.90}}{0.61} & \hotSD{34.08}{0.08} & \hotSD{16.48}{0.16} & \hotSD{\underline{0.98}}{0.04} & \hotSD{58.07}{0.22} & \hotSD{24.5}{0.3} & \hotSD{\underline{18.1}}{0.2} & \hotSD{\underline{12.1}}{0.4} & \hotSD{\underline{9.3}}{0.6} \\
\ABBR{} & 256 & 354.1 & \hotSD{\textbf{56.10}}{0.36} & \hotSD{33.85}{0.12} & \hotSD{16.42}{0.10} & \hotSD{\underline{0.98}}{0.03} & \hotSD{58.53}{0.10} & \hotSD{\underline{22.4}}{0.5} & \hotSD{\textbf{16.8}}{0.3} & \hotSD{\textbf{10.9}}{0.3} & \hotSD{\textbf{8.5}}{0.2} \\
\bottomrule
\end{tabular*}
\par\vspace{2pt}
\noindent\parbox{\textwidth}{\raggedright
$E_{\mathrm{pos}}, E_{\mathrm{rot}}, E_{\mathrm{finger}}$: object position, orientation, and finger-joint angle errors (medians). CPU core-s: mean per demonstration. AUC denotes ADD-AUC10, the area under the object-pose accuracy curve for ADD thresholds up to 10\,cm. The $\gamma$ columns report violation rates (\%). \textbf{Bold} and \underline{underline} denote the best and second-best values, respectively (excluding IK).}
\end{table*}

\Needspace{15\baselineskip}
% Limit wrapping to the figure and caption; restore full width afterward.
\begin{wrapfigure}[14]{r}{0.57\textwidth}
% \Needspace{18\baselineskip}
% \begin{wrapfigure}{r}{0.52\textwidth}
\vspace{-10pt}
\centering
\includegraphics[width=\linewidth]{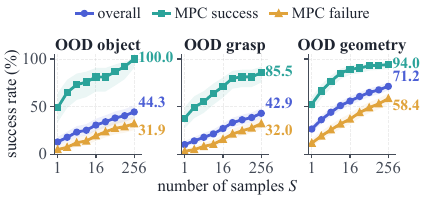}
\caption{\ABBR{} performance across three \texttt{HOT3D} OOD subsets. For each subset, we report \ABBR{}'s SRs on MPC-success, MPC-failure, and all demonstrations. }
\label{fig:ood_diversity}
\end{wrapfigure}

\textbf{\ABBR{}'s generalization.} \Cref{fig:ood_diversity} reports \ABBR{}'s performance on three OOD subsets. Test-time scaling improves SR across all three: on MPC-success demonstrations, the mean SR across subsets rises from $46.2\%$ at $S{=}1$ to $93.2\%$ at $S{=}256$. With $S{=}256$, \ABBR{} also solves $31.9$--$58.4\%$ of demonstrations that MPC fails to solve with 3,000 samples, expanding coverage beyond the seed optimizer with fewer samples.

We apply \ABBR{} trained on 28k seed trajectories to the 360k augmented demonstrations with up to $S{=}256$ samples, only $8.5\%$ of MPC's sampling budget. On the 264k MPC-failure demonstrations, \ABBR{} solves 127k ($48.01\%$). Together with the 95k MPC-success trajectories, this expands the dataset to 223k feasible trajectories across 3.3k geometries. On the MPC-success demonstrations, \ABBR{} achieves $89.30\%$ SR. These results support using \ABBR{} as a primary generator for large-scale human-to-robot datasets. See App.~\ref{sec:hot3d-dataset-expansion} for details.

\textbf{\ABBR{}'s efficiency.}
We compute the cost of a practical deployment in which DIAL-MPC generates a small seed dataset and \ABBR{} retargets the remaining demonstrations. At a success rate of 26.47\%, obtaining 28k seed demonstrations requires about 108k MPC attempts and 115.9k CPU core-h. Running \ABBR{} on the remaining 332k demonstrations at $S{=}256$ costs 32.6k CPU core-h. \ABBR{} solves 127k MPC-failure demonstrations and achieves $88.91\%$ SR on the 67k MPC-success demonstrations outside the seed set. This strategy yields about 215k trajectories at 148.6k CPU core-h, compared with 390k CPU core-h for the full DIAL-MPC run ($2.63\times$ lower cost).

\subsection{Solving High-Precision and Long-Horizon Retargeting}
\label{sec:exp:difficulty}

This section examines the scaling challenges along the difficulty axis where we use high-precision motion capture to collect human demonstrations for more complex tasks, and employ \ABBR{} to improve dataset generation efficiency. \new{Furthermore,~\Cref{sec:exp:bc_s2r} discusses using \ABBR{}-generated data for policy learning with behavioral cloning and sim-to-real policy deployment.}

\textbf{Scaling challenges for high-precision and long-horizon retargeting.} Existing dexterous retargeting methods primarily focus on shorter tasks that tolerate centimeter-level errors~\citep{spider,dextrack,dexplore,zhu2026dexflywheel} (see \Cref{tab:retargeting-dataset-comparison} for details), whereas precise manipulation often demands millimeter precision over long horizons. Scaling along this axis presents two challenges. \textit{First}, human motion recovered from visual observations, as in \texttt{HOT3D}, lacks the precision needed for effective optimization. \textit{Second}, RL and MPC require significantly more samples to optimize longer trajectories under tighter precision constraints, motivating higher-quality human motion references and more sample-efficient retargeting.

Accordingly, we collect human demonstrations using 15 OptiTrack cameras and an instrumented wearable exoskeleton to improve human motion precision~\citep{furtado2019comparative}. We consider two tasks: 1) \texttt{ShapeFilter} inserts a cross-shaped part into a matching hole with a tolerance of $1.96$\,mm; its dataset contains 3,037 demonstrations augmented from 605 human demonstrations. 2) \texttt{NIST}, an industrial assembly task~\citep{lian2021benchmarking}, involves grasping, transporting, aligning, and screwing a knob onto a threaded post; success requires horizontal error below $1$\,mm, screwing down by at least $8$\,mm, and axis error below $0.003$\,rad. Its dataset contains 358 demonstrations. Our real-to-sim pipeline reconstructs both tasks in SuperDex, where DIAL-MPC optimizes each reference trajectory for feasibility, providing the seed datasets and the baseline. Dataset and optimization details are in App.~\ref{sec:appendix_shapefilter} and App.~\ref{sec:appendix_nist}; demonstration examples are in App.~\ref{sec:retargeting_demos_shapefilter} and App.~\ref{sec:retargeting_demos_nist}.

\begin{figure}[!t]
\centering
\begin{subfigure}[t]{0.5\linewidth}
  \centering
  \includegraphics[width=\linewidth]{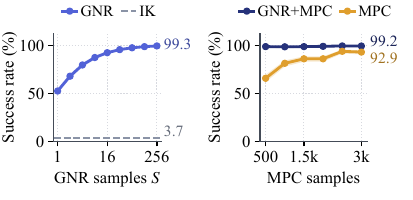}
  \caption{\texttt{ShapeFilter}}
  \label{fig:difficulty_results_shapefilter}
  \label{fig:difficulty_results_shapefilter_mpc}
\end{subfigure}%
\begin{subfigure}[t]{0.5\linewidth}
  \centering
  \includegraphics[width=\linewidth]{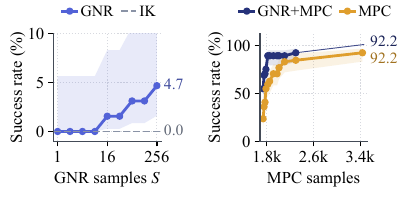}
  \caption{\texttt{NIST}}
  \label{fig:difficulty_results_nist}
  \label{fig:difficulty_results_nist_open_loop}
\end{subfigure}

\caption{\ABBR{}'s performance on high-precision and long-horizon tasks. On \texttt{ShapeFilter}, open-loop \ABBR{} improves with the number of samples $S$, and \ABBR{} warm starts reduce MPC sampling requirements. On \texttt{NIST}, open-loop sampling alone is insufficient, while using \ABBR{} to initialize MPC (\ABBR{}$+$MPC) outperforms MPC from scratch at matched sampling budget.}
\label{fig:difficulty_results}
\end{figure}

\textbf{\ABBR{} training and evaluation.} We use 2,448/589 training/test demonstrations for \texttt{ShapeFilter} and 294/64 for \texttt{NIST}. Following \Cref{sec:gnr_deployment}, we evaluate \ABBR{} open-loop and as an on-manifold initialization for MPC (denoted as \ABBR{}+MPC). On \texttt{ShapeFilter}, IK achieves only $3.7\%$ SR, while open-loop success rises from $52.4\%$ at $S{=}1$ to $99.3\%$ at $S{=}256$. At $S{=}64$, \ABBR{} exceeds MPC's $92.9\%$ SR with 3,000 samples ($M{=}6$, $n_s{=}500$), using only $1.6\%$ of its CPU cost. Combined with MPC, \ABBR{} achieves around $99\%$ SR across $M{=}1$ to $M{=}6$ iterations and consistently outperforms vanilla MPC initialized from IK at each $M$ (\Cref{fig:difficulty_results_shapefilter}). On \texttt{NIST}, open-loop \ABBR{} reaches $4.7\%$ SR at $S{=}256$, compared with $0\%$ for IK (\Cref{fig:difficulty_results_nist_open_loop}), motivating closed-loop feedback for timely corrections over long horizons. With \ABBR{} initialization, MPC reaches $92.2\%$ SR using 2,296 samples instead of the 3,440 used by vanilla MPC, a $33.3\%$ reduction (\Cref{fig:difficulty_results_nist}).

\begin{figure}[!t]
\centering
\begin{subfigure}[t]{0.5\linewidth}
  \centering
  \includegraphics[width=\linewidth]{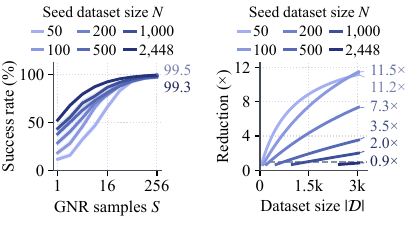}
  \caption{\texttt{ShapeFilter}}
  \label{fig:difficulty_efficiency_shapefilter}
  \label{fig:difficulty_efficiency_shapefilter_speedup}
\end{subfigure}%
\begin{subfigure}[t]{0.5\linewidth}
  \centering
  \includegraphics[width=\linewidth]{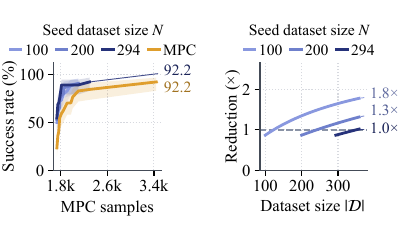}
  \caption{\texttt{NIST}}
  \label{fig:difficulty_efficiency_nist}
  \label{fig:difficulty_efficiency_nist_speedup}
\end{subfigure}
\caption{\ABBR{}'s efficiency in data generation. For each task, left: test-time scaling using \ABBR{} trained with different seed dataset sizes $N$. Right: CPU cost reduction as dataset size $|\mathcal{D}|$ grows. Costs include MPC seed generation and \ABBR{} generation.}
\label{fig:difficulty_efficiency}
\vspace{-1.5em}
\end{figure}

\Needspace{135pt}

\begin{wraptable}[11]{r}{0.40\textwidth}
\vspace{-\baselineskip}
% \begin{wraptable}{r}{0.40\textwidth}
\centering
\scriptsize
\setlength{\tabcolsep}{2pt}
\caption{Comparing efficiency and SR using MPC and \ABBR{}.}
\label{tab:difficulty-marginal-cost}
\begin{tabular*}{\linewidth}{@{\extracolsep{\fill}}lrrr@{}}
\toprule
Method & Samples & SR (\%) $\uparrow$ & core-s $\downarrow$ \\
\midrule
\multicolumn{4}{@{}l}{\texttt{ShapeFilter}} \\
MPC & 2,500 & 93.7 & 10,124 \\
\ABBR{} & 77 & 94.7 & 678 \\
\midrule
\multicolumn{4}{@{}l}{\texttt{NIST}} \\
MPC & 3,440 & 92.2 & 795,777 \\
\ABBR{}$+$MPC & 1,976 & 89.1 & 266,199 \\
\bottomrule
\end{tabular*}
% \par\vspace{2pt}
\end{wraptable}

\textbf{Data and test-time scaling of \ABBR{}.} Larger seed datasets improve \ABBR{}'s performance but require more costly MPC generation of the seed dataset. We ablate the seed dataset size $N$ in \Cref{fig:difficulty_efficiency_shapefilter,fig:difficulty_efficiency_nist}. On \texttt{ShapeFilter}, open-loop \ABBR{} success at $S{=}1$ increases from $11.5\%$ with 50 training demonstrations to $52.4\%$ with 2,448. Even with only 50 demonstrations, test-time scaling raises success to $99.5\%$ at $S{=}256$. On  \texttt{NIST}, we use \ABBR{} to provide on-manifold initializations for MPC. Increasing MPC samples raises \ABBR{}+MPC success from $54.7\%$ to $92.2\%$ with $N=294$ demonstrations and from $50.0\%$ to $90.6\%$ with $N=100$. \ABBR{} trained with $N=294$ and $N=100$ uses $33.3\%$ and $38.8\%$ fewer MPC samples, respectively, than IK-initialized MPC at $92.2\%$ success. Therefore, \ABBR{}'s test-time scaling complements data scaling, reducing reliance on costly MPC-generated seed data.

\textbf{\ABBR{}'s efficiency.} \Cref{tab:difficulty-marginal-cost} reports that \ABBR{} reduces CPU cost by $14.9\times$ on \texttt{ShapeFilter} and $3.0\times$ on \texttt{NIST} when achieving similar SRs. We then compute the total cost of generating each dataset, including MPC generation of $N$ successful seeds and retargeting the remaining $|\mathcal{D}|-N$ demonstrations with \ABBR{} or \ABBR{}$+$MPC. For 3,037 \texttt{ShapeFilter} demonstrations, $N{=}50,S{=}77$ gives a $94.8\%$ dataset success rate at 765.6 CPU core-h, versus $93.7\%$ at 8,541.0 core-h for MPC, an $11.2\times$ reduction.  (\Cref{fig:difficulty_efficiency_shapefilter_speedup}). For 358 \texttt{NIST} demonstrations, $N{=}100$ with a total MPC sample budget of 1,976 gives a $92.1\%$ dataset success rate at 44,516.7 core-h, versus $92.2\%$ at 79,135.6 core-h for MPC, a $1.8\times$ reduction (\Cref{fig:difficulty_efficiency_nist_speedup}).

\begin{figure}[H]
\centering
\includegraphics[width=\linewidth]{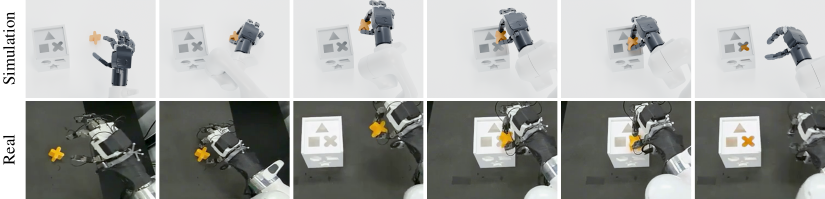}
\caption{\new{Real-world policy deployment on \texttt{ShapeFilter} using a Tesollo DG-5F hand and a Franka FR3 arm. Top: a \ABBR{}-retargeted demonstration in the rendered simulation dataset. Bottom: a real-world policy rollout using a diffusion policy trained with \ABBR{}-generated data.}}
\label{fig:sim_real}
\end{figure}

\Needspace{18\baselineskip}
\new{
\subsection{\ABBR{} for Policy Learning}
\label{sec:exp:bc_s2r}

\begin{wrapfigure}{r}{0.33333\linewidth}
\vspace{-\intextsep}
\centering
\includegraphics[width=\linewidth]{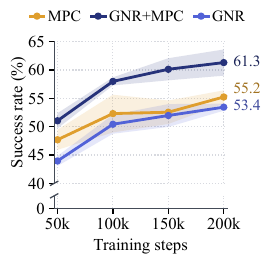}
\caption{SR curves of diffusion policy training using data produced by different generators.}
\label{fig:bc_sr_curves}
\end{wrapfigure}

\ABBR{} can generate demonstrations for dexterous policy learning. We train diffusion policies with behavioral cloning (BC)~\citep{chi2024diffusionpolicy} and show that \ABBR{} produces successful policies while significantly reducing generation cost. Then, we demonstrate that \ABBR{} achieves successful sim-to-real policy transfer without real-robot demonstrations on \texttt{ShapeFilter}.

\textbf{Examining Data Quality for BC.} We train state-based diffusion policies for 200k steps on data generated by \ABBR{}, \ABBR{}+MPC, and MPC, with three random seeds per dataset. We evaluate every 50k steps using 500 episodes per evaluation. As reported in~\Cref{fig:bc_sr_curves}, using \ABBR{} data alone achieves similar success to MPC, while \ABBR{}+MPC improves SR by 6.1\,pp over MPC and 7.9\,pp over \ABBR{} alone at 200k steps. This is because closed-loop feedback further enhances precision and \ABBR{} provides smoother and more feasible initializations, showcasing a trade-off between generation cost and precision.

\textbf{Sim-to-Real Policy Deployment.} Based on the  data quality analysis above, we train a visuomotor policy on \ABBR{}+MPC data rendered in SuperDex~\citep{superdex2026} and deploy it on the real robot. We use the same four-camera setup in simulation and on the real robot, with left and right overhead cameras, a wrist camera, and a front camera. We train the policy without real-robot demonstrations. The policy achieves $38\%$ success over 200 episodes in simulation and $30\%$ success over 10 real-robot episodes (\Cref{fig:sim_real}). To our knowledge, this is the first retargeting study to demonstrate millimeter-precision sim-to-real transfer for five-fingered dexterous manipulation, while existing studies generally use centimeter precision over a few seconds~\citep{spider,dextrack,chord,dexmachina}.}

\section{Conclusion}
\label{sec:conclusion}

We propose Generative Neural Retargeting (\ABBR{}), which learns from dynamically feasible trajectories and amortizes physics-based retargeting across demonstrations. Increasing test-time sampling boosts \ABBR{}'s success, complementing data scaling when feasible training trajectories are costly. \ABBR{} generalizes across motions and geometries, achieves higher success than MPC with far fewer samples, and retargets demonstrations on which MPC fails. We apply \ABBR{} within a real-to-sim data engine that scales retargeting along two axes: diversity, producing a retargeting dataset with 223k robot trajectories spanning 3.3k geometries; and difficulty, producing 3.4k robot trajectories for two challenging millimeter-precision, long-horizon tasks. \new{We demonstrate that behavioral cloning policies trained using \ABBR{} data achieve sim-to-real transfer to a five-fingered dexterous hand on a millimeter-precision task without any real robot demonstrations.}

% arXiv version: the ICLR-specific AI use, ethics, and reproducibility statements are omitted;
% they remain in iclr2027_conference.tex. References follow the conclusion directly.

%\subsubsection*{Author Contributions}
%If you'd like to, you may include  a section for author contributions as is done
%in many journals. This is optional and at the discretion of the authors.

%\subsubsection*{Acknowledgments}
%Use unnumbered third level headings for the acknowledgments. All
%acknowledgments, including those to funding agencies, go at the end of the paper.

\bibliography{iclr2027_conference}
\bibliographystyle{iclr2027_conference}

% Use tighter line spacing only in the appendix table of contents.
\newpage
\appendix
\etocdepthtag.toc{appendix}
\begin{center}
{\Large\bf Appendix}
\end{center}
\etocsettagdepth{mainbody}{none}
\etocsettagdepth{appendix}{subsection}
\etocsettocstyle{}{}
{\setlength{\parskip}{0pt}\linespread{0.90}\selectfont\tableofcontents}
\newpage
\raggedbottom
\section{Extended Related Work}
\label{sec:app:extended-related-work}

\subsection{Comparing Retargeting Datasets}
\label{sec:app:comparing-retargeting-datasets}

\Cref{tab:retargeting-dataset-comparison} compares datasets used by recent retargeting works along the diversity and difficulty axes. As discussed in~\Cref{sec:intro}, the computational cost and engineering burden of physics-based retargeting limit scaling along both axes.

Along the diversity axis, most works evaluate on a few thousand trajectories with limited object geometries. We apply \ABBR{} within our real-to-sim data engine (\Cref{sec:r2s}) to scale both the number of trajectories and the range of geometries from egocentric videos in \texttt{HOT3D}~\citep{hot3d}. This produces 223k paired human and feasible robot trajectories.

Along the difficulty axis, existing works primarily evaluate on short-horizon tasks lasting a few seconds~\citep{spider,dextrack}, or use longer demonstrations lasting 10 to 30 seconds with success criteria that tolerate centimeter-level errors~\citep{chord,dexmachina}. Long-horizon tasks requiring millimeter-level precision remain underexplored in existing retargeting work. Such tasks require accurate human motion references, but motion recovered from egocentric videos may lack the precision needed for successful retargeting. We therefore use motion capture with a wearable exoskeleton to collect high-precision human demonstrations for \texttt{ShapeFilter} and \texttt{NIST}. These long-horizon, multi-stage tasks require millimeter-level precision and are, to our knowledge, among the \textit{most challenging} tasks studied in dexterous retargeting.

\begin{table}[H]
\centering
\caption{Recent physics-based retargeting methods on the two scaling axes: \emph{diversity} (number of demonstrations and distinct geometries) and \emph{difficulty} (task precision and average demonstration length).}
\label{tab:retargeting-dataset-comparison}
\footnotesize % v2 size: appendix tables use footnotesize (<=5 columns) or scriptsize (wider)
\setlength{\tabcolsep}{5pt}
\begin{tabular}{@{}lrrrl@{}}
\toprule
Method & Demonstrations & Geometries & Precision (mm) & Horizon \\
\midrule
DexTrack \citep{dextrack} & 3,585 & 257 & $100$ & short \\
AdaDexTrack \citep{adadextrack} & 2{,}765 & 50 & $100$ & short \\
ManipTrans \citep{maniptrans} & 3{,}300 & 1,200 & 30-80 & short \\

% DexMachina \citep{dexmachina} & 7 & 5 & $10$--$90$ & short to long \\
% yue's claude update: range/compound written with an en dash (--); house style avoids dashes
DexMachina \citep{dexmachina} & 7 & 5 & $10$ to $90$ & short to long \\
% REGRIND \citep{feng2026regrind} & 2 & 2 & --- & 9.0\,s \\
SPIDER \citep{spider} & 2{,}885 & 103 & $100$ & short \\
Do-As-I-Do \citep{paliwal2026dexterous} & 500 & N/A & $100$ & short \\
DexCanvas \citep{xu2025dexcanvas} & 12{,}000 & 30 & $50$ & short to long \\
DexFlyWheel \citep{zhu2026dexflywheel} & 2{,}000 & 80 & $20$ & short \\
CHORD \citep{chord} & 4{,}739 & 10 & $150$ & short to long \\
\midrule
% yue's claude update: range/compound written with an en dash (--); house style avoids dashes
\ABBR\ (\texttt{HOT3D}) & \textbf{223k} & \textbf{3.3k} & 150 & short (2\,s) \\
\ABBR\ (\texttt{ShapeFilter}) & 3{,}037 & 1 & $\mathbf{1.96}$ & long (11.0\,s) \\
\ABBR\ (\texttt{NIST}) & 358 & 1 & $\mathbf{1.0}$ & long (\textbf{29\,s}) \\
\bottomrule
\end{tabular}
\end{table}

\subsection{Dexterous Retargeting}

\paragraph{Robot Learning from Human Data.}
Human video provides a scalable source of manipulation data~\citep{shaw2023videodex,bharadhwaj2024track2act,papagiannis2025r+,liu2025egozero,ma2026humanscale}. Robot systems increasingly learn from internet and egocentric video~\citep{shaw2023videodex,wang2023mimicplay,li2024okami,wang2026humanego,freeman2026warped,wang2026humanoidumi,zhao2026towards}, but differences in morphology and actuation create an embodiment gap that prevents human motions from being transferred directly to robots~\citep{chen2026unit,zhao2026towards,wu2026foundation-lingbotvla2}. Retargeting has emerged as a promising approach to transfer human motions to robots~\citep{spider,dexmachina,dariush2008online}.

\paragraph{Kinematic Retargeting.}

Kinematic approaches match human and robot fingertips or keypoints and recover robot configurations with IK~\citep{dariush2008online,rohling1993optimized,antotsiou2018task,dariush2008whole,yin2025geometric,handa2020dexpilot,qin2023anyteleop,malate2026smooth,wu2026toporetarget}. Source motion may come from motion capture systems, wearable hardware, or video~\citep{bahety2024screwmimic,bahety2025safemimic,li2024okami,park2025demodiffusion,ma2026humanscale,wu2026foundation-lingbotvla2}. Existing works have explored object-aware retargeting~\citep{li2024okami}, policy refinement for retargeting~\citep{park2025demodiffusion}, and visual synthesis to reduce geometric or visual mismatch~\citep{freeman2026warped,shen2026ld4wam,wu2026foundation-lingbotvla2}, but kinematic constraints alone do not ensure dynamically feasible trajectories~\citep{malate2026smooth}.

\paragraph{Physics-based Retargeting.}
Physics-based methods produce dynamically feasible trajectories through RL~\citep{dextrack,chord,dexmachina,feng2026minimalist,adadextrack} or sampling-based control~\citep{spider,paliwal2026dexterous,xue2025full,malate2026smooth}. RL can suffer from costly and unstable policy optimization and often requires careful reward engineering~\citep{feng2026minimalist,spider,lee-rl-aloha,fukazawa2025rl-behavior-reg}. Sampling-based control avoids policy training but still optimizes each trajectory independently. \ABBR{} learns a generative model of dynamically feasible trajectories to improve sample-efficiency for retargeting new demonstrations. We deploy \ABBR{} open-loop, or use \ABBR{} to provide MPC with on-manifold initialization for contact-rich tasks. \ABBR{} generalizes to OOD demonstrations and solves demonstrations on which MPC fails, thus expanding the dataset while boosting efficiency.

There are works utilizing retargeting priors from RL generated trajectories to enhance retargeting~\citep{dextrack,adadextrack}. However, these methods rely on costly RL training, complex curriculum design, or reward engineering, limiting their ability to generate far larger retargeting datasets. \ABBR{} provides a scalable retargeting data engine through a general recipe: combining a simple seed generator that requires neither task-specific heuristics nor curriculum design with an efficient, generalizable neural retargeter with test-time scaling. This design enables retargeting at substantially larger scales across more diverse and challenging tasks.

Several works distill sampling-based MPC into task-specific robot control policies~\citep{zhou2024diffusion-mpc,kurtz2026generative-mpc}. \ABBR{} instead focuses on retargeting conditioned on human motion, learning a single multi-task retargeter that generalizes across a stream of diverse and challenging human demonstrations provided by a scalable real2sim data engine.

\section{Discussions}
\label{sec:ext-discuss}

% \new{\subsection{How Much Does the Cross-Episode Prior Help?}

% We ablate the benefit of sharing knowledge across demonstrations. Our data scaling experiments provide initial evidence in \Cref{fig:difficulty_efficiency}: on \texttt{ShapeFilter}, the success rate increases from 11.5\% to 52.4\% when sampling once from \ABBR{} as we increase the number of training demonstrations from 50 to 2,448 due to stronger cross-episode prior.

% Now, we add an experiment to ablate the benefit of the cross-episode prior for \ABBR{}. Specifically, we fit a \ABBR{} model for each test episode using only successful DIAL-MPC samples from that episode. We compare \ABBR{} trained across all episodes on the training set with \ABBR{} trained within each episode on the \texttt{ShapeFilter} test set. As shown in~\Cref{tab:shapefilter_cross_episode}, \ABBR{} with the cross-episode prior consistently achieves higher success across all sampling budgets $S$.}

% \begin{table}[H]
% \centering
% \begin{tabular}{lcc}
% \toprule
% Method & $S$ & SR (\%) \\
% \midrule
% \ABBR{} & 16 & 92.3 \\
%  & 32 & 95.4 \\
%  & 64 & 97.4 \\
%  \midrule
% \ABBR{} (within episode) & 16 & 37.8 \\
%  & 32 & 57.2 \\
%  & 64 & 78.1 \\
% \bottomrule
% \end{tabular}
% \caption{Comparison between \ABBR{} and \ABBR{} learning within episodes to ablate the effect of cross-episode learning.}
% \label{tab:shapefilter_cross_episode}
% \end{table}

\new{\subsection{Comparison with RL Baselines}
\label{app:cmp-with-rl}

RL has emerged as a promising approach for dexterous retargeting. However, RL can suffer from training instability and sample inefficiency~\citep{spider,dextrack}. Therefore, we compare \ABBR{} primarily against sampling-based optimizers in~\Cref{sec:exp:diversity} and use DIAL-MPC for seed data generation because it avoids costly per-trajectory policy training.

\paragraph{Introducing RL baselines.} We consider two RL-based methods, DexTrack~\citep{dextrack} and ManipTrans~\citep{maniptrans}. Both use PPO~\citep{schulman2017proximal} for per-trajectory optimization and reuse a shared model to facilitate RL optimization.

Both RL baselines first learn a shared model across demonstrations for motion tracking, similar to how \ABBR{} learns from seed demonstrations. DexTrack shares a hand-object tracker~\citep{dextrack}, and ManipTrans shares a hand-motion imitator~\citep{maniptrans} across demonstrations. Both then reuse the shared model to facilitate RL when retargeting additional demonstrations. DexTrack and ManipTrans report a total of only 3,585 and 3,300 trajectories. The scale is limited by the cost of both policy optimization and seed generation, \textit{which also motivates \ABBR{} to scale up human-to-robot retargeting to far larger datasets.}

We first measure the cost of generating seed data with per-trajectory RL, as in DexTrack and ManipTrans, and show that it is prohibitively expensive for a larger dataset as \texttt{HOT3D}. To make seed generation affordable at a larger scale, we adapt both methods by training the shared tracker and imitator on the same MPC seed dataset we used for \ABBR{}, instead of on trajectories generated by per-trajectory PPO. We compare RL baselines with and without the shared model against \ABBR{} on the same \texttt{HOT3D} test set.

\paragraph{Seed generation with per-trajectory RL.}
We first evaluate per-trajectory PPO in DexTrack and ManipTrans to measure the cost of generating seed demonstrations, which are required to initialize their shared tracker and imitator. We train each policy for up to 540k environment steps (about $S=10,000$ rollouts) and report SR and environment steps in~\Cref{tab:hot3d_rl_baselines}.

At 3,000 rollouts, DexTrack achieves $15.0\%$ SR at an estimated $1.95$ CPU core-h per trajectory, and ManipTrans achieves $12.3\%$ SR at $5.43$ CPU core-h per trajectory. DIAL-MPC achieves $28.13\%$ SR with $0.75$ CPU core-h per trajectory (~\Cref{sec:exp:diversity}).

In practice, generating 28k successful seeds with DIAL-MPC costs 116k CPU core-h, equivalent to 9.4 days on 512 CPU cores. At the per-trajectory cost and SR in~\Cref{tab:hot3d_rl_baselines}, DexTrack would require about 371k CPU core-h, or about 30 days on the same 512 CPU cores, to generate the same number of successful seeds.

\begin{table}[H]
\centering
\caption{RL baselines for seed data generation, without the shared model, on the \texttt{HOT3D} test set. Steps denote environment steps used in simulation rollouts. For \ABBR{} and DIAL-MPC, $S$ denotes the number of samples drawn. For RL, $S$ denotes the number of training rollouts within the environment-step budget.}
\label{tab:hot3d_rl_baselines}
\small
\begin{tabular}{@{}lrrr@{}}
\toprule
Method & $S$ & Steps & SR (\%) \\
\midrule
\ABBR{} & 256 & 13k & $\mathbf{56.10}$ \\
DIAL-MPC~\citep{xue2025full} & 3,000 & 162k & $28.13$ \\
\midrule
\multirow{4}{*}{DexTrack~\citep{dextrack}} & 270 & 14k & 9.2 \\
 & 1,000 & 54k & 12.0 \\
 & 3,000 & 162k & 15.0 \\
 & 10,000 & 540k & 16.0 \\
\midrule
\multirow{4}{*}{ManipTrans~\citep{maniptrans}} & 270 & 14k & 9.7 \\
 & 1,000 & 54k & 10.7 \\
 & 3,000 & 162k & 12.3 \\
 & 10,000 & 540k & 14.7 \\
\bottomrule
\end{tabular}
\end{table}

\paragraph{RL baselines with shared learning.}

To make shared training affordable, we use the same seed demonstrations to train \ABBR{} and to pretrain the DexTrack tracker and the ManipTrans hand-motion imitator. Following DexTrack~\citep{dextrack}, we roll out the shared tracker to obtain a baseline action sequence and train a residual PPO policy on top of it for each trajectory. Following ManipTrans~\citep{maniptrans}, we freeze the hand-motion imitator and use its output at each control step as the base action for a residual PPO policy.

We evaluate all three methods on the \texttt{HOT3D} test set and report results in~\Cref{tab:hot3d_rl_shared}. \ABBR{} reaches $56.10\%$ SR using 13k environment steps by sampling directly from its model, without policy optimization. DexTrack and ManipTrans need about $11.7\times$ more rollouts to exceed $50\%$ SR ($53.3\%$ and $51.7\%$ at 3,000 rollouts), requiring about $32\times$ and $73\times$ more CPU core-h per episode.

For the full \texttt{HOT3D} dataset, \ABBR{} adds 127k successful trajectories to the 95k DIAL-MPC successes, yielding about 223k trajectories at an additional cost of only 26k CPU core-h (2.1 days on 512 CPU cores). At the per-episode costs in~\Cref{tab:hot3d_rl_shared}, running DexTrack and ManipTrans with 3,000 rollouts on the same 264k MPC-failure demonstrations would instead require about 845k and 1.95M CPU core-h, or about 69 and 158 days on 512 CPU cores.

\begin{table}[H]
\centering
\caption{Comparison with RL baselines using a shared model pretrained on the same seed dataset on \texttt{HOT3D}. We report mean and standard deviation over three seeds.}
\label{tab:hot3d_rl_shared}
\small
\begin{tabular}{@{}lrrrr@{}}
\toprule
Method & $S$ & Steps & core-h / episode & SR (\%) \\
\midrule
\ABBR{} & 256 & 13k & 0.10 & $\mathbf{56.10} \pm 0.36$ \\
\midrule
DexTrack + shared tracker & 270 & 14k & 0.25 & $36.7 \pm 0.92$ \\
DexTrack + shared tracker & 1,000 & 54k & 1.00 & $46.7 \pm 1.94$ \\
DexTrack + shared tracker & 3,000 & 162k & 3.19 & $53.3 \pm 4.40$ \\
\midrule
ManipTrans + shared imitator & 270 & 14k & 0.83 & $30.0 \pm 0.57$ \\
ManipTrans + shared imitator & 1,000 & 54k & 2.45 & $45.9 \pm 1.30$ \\
ManipTrans + shared imitator & 3,000 & 162k & 7.35 & $51.7 \pm 5.60$ \\
\bottomrule
\end{tabular}
\end{table}
}

\paragraph{More about RL baselines.}

1) \textbf{DexTrack~\citep{dextrack}} first retargets individual trajectories with per-trajectory PPO to generate demonstrations for training a shared hand-object tracker with RL and imitation learning. In their paper, 100 trajectories are first sampled from each training dataset and optimized with per-trajectory PPO to train the initial tracker. DexTrack then uses the tracker's action sequence as the baseline for residual per-trajectory RL, and employs homotopy optimization, which solves a sequence of progressively harder tracking tasks, to obtain additional tracking demonstrations.

2) \textbf{ManipTrans~\citep{maniptrans}} first trains a shared hand-motion imitator with PPO on hand-only motion data without objects. It then freezes the imitator and trains a residual PPO policy for each hand-object trajectory. The residual policy uses per-step simulated contact forces as both observations and reward terms, together with a gravity and friction curriculum. Extracting the contact forces at each control step in a high-precision simulator can be slow. Therefore, our experiments in~\Cref{tab:hot3d_rl_shared} also show that ManipTrans costs more compute to solve a trajectory compared to DexTrack.

Compared with RL baselines, sampling-based optimizers such as DIAL-MPC are often more efficient on many retargeting tasks~\citep{spider}, because policy optimization in RL can be costly and unstable~\citep{spider,schulman2017proximal}. RL methods also often rely on careful reward engineering or curriculum design~\citep{chord,dextrack}, which adds complexity to algorithms. \ABBR{} uses MPC as its seed generator. Through its generalization and test-time scaling, \ABBR{} can solve demonstrations with unseen objects, geometries, or motions without any additional optimization or task-specific reward engineering.

\new{\subsection{Precision Requirements}

For \texttt{ShapeFilter} and \texttt{NIST}, we consider millimeter and submillimeter precision requirements over long horizons. Prior works generally use centimeter requirements over a few seconds~\citep{dextrack,dexmachina,spider}. Solving higher-precision tasks requires more accurate human references and more sampling for physics-based optimization. For these two tasks, we use motion capture with a wearable exoskeleton to obtain more precise human references, and use \ABBR{} to reduce the sampling cost of generating demonstrations.

For \texttt{HOT3D}, we derive human references from egocentric videos, following prior work on dexterous retargeting from video~\citep{dexmachina,dextrack,paliwal2026dexterous}. Recovering precise hand and object motion from these videos is challenging because of limited accuracy of reconstructed references. Following CHORD~\citep{chord}, we use 150\,mm for position and $40^\circ$ for orientation.

We also show that \ABBR{}'s performance gains are consistent when using tighter precision requirements. We evaluate \ABBR{} and DIAL-MPC~\citep{xue2025full} on \texttt{HOT3D} with tighter precision requirements. As shown in~\Cref{tab:hot3d_precision}, \ABBR{}, learned using the data generated with the 150mm requirements, retains higher SRs compared to DIAL-MPC under different precision requirements.}

\begin{table}[H]
\centering
\caption{Success rates on \texttt{HOT3D} under different precision requirements.}
\label{tab:hot3d_precision}
\small
\begin{tabular}{@{}rrrr@{}}
\toprule
Position (mm) & Orientation ($^\circ$) & DIAL-MPC SR (\%) & \ABBR{} SR (\%) \\
\midrule
150 & 40 & 28.13 & \textbf{56.10} \\
100 & 30 & 23.7 & \textbf{29.9} \\
50 & 20 & 14.5 & \textbf{18.9} \\
\bottomrule
\end{tabular}
\end{table}

\new{\subsection{Why Does \ABBR{} Generalize?}

\ABBR{} generalizes to unseen objects, geometries, and motions on \texttt{HOT3D}. Because the egocentric videos in \texttt{HOT3D} cover a limited set of objects, our real-to-sim engine broadens the training set coverage by stretching and deforming object meshes and varying motions and physical parameters. Objects with different overall geometries can share local geometry at contact, so these augmentations expose \ABBR{} to a wider range of contact configurations for grasping and moving objects. Test-time scaling complements this broader coverage by increasing the likelihood of generating a feasible trajectory with more \ABBR{} samples.

\ABBR{} also generalizes to unseen poses and motions on \texttt{ShapeFilter} and \texttt{NIST}. Beyond data scaling and test-time scaling, we show that this spatial generalization can be attributed in part to the design choice of predicting residual corrections to the IK reference versus predicting absolute controls, especially when few candidates are sampled (see \Cref{fig:appendix-spatial-support-residual} for details).

}

\new{\subsection{\ABBR{} Closed-Loop Deployment without MPC}

\texttt{NIST} is among the hardest tasks in dexterous retargeting studies, and is difficult to solve in open loop because of its long horizon and sub-millimeter precision requirements. Open-loop \ABBR{} achieves only $4.7\%$ success at $S=256$. This is because retargeting errors compound over long horizons without feedback, motivating closed-loop deployment.

\ABBR{}+MPC has been shown to be simple and effective in our experiments in~\Cref{sec:exp:difficulty}. We also experimented with a different closed-loop deployment of \ABBR{} in a receding-horizon manner. Specifically, \ABBR{} predicts actions over a 64-step window and uses the latest states to predict the next window. We test this closed-loop \ABBR{} without MPC on \texttt{NIST}. Across three seeds, receding-horizon \ABBR{} improves SR by 1.04\,pp over the open-loop baseline of $4.7\%$. The under-performance is due to the limited data with only about 300 demonstrations, insufficient for learning high-precision and long-horizon retargeting on \texttt{NIST}. In this setting, sampling-based optimization can be necessary to solve the task by exploring control sequences beyond the learned manifold using limited data.

\begin{table}[H]
\centering
\caption{Open-loop and closed-loop receding-horizon \ABBR{} on \texttt{NIST}.}
\label{tab:nist_closed_loop_gnr}
\small
\begin{tabular}{@{}lcc@{}}
\toprule
 & Open-loop \ABBR{} & Closed-Loop \ABBR{} w/o MPC \\
\midrule
SR (\%) & 4.7 & \textbf{5.7} \\
\bottomrule
\end{tabular}
\end{table}

}

\new{\subsection{Cost Function vs. Success Verifier}

Sampling-based methods such as DIAL-MPC refine successive time windows when run in a receding-horizon manner, and rely on a cost function to guide sampling within each window. At each time window, DIAL-MPC samples across iterations and uses a cost-weighted average to produce the control sequence for the next time window. At the final window, the averaged control sequence is used for success verification. In contrast, \ABBR{} directly proposes multiple complete trajectories in open loop, each of which can be verified for success in simulation in parallel.

In this section, we ablate trajectory selection for \ABBR{} and DIAL-MPC to evaluate \ABBR{}'s sample-efficiency gains under cost-based and success-based selection. We evaluate \ABBR{} and DIAL-MPC on the same \texttt{HOT3D} test set used in~\Cref{sec:exp:diversity}.

\textbf{1) \ABBR{} using the cost function for trajectory selection.}
We run \ABBR{} with only 256 samples and use the cost function to select the minimum-cost trajectory. Verifying the selected trajectory yields $34.2\%$ SR, outperforming DIAL-MPC at $28.1\%$ SR with up to 3,000 samples.

\textbf{2) \ABBR{}'s sample efficiency when DIAL-MPC uses the success verifier.}
DIAL-MPC uses the cost function to guide sampling and to average sampled trajectories into the final trajectory. To enable a success verifier, we modify DIAL-MPC to apply the verifier to the rollouts from its last sampling iteration. DIAL-MPC anneals its sampling noise over iterations, so the last iteration samples around the most refined mean. For our DIAL-MPC baseline with 3,000 samples (300 samples $\times$ 10 iterations), this yields 300 candidates at the last iteration. We report the SR and the optimization and verification costs in~\Cref{tab:hot3d_oracle_selection}. With success verification, DIAL-MPC's SR increases from $28.1\%$ to $50.4\%$, at an additional 533.6 core-s per demonstration for verifying the rollouts in simulation. Overall, \ABBR{} using $S=256$ still achieves a higher SR ($56.1\%$) with $9\times$ less CPU compute.

\begin{table}[H]
\centering
\caption{SR and CPU cost of \ABBR{} and DIAL-MPC using the success verifier on the \texttt{HOT3D} test set.}
\label{tab:hot3d_oracle_selection}
\small
\begin{tabular}{@{}lrrrr@{}}
\toprule
Method & $S$ & Optimization (core-s) & Verification (core-s) & SR (\%) \\
\midrule
\ABBR{} & 256 & -- & 354.1 & \textbf{56.1} \\
% DIAL-MPC & 3,000 & 2,689.3 & 533.6 & 28.13 \\
DIAL-MPC + success verification & 3,000 & 2,689.3 & 533.6 & 50.4 \\
\bottomrule
\end{tabular}
\end{table}
}

\new{\subsection{Why Does Policy Deployment Validate Physics Fidelity in Simulation?}
\label{app:policy-physics-fidelity}

A common way to validate the physics fidelity of retargeted trajectories is to replay them open loop on a real robot~\citep{spider}. However, open-loop replay is highly sensitive to small errors in the initial object pose and in robot actuation, especially on tasks such as \texttt{ShapeFilter} that require millimeter-level precision over long horizons. Without closed-loop feedback, these errors accumulate over the trajectory. Small changes in contact geometry and variability in friction make it hard to predict real-world contact outcomes~\citep{yu2016more}. A failed real-world replay could therefore reflect reset errors rather than physics fidelity during the sim-to-real transfer.

A closed-loop policy can correct small errors through feedback, and therefore better isolates whether the contact dynamics learned in simulation transfer to the real world. Our visuomotor policy in~\Cref{sec:exp:bc_s2r} is trained using only trajectories generated in simulation, without any real-robot demonstrations. The RGB images are rendered observations and do not alter the trajectories or the physics, so the policy learns its contact behavior entirely from simulated physics. Its real-robot SR ($30\%$) is close to its simulated SR ($38\%$) despite the additional visual gap (\Cref{fig:sim_real}), which indicates that the simulated contact dynamics match the real system.
}

\section{Experimental Settings}
\label{sec:impl}
\label{sec:optimizer_hyperparameters}

\subsection{\ABBR{} Architecture and Training}
\label{sec:gnr-architecture}

The \ABBR{} model for \texttt{HOT3D} combines a Point Transformer V3 geometry encoder, taking in 1,024 points as inputs, with a temporal U-Net of widths $(128,256,512)$. The geometry, conditioning, and time embeddings have dimensions 96, 256, and 128, respectively. The model has 19.57M trainable parameters and uses ten Euler steps for inference. We train it on 28,480 episodes using bfloat16 mixed precision and batch size 128 on one NVIDIA A100.

\texttt{ShapeFilter} and \texttt{NIST} models use temporal U-Net widths $(128,256,512)$, kernel size 5, group normalization, and a 128-dimensional time embedding. \texttt{ShapeFilter} conditions on the IK reference, arm joint state, and either the object pose or a geometry encoding from Point Transformer V3, using 64-frame sequences for training and inference. The main experiments use object poses. Geometry conditioning achieves similar performance and is reported in \Cref{tab:shapefilter-conditioning}. \texttt{NIST} conditions on the reference and knob pose. We train \texttt{ShapeFilter} for 100 epochs and \texttt{NIST} for 200 epochs. Both use AdamW with a constant learning rate of $4\times10^{-4}$, weight decay $10^{-6}$, gradient clipping at 1.0, EMA decay 0.999, batch size 128, and mixed precision on one GPU. \new{At inference, both models process demonstrations in 64-frame windows, repeating generation over the full sequence in open-loop.} 

\subsection{Evaluation}
\label{sec:evaluation-protocol}

We use $S$ for the number of \ABBR{} samples per demonstration, $n_s$ for samples per MPC optimizer iteration, and $M$ for the maximum number of iterations in each optimization, giving a sample budget of $n_sM$. We compare different methods with the success criterion reported in \Cref{tab:success-criteria}.

\begin{table}[H]
\centering
\caption{Success criteria of different benchmarks.}
\label{tab:success-criteria}
\footnotesize % v2 size: appendix tables use footnotesize (<=5 columns) or scriptsize (wider)
\setlength{\tabcolsep}{5pt}
\begin{tabular}{@{}p{0.17\linewidth}p{\dimexpr0.83\linewidth-2\tabcolsep\relax}@{}}
\toprule
Benchmark & Success criterion \\
\midrule
\texttt{ShapeFilter} & \raggedright Object centroid lies in the insertion region
$[-80,80]\!\times\![0,160]\!\times\![-80,80]$~mm. \tabularnewline
\addlinespace[3pt]
\texttt{NIST} & \raggedright Horizontal error $<1$~mm, screwing down by at least $8$~mm,
and axis error $<0.003$~rad. \tabularnewline
\addlinespace[3pt]
\texttt{HOT3D} & \raggedright Object position error $<150$~mm and orientation error $<40^\circ$.\newline
Mean capsule penetration $\leq5$~mm.\newline
Median squeeze force after grasp release $\geq1$~N.\newline
Per-finger force 95th percentile $\leq100$~N. \tabularnewline
\addlinespace[3pt]
SPIDER & \raggedright Mean object-position error $\leq0.10$~m and mean
object-orientation error $\leq0.50$~rad. \tabularnewline
\bottomrule
\end{tabular}
\end{table}

\subsection{Optimizer Baselines}
\label{sec:optimizer-baselines}

\paragraph{DIAL-MPC~\citep{xue2025full}.}
DIAL-MPC perturbs the control sequence using annealed sampling noise, evaluates the resulting simulation rollouts, and updates the control sequence. \Cref{tab:dial-settings} summarizes the configurations and hyperparameters used in the main experiments.

\new{When initializing DIAL-MPC with \ABBR{} (denoted as \ABBR{}+MPC), we can reduce the number of iterations $M$ or the number of samples per iteration $n_s$. For \texttt{ShapeFilter}, we reduce $M$ and keep $n_s=500$ fixed, giving total budgets of 500 to 3,000 samples for \ABBR{}+MPC in~\Cref{sec:exp:difficulty}. For \texttt{NIST}, we reduce both $M$ and $n_s$ across all three stages. For example, to obtain a total budget of 2,296 samples, we set $M=8$ in each stage and use $n_s=72$, $72$, and $143$ for three stages, respectively. \ABBR{} provides the initial control sequence before the first stage of DIAL-MPC.}

\begin{table}[H]
\centering
\caption{DIAL-MPC settings for the main experiments. $n_s$ denotes samples per iteration and $M$ the maximum number of MPC iterations; Samples gives $n_sM$ for each phase.}
\label{tab:dial-settings}
\scriptsize
\setlength{\tabcolsep}{4.5pt}
\begin{tabular}{@{}p{0.16\linewidth}lrrrrrl@{}}
\toprule
Benchmark & Phase & $n_s$ & $M$ & Samples & Horizon & Stride & Active coordinates \\
\midrule
\texttt{ShapeFilter} & insertion & 500 & 6 & 3,000 & 20 & 5 & arm and hand \\
& \multicolumn{7}{p{\dimexpr0.84\linewidth-2\tabcolsep\relax}@{}}{\raggedright Exploitation fraction $0.15$; exploitation noise $0.5$; temperature $0.08$; trajectory-noise decay $0.7$.} \\
\midrule
\texttt{NIST} & wrist alignment & 86 & 10 & 860 & 7 & 2 & end effector \\
& finger closure & 86 & 10 & 860 & 6 & 1 & thumb and index \\
& final refinement & 172 & 10 & 1,720 & 13 & 4 & arm and hand \\
& \multicolumn{7}{p{\dimexpr0.84\linewidth-2\tabcolsep\relax}@{}}{\raggedright Initial and final noise scales $1.0$; exploitation fraction $0.15$; exploitation noise $0.5$.} \\
\midrule
\texttt{HOT3D} & tracking & 300 & 10 & 3,000 & full & full & hand \\
& \multicolumn{7}{p{\dimexpr0.84\linewidth-2\tabcolsep\relax}@{}}{\raggedright Temperature $0.03$; trajectory-noise decay $0.99$; sampling noise scale decreases log-linearly from $0.10$ to $0.04$.} \\
\bottomrule
\end{tabular}
\end{table}

We evaluate SPIDER~\citep{spider} on SPIDER's multi-embodiment benchmark in~\Cref{sec:multi-embodiment}. We use SPIDER's public implementation and configurations, which combine iterative physics optimization with annealed sampling noise and contact guidance. It samples $n_s{=}1{,}024$ candidates per optimizer iteration and runs up to $M{=}16$ MPC iterations on \texttt{OakInk} and $M{=}32$ on \texttt{GigaHands}, with early stopping.

% MPPI uses a fixed noise scale of $1.0$, whereas SPIDER anneals it to $0.01$. The results are reported in~\Cref{sec:multi-embodiment}.

\paragraph{CEM, iCEM, and CMA-ES.}
We evaluate CEM, iCEM, and CMA-ES as additional sampling-based baselines. The optimizer settings are reported in~\Cref{tab:sampling-baseline-settings}.

\begin{table}[H]
\centering
\caption{Sampling-based optimizer settings.}
\label{tab:sampling-baseline-settings}
\scriptsize
\setlength{\tabcolsep}{4.2pt}
\begin{tabular}{@{}llrrrp{0.55\linewidth}@{}}
\toprule
Method & Benchmark & $n_s$ & $M$ & Samples & Sampling settings \\
\midrule
MPPI & \texttt{HOT3D} & 300 & 10 & 3,000 & \raggedright Softmax-weighted update using DIAL-MPC sampling noise settings. \tabularnewline
\midrule
CEM & \texttt{ShapeFilter} & \begin{tabular}[t]{@{}r@{}}32\\500\end{tabular} & 6 & \begin{tabular}[t]{@{}r@{}}192\\3,000\end{tabular} & \raggedright Elite ratio $0.10$; mean/variance smoothing $0.10$; white Gaussian noise; fixed sample count; optimizer reset at the start of each optimization. \tabularnewline
\midrule
iCEM & \begin{tabular}[t]{@{}l@{}}\texttt{ShapeFilter}\\\texttt{HOT3D}\end{tabular} & \begin{tabular}[t]{@{}r@{}}32\\300\end{tabular} & \begin{tabular}[t]{@{}r@{}}6\\10\end{tabular} & \begin{tabular}[t]{@{}r@{}}192\\3,000\end{tabular} & \raggedright Elite ratio $0.10$; smoothing $0.10$; colored-noise exponent $2.0$.\newline
 \tabularnewline
\midrule
CMA-ES & \begin{tabular}[t]{@{}l@{}}\texttt{ShapeFilter}\\\texttt{HOT3D}\end{tabular} & \begin{tabular}[t]{@{}r@{}}32\\300\end{tabular} & \begin{tabular}[t]{@{}r@{}}6\\10\end{tabular} & \begin{tabular}[t]{@{}r@{}}192\\3,000\end{tabular} & \raggedright Full covariance over optimized control dimensions. \tabularnewline
\bottomrule
\end{tabular}
\end{table}

\new{\subsection{Behavioral Cloning}}
\label{app:bc-and-sim2real-settings}

\new{
\paragraph{State-based Diffusion Policy Settings.} We use a state-based diffusion policy to conduct data quality analysis on \texttt{ShapeFilter}. We build the policy using a conditional 1D U-Net with channel widths $(512, 1024, 2048)$. The observations to the policy span two time steps and include arm and hand joint positions and velocities and the poses of objects. The policy predicts 16 actions, each consisting of a target hand pose and hand joint positions, and executes the first 8 actions before predicting the next action chunk. We train each policy for 200k gradient steps with a batch size of 1{,}024, an initial learning rate of $3\times10^{-4}$ with cosine decay, and an exponential moving average of the weights. We use 100 diffusion timesteps during training and 10 DDIM denoising steps at inference. We evaluate the policies in closed loop every 50{,}000 training steps on 500 episodes per evaluation seed. In~\Cref{fig:bc_sr_curves}, the curves show the mean and one standard deviation across three training seeds.

\paragraph{Vision-based Diffusion Policy Settings.} For real-world deployment on \texttt{ShapeFilter}, we train a visuomotor diffusion policy on \ABBR{}+MPC data for 200k steps. The dataset is rendered in SuperDex~\citep{superdex2026}. We use DINOv3 ViT-S/16~\citep{simeoni2025dinov3} to encode RGB images from two overhead cameras, a right wrist camera, and a front camera. We crop each $256\times410$ image to $224\times384$ pixels and use the image features from the three most recent observations, together with robot proprioception, to condition a U-Net backbone.  For training, the policy uses learning rate 1e-4, cosine schedule over 200k steps, batch 512, EMA 0.9999. For inference, the policy uses 20 DDIM steps to predict 16 actions in a chunk and executes the first 4 actions at 20\,Hz before predicting the next chunk.
}

\section{Data Engine Details}
\label{app:data-engine}

% Optimizer hyperparameters are listed in App.~\ref{sec:optimizer_hyperparameters}; \texttt{HOT3D} success thresholds and augmentation magnitudes appear in App.~\ref{sec:appendix_hot3d}.

\subsection{Real-to-Sim Transfer and Kinematic Retargeting}
\label{app:de-retargeting}

\paragraph{Capture to simulated human hand.}
The capture system and simulator represent joint motion differently and use different rest poses and relative bone lengths. We transfer motion through bone positions and orientations in a shared coordinate system. After transforming the captured motion into the simulator's coordinates, we use forward kinematics to obtain $21$ hand landmarks and construct coordinate frames for $19$ bones. The simulator's IK converts these targets into a $33$-DoF hand configuration while preserving the simulated hand's relative bone lengths. Mirroring is applied to both the recorded hand representation and the simulated hand model to align their handedness. We obtain joint positions from the hand landmarks rather than the bone-frame origins, which coincide with the wrist in the rest pose in this representation.

\paragraph{Simulated human hand to robot.}
The DG-5F's limited thumb motion across the palm and rigid palm prevent it from reproducing all human hand poses. We use the wrist pose and five fingertip positions as IK targets, initializing each frame with the previous solution to encourage smooth motion. Our simulator uses a right-handed DG-5F model, so we mirror the complete scene across the sagittal plane for left-handed segments. The resulting robot configurations and recorded object poses form the kinematic reference $\bar{\bm{x}}_{0:T}$.

\subsection{Skill Segmentation}
\label{app:de-segmentation}

We segment multi-skill recordings into single skills, while single-skill recordings are used directly.

\paragraph{Contact heuristic.}
When 3D hand and object states are available, as in \texttt{HOT3D}, we estimate hand-object contact from their relative positions. A segment is an uninterrupted sequence of frames in which at least two fingers lie within $20$\,mm of the object surface, measured using a k-d tree to search the hand landmarks obtained by forward kinematics. A segment ends if the object remains set down for more than $5$\,s. Gaps shorter than $0.5$\,s are merged, and very short segments are discarded. Each segment starts at first contact and extends past release. Segments are labeled left-handed, right-handed, or two-handed. Applied to \texttt{HOT3D}, this yields roughly $5{,}000$ segments over $33$ objects: about $40\%$ two-handed, $40\%$ right-handed, and $20\%$ left-handed. The subset used by the data engine is reported in \Cref{tab:hot3d-generation-census}.

\paragraph{Vision-language segmenter.}

When only RGB video is available, we use a vision-language model (VLM) to identify hand-object contact in sampled frames. We group consecutive frames by contact state, discard isolated single-frame changes, and merge gaps shorter than $0.7$\,s. The VLM then reviews sampled frames from each segment to correct which hand and object are involved. Otherwise, we detect contact using the 3D hand and object positions as discussed in the contact heuristic method above.

\subsection{Grasp Establishment}
\label{app:de-grasp}

Segments may begin with the object already grasped. IK aligns the fingertips but can leave the DG-5F palm or knuckles intersecting the object. This initial penetration can eject the object or trap it against the hand, producing low tracking error without a valid grasp.

We therefore re-establish the grasp in simulation. The hand starts open and closes around the object while the object is fixed in place with gravity disabled. After roughly $12$ frames, the constraint is gradually released and gravity is restored. Force and penetration are evaluated after the constraint is released.

\subsection{Contact-Force Labels}
\label{app:de-force}

We use a signed distance field to represent the object's collision geometry. A penalty contact model is employed to compute the normal force from penetration depth. The field gradient provides the surface normal. We execute the optimized control sequence open loop with the same proportional-derivative (PD) controller, recording the position, normal, penetration depth, and force at every contact point. We close holes in object meshes to make them watertight, so the signed distance field has a well-defined inside and outside. We run physics at a higher rate than control to reduce penetration between control updates, and apply a short temporal median filter per finger and force component to remove outliers.

\new{\subsubsection{Stiffness and Forces.}
\label{app:stiffness-force}

We generate contact-force labels by replaying the retargeted controls in simulation. During trajectory optimization, the fingers may penetrate slightly into the object. The position controller can maintain this penetration, producing large simulated forces under a penalty contact model~\citep{zakka2026mjlab}. Trajectories optimized for motion tracking can therefore generate large contact forces~\citep{xue2025full,spider}, a commonly seen issue in modern simulators~\citep{zakka2026mjlab,superdex2026}.

For example, when using MuJoCo Warp as the simulator~\citep{zakka2026mjlab} to replay an XHand trajectory retargeted by SPIDER~\citep{spider}, the hand lifts an object weighing $5.06$\,N successfully, and the object-position error is only $0.027$\,mm. However, the maximal per-finger force reaches $739$\,N. On \texttt{HOT3D}, at the default SuperDex stiffness of $10^9$\,Pa/m, $47.6\%$ of IK replays exceed $100$\,N even for small objects like Aria glasses. Many existing retargeting works primarily minimize object and finger tracking errors~\citep{spider} without considering  whether the contact forces are realistic. To generate more realistic forces, we lower the contact stiffness and set the force cap at $100$\,N to avoid demonstrations with excessive force magnitudes in the dataset.

\paragraph{Ablation of Stiffness.} We use a contact stiffness of $10^8$\,Pa/m, one tenth of the SuperDex default, and a $100$\,N per-finger force cap. We ablate stiffness using two settings. First, we keep the cap fixed at $100$\,N. Second, we match it to the 84.1st percentile of the IK replay force distribution at each stiffness, corresponding to the percentile of $100$\,N at $10^8$\,Pa/m. We compare \ABBR{} with 16 samples, IK, and DIAL-MPC with 3,000 samples under varying stiffness and force cap settings. As reported in \Cref{tab:hot3d-stiffness}, with a fixed force cap, IK violates the cap more often as stiffness increases, and the SRs of IK, \ABBR{}, and DIAL-MPC all decrease. At every tested stiffness, \ABBR{} at $S=16$ achieves a higher SR than IK by $16.2$--$22.7$\,pp and DIAL-MPC by $12.9$--$15.5$\,pp. The same conclusion holds when we adjust the force cap to match the percentile of the IK replay force distribution. With these caps, \ABBR{} outperforms IK by $22.7$--$29.1$\,pp and DIAL-MPC by $11.6$--$15.3$\,pp.

\begin{table}[H]
\centering
\caption{Ablation of contact stiffness on the \texttt{HOT3D} test set. We compare \ABBR{}, IK, and DIAL-MPC within each setting using either a fixed $100$\,N per-finger force cap or a cap matched to the percentile of the IK force distribution corresponding to the default $100$\,N at $10^8$\,Pa/m we used for the paper.}
\label{tab:hot3d-stiffness}
\label{tab:hot3d_contact_stiffness}
\small
\setlength{\tabcolsep}{4pt}
\begin{tabular}{@{}lrrrr@{}}
\toprule
Stiffness (Pa/m) & Force cap (N) & IK SR (\%) & DIAL-MPC SR (\%) & \ABBR{} $S=16$ SR (\%) \\
\midrule
$10^8$                    & 100 & 18.3 & 28.1 & \textbf{41.0} \\
$3 \times 10^8$           & 100 & 16.3 & 24.0 & \textbf{37.3} \\
$10^9$ (SuperDex default) & 100 & 14.5 & 16.8 & \textbf{32.3} \\
$3 \times 10^9$           & 100 & 11.6 & 13.3 & \textbf{27.8} \\
\midrule
$10^8$                    & 100 & 18.3 & 28.1 & \textbf{41.0} \\
$3 \times 10^8$           & 169 & 22.1 & 34.6 & \textbf{49.9} \\
$10^9$ (SuperDex default) & 278 & 28.4 & 44.4 & \textbf{57.2} \\
$3 \times 10^9$           & 387 & 31.1 & 48.6 & \textbf{60.2} \\
\bottomrule
\end{tabular}
\end{table}

}

\Needspace{16\baselineskip}
\begin{wrapfigure}[15]{r}{0.33\linewidth}
  \vspace{-\intextsep}
  \centering
  \includegraphics[width=\linewidth]{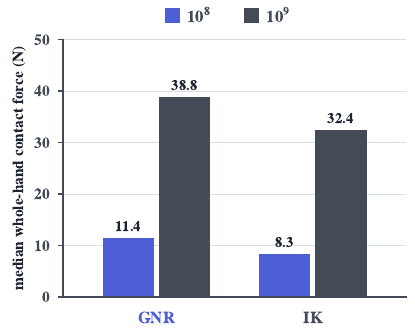}
  \caption{\new{Median whole-hand contact force on \texttt{HOT3D} under the $10^8$ and $10^9$ stiffness settings.}}
  \label{fig:hot3d-stiffness-force}
\end{wrapfigure}

\begin{figure}[H]
  \centering
  \includegraphics[width=0.73\linewidth]{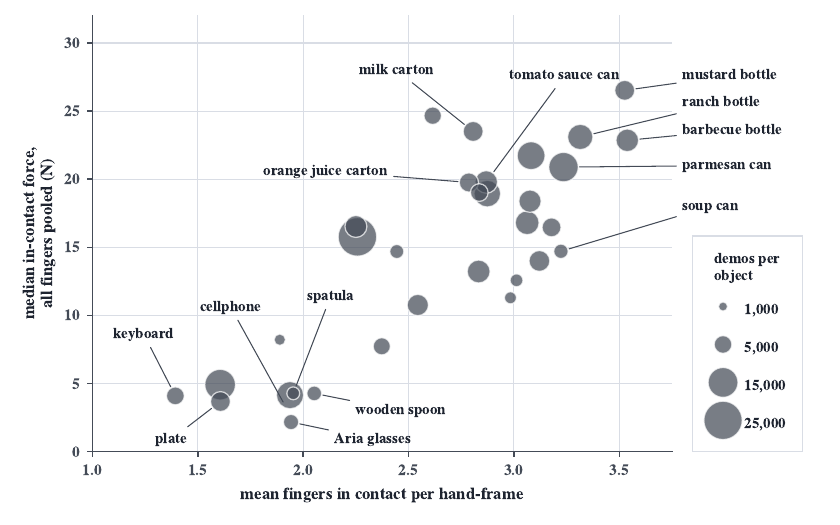}
  \caption{\new{Contact statistics by object in \texttt{HOT3D}. Each point shows the mean number of fingers in contact per hand-frame and the median contact force pooled over fingers and demonstrations.}}
  \label{fig:hot3d-object-force}
\end{figure}

\new{\noindent\textbf{Force Label Quality.}\quad At the lower-stiffness setting ($10^8$), the median whole-hand contact force is $11.4$\,N for \ABBR{} and $8.3$\,N for IK, compared with $38.8$ and $32.4$\,N at the default setting ($10^9$), respectively (\Cref{fig:hot3d-stiffness-force}), showing that we can use the lower stiffness to reduce excessive simulated forces.

We also examine the force labels from 223k \texttt{HOT3D} robot demonstrations. Bottles, cans, and cartons held with a whole-hand grasp have more fingers in contact and higher median contact forces than thin or small objects such as glasses and the spatula (\Cref{fig:hot3d-object-force}). Prior studies similarly report higher grip forces during power grasps than during several other ways of manipulating the same object~\citep{martinbrevet2017force}. The simulated magnitudes depend on the contact model and controller.\par}
\WFclear

\begin{figure}[H]
  \centering
  \includegraphics[width=0.70\linewidth]{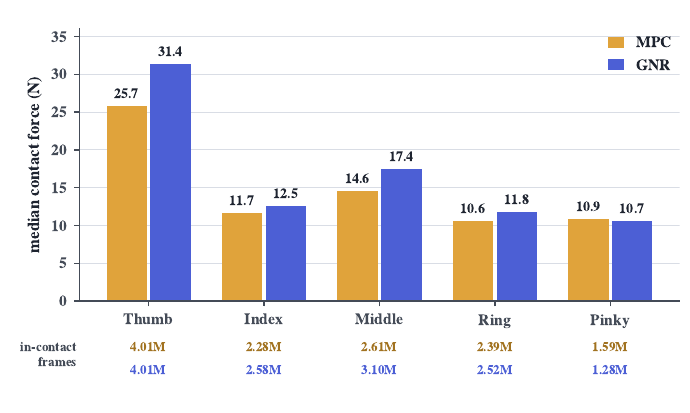}
  \caption{\new{Per-finger median simulated contact force and the number of contact frames for MPC and \ABBR{} demonstrations. For the per-object and per-finger statistics, we sum contact forces over each finger's distal links and count a finger as in contact when its force exceeds $0.1$\,N.}}
  \label{fig:hot3d-finger-force}
\end{figure}

\new{The thumb contacts the object most often ($81\%$ of hand-frames) and has the highest median contact force for both MPC and \ABBR{} (\Cref{fig:hot3d-finger-force}). This is consistent with human cylindrical grasps, where the thumb contributes the largest share of grip force~\citep{kinoshita1996grip,kuo2013force}.\par}

\subsection{Perturbations and Domain Randomization}
\label{app:perturbations}

\paragraph{\texttt{HOT3D} discovery phase.}
For each source segment, we evaluate the original scene and 16 perturbations of object translation, start time, scale, rotation, and friction. Translation shifts by $15$\,mm in either direction along the palm normal or grasp axis; start time shifts by $-5$, $-3$, $+3$, or $+5$ frames; scale is multiplied by $1.15$ or $1.30$; rotation shifts by $10^\circ$ in either direction about the up or grasp axis; and friction is multiplied by $0.5$ or $2.0$.

\paragraph{\texttt{HOT3D} scaling phase.}
We vary object density and surface friction and modify object geometry. Mass is computed from mesh volume and a density sampled from an object-specific range estimated by a large language model. For each resulting variant, controls are re-optimized and evaluated under physics using the same criterion $\Gamma$.

\paragraph{\texttt{ShapeFilter} domain randomization.}
Object and scene poses are perturbed per demonstration as shown in \Cref{tab:shapefilter-domain-randomization}; the box and lid move rigidly together. Captured demonstrations add variation beyond these ranges: initial cross positions span approximately $\pm167$\,mm and $\pm121$\,mm along the two table axes.

\begin{table}[H]
\centering
\caption{\texttt{ShapeFilter} domain-randomization settings.}
\label{tab:shapefilter-domain-randomization}
\footnotesize
\setlength{\tabcolsep}{5pt}
\begin{tabular}{@{}lll@{}}
\toprule
Scene element & Perturbation & Sampling range \\
\midrule
Cross-shaped object & horizontal position & $[-30,30]$ mm per axis \\
Cross-shaped object & yaw & $[-10,10]^{\circ}$ \\
Box and lid & horizontal position & $[-30,30]$ mm per axis \\
Box and lid & yaw & $[-2.5,2.5]^{\circ}$ \\
Complete scene & vertical position & $[-20,20]$ mm \\
\bottomrule
\end{tabular}
\end{table}

\subsection{Geometry Variants}
\label{app:de-geometry}

The scaling phase varies object physics and geometry as described in App.~\ref{app:perturbations}.

% yue's claude update: App.~\ref on a table/figure rendered as "App. N"; changed to \Cref
\Cref{tab:de-geometry-menu} lists the five geometry augmentations. Deformation and stretching preserve the correspondence between vertices in the original and modified meshes, allowing grasp contact points to transfer to the variant. The other three regenerate the mesh from a reference image with SAM3D~\citep{chen2026sam3d} and do not preserve this correspondence. Together, they define $200$ variants per object, or $6{,}600$ possible meshes across the $33$ source objects.

We replace the object mesh in a successfully retargeted scene, then re-optimize the control sequence. We sample $10$ geometry variants for each of $728$ source motions, with $24$ random optimizer seeds per pair, which gives $174{,}720$ trials. The $7{,}280$ motion variants involve $3{,}976$ distinct meshes, since motions involving the same object sample from the same set of geometry variants. Among these samples, $2{,}723$ yield at least one successful trajectory. Together with the $33$ original meshes, they give the $2{,}756$ geometries in the original MPC-success dataset.

\begin{table}[H]
\centering
\footnotesize % v2 size: appendix tables use footnotesize (<=5 columns) or scriptsize (wider)
\setlength{\tabcolsep}{5pt}
\caption{Geometry augmentations and variants per object.}
\label{tab:de-geometry-menu}
\begin{tabular}{@{}llr@{}}
\toprule
Edit & Implementation & Variants per object \\
\midrule
Regeneration          & Regenerate from the reference image with a new random seed & 20 \\
Deformation           & Deform the mesh by different amounts along each axis        & 60 \\
Occlusion completion  & Mask part of the reference image and complete the shape    & 20 \\
Rotation              & Rotate about the vertical axis, then regenerate            & 20 \\
Stretch               & Scale the original mesh along a single axis                & 80 \\
\midrule
\multicolumn{2}{@{}l}{Total}                                                        & 200 \\
\bottomrule
\end{tabular}
\end{table}

\section{\texttt{HOT3D}}
\label{sec:appendix_hot3d}

\subsection{Data Generation}
\label{sec:hot3d-data-generation}

Egocentric hand-object recordings are retargeted to the DG-5F hand with 26-dimensional joint control. The data engine augments 780 human motions from 210 recordings and 33 objects by varying pose, timing, rotation, scale, friction, and geometry. Among the 360,150 evaluated demonstrations, DIAL-MPC succeeds on 95,344 (26.47\%). \ABBR{} solves another 127,136 MPC-failure demonstrations, giving 222,480 successful trajectories within this population (\Cref{tab:hot3d-generation-census}).

\begin{table}[H]
\centering
\footnotesize
\setlength{\tabcolsep}{5pt}
\caption{\texttt{HOT3D} statistics for the 360,150 evaluated demonstrations. Columns separate demonstrations using original and modified object geometries; original geometries are augmented in pose, timing, mass, and friction. Trajectory counts and distinct geometry counts are reported separately.}
\label{tab:hot3d-generation-census}
\begin{tabular}{@{}lrrr@{}}
\toprule
Quantity & \shortstack{Original\\geometries} & \shortstack{Modified\\geometries} & Total \\
\midrule
Evaluated demonstrations & 186,310 & 173,840 & 360,150 \\
DIAL-MPC successes & 47,426 & 47,918 & 95,344 \\
DIAL-MPC failures & 138,884 & 125,922 & 264,806 \\
Additional \ABBR{} successes ($S{=}256$) & 74,391 & 52,745 & 127,136 \\
MPC or \ABBR{} successes & 121,817 & 100,663 & 222,480 \\
\midrule
Source motions & 780 & 728 & 780 \\
Distinct geometries & 33 & 3,976 & 4,009 \\
Geometries with MPC successes & 33 & 2,722 & 2,755 \\
Geometries with MPC or \ABBR{} successes & 33 & 3,265 & 3,298 \\
\bottomrule
\end{tabular}
\end{table}

\paragraph{Augmentation.}
Discovery and scaling perturbations are summarized in App.~\ref{app:perturbations}; geometry variants and their generation are detailed in App.~\ref{app:de-geometry}.

\paragraph{Success criterion.}
Success uses a position error threshold of 150\,mm (\Cref{tab:success-criteria}) following CHORD~\citep{chord}, rotation error below $40^\circ$, mean capsule penetration at most 5\,mm, median squeeze force after grasp release at least 1\,N, and per-finger 95th-percentile force at most 100\,N.

\subsection{Training \& Test Set}
\label{sec:hot3d-training-test-set}
\ABBR{} is trained on 28,480 DIAL-MPC successes from pose, timing, mass, and friction augmentations. The training set contains 1,838,731 frames (17.03 hours). Each demonstration has about 64 frames (2.15\,s at 30\,Hz). The dataset covers 23 objects and 353 source motions.

Because sampling-based optimizers are costly, we randomly sample a test set with 1,000 demonstrations and compare \ABBR{} with other sampling-based optimizers on this set. It comprises 203 OOD-object, 338 OOD-grasp, and 459 OOD-geometry demonstrations. \ABBR{} results are reported in \Cref{fig:ood_diversity}. Beyond this test set, we also apply \ABBR{} to the 360k full dataset. Results are reported in~\Cref{sec:hot3d-dataset-expansion}.

\subsection{Evaluation}
\label{sec:hot3d-evaluation}
\paragraph{Test set.}
\label{sec:hot3d-variation-atlas}

The \texttt{HOT3D} dataset contains 33 scanned objects in total. \ABBR{} used in~\Cref{sec:experiments} is trained on 23 objects and 353 source motions. The 1,000 test demonstrations are divided into three disjoint subsets: 203 with unseen objects, 338 with unseen grasps, and 459 with unseen geometries.

\begin{figure}[H]
\centering
\includegraphics[width=0.86\textwidth]{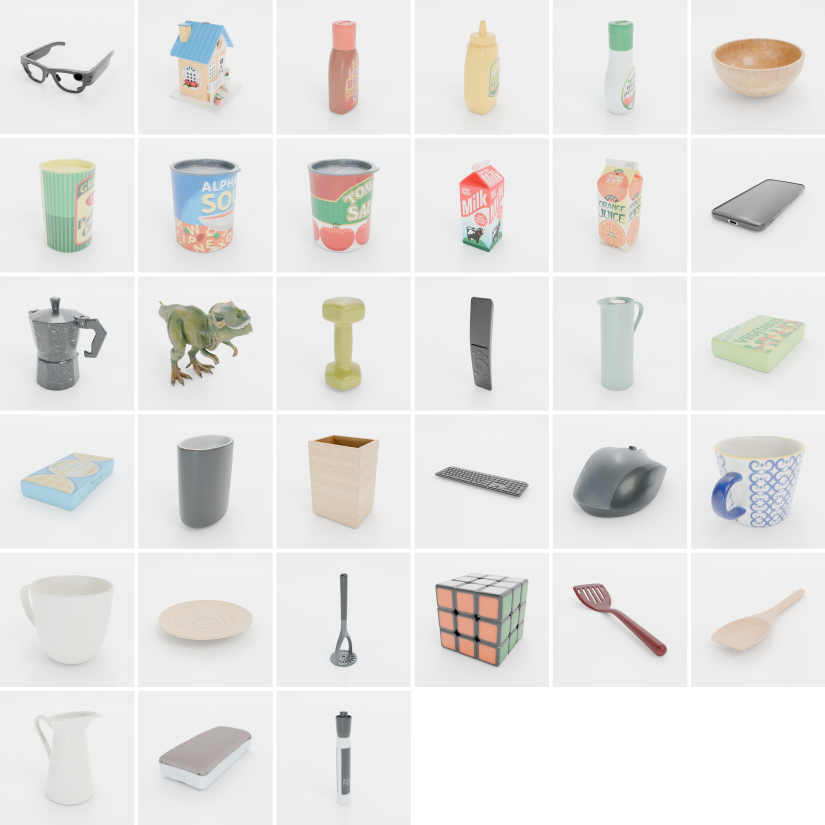}
\caption{Objects in \texttt{HOT3D}.}
\label{fig:appendix-hot3d-object-catalog}
\end{figure}

\begin{figure}[H]
\centering
\includegraphics[width=0.96\textwidth]{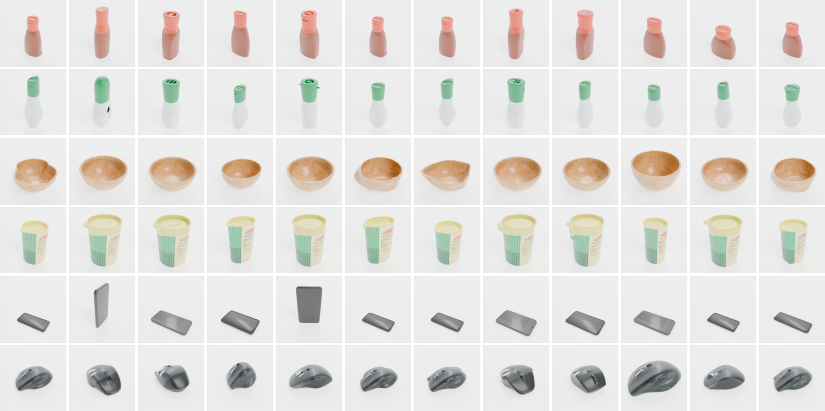}
\caption{Examples of geometry variants. Each row uses the same original object.}
\label{fig:appendix-hot3d-geometry-atlas}
\end{figure}

\begin{figure}[H]
\centering
\includegraphics[width=0.96\textwidth]{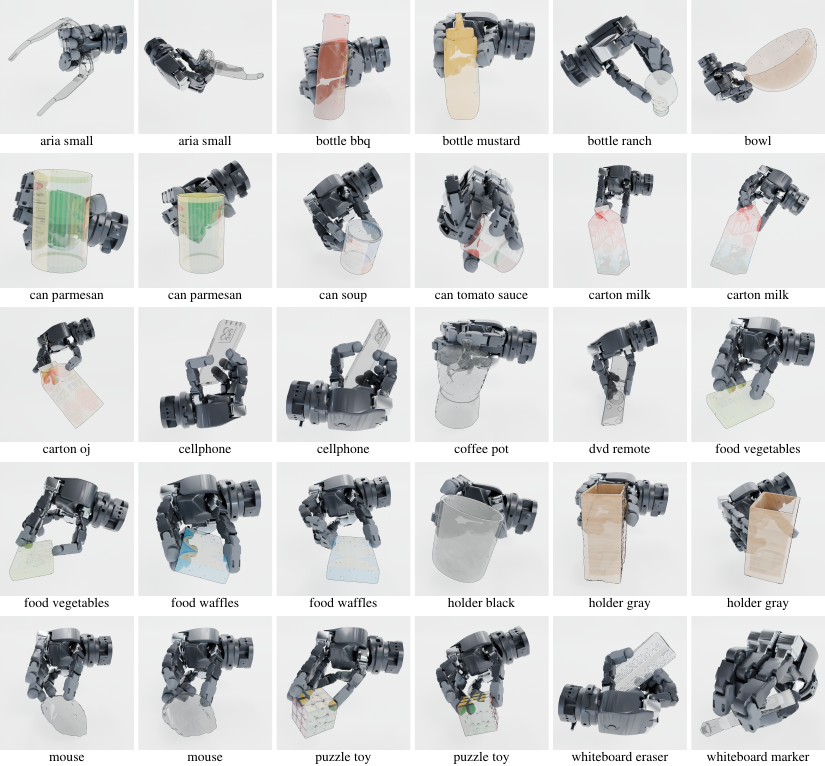}
\caption{Examples of grasps produced by \ABBR{} on \texttt{HOT3D}. Objects are rendered semi-transparent with an outline to show finger placement.}
\label{fig:appendix-hot3d-grasp-atlas}
\end{figure}

On the test set, \ABBR{} success rises from 12.81\% to 44.33\% on OOD objects, from 10.06\% to 42.90\% on OOD grasps, and from 26.36\% to 71.24\% on OOD geometry as $S$ increases from 1 to 256.

\Needspace{25\baselineskip}
\subsection{Dataset Expansion}
\label{sec:hot3d-dataset-expansion}

We apply \ABBR{} to the 264k MPC-failure demonstrations. For this expansion, \ABBR{} is trained on 28,480 MPC successes. A demonstration succeeds if any of its $S$ candidates succeeds; sampling stops upon success.

\begin{figure}[H]
\centering
\includegraphics[width=\linewidth]{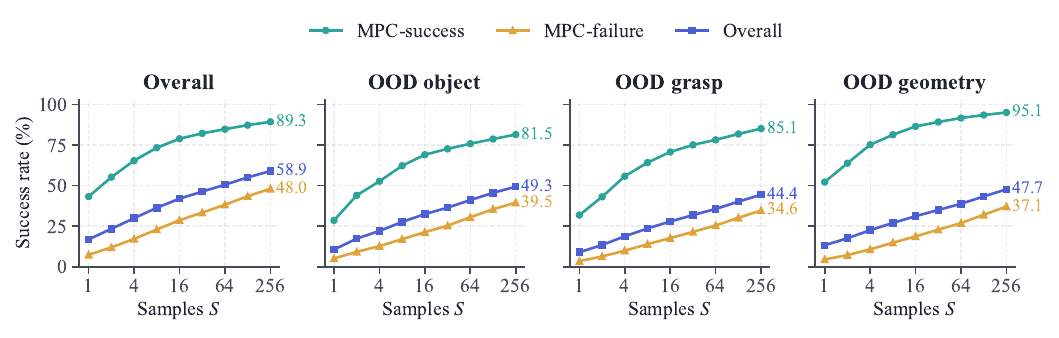}
\caption{Test-time scaling on 360k \texttt{HOT3D} demonstrations: 95k MPC-success and 264k MPC-failure demonstrations. The curves show \ABBR{}'s SR overall and across three OOD subsets.}
\label{fig:hot3d-dataset-expansion}
\end{figure}

\ABBR{} solves 19,384 MPC-failure demonstrations (7.32\%) at $S{=}1$, increasing to 101,129 (38.19\%) at $S{=}64$, 115,017 (43.43\%) at $S{=}128$, and 127,136 (48.01\%) at $S{=}256$ (\Cref{fig:hot3d-dataset-expansion}). Combining these additional trajectories with the 95k MPC successes yields 222,480 feasible trajectories. Geometry coverage grows from 2,756 to 3,298 (3.3k), including 542 new geometries.

We also apply the same model to the 95k MPC-success demonstrations. SR increases from 43.22\% at $S{=}1$ to 84.75\% at $S{=}64$ and 89.30\% at $S{=}256$ (85,143 successes). Among the demonstrations outside the training set, SR rises from 40.68\% to 88.91\%. Across the 360k demonstrations, \ABBR{} solves 212,279 demonstrations (58.94\%) at $S{=}256$.

\subsection{Trajectory Analysis}

\begin{figure}[H]
\centering
\includegraphics[width=\textwidth]{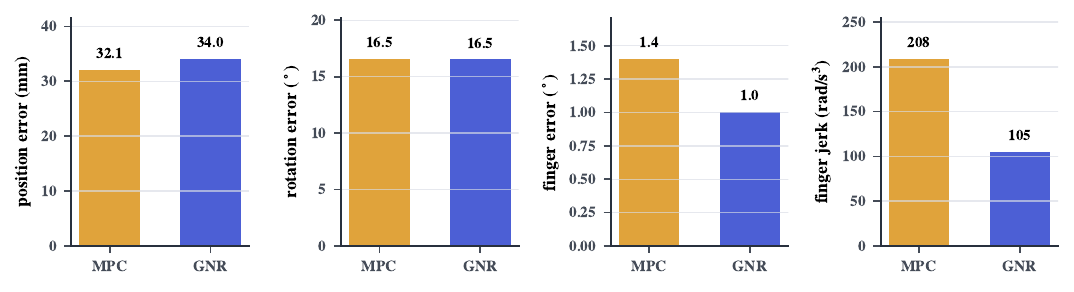}
\caption{\texttt{HOT3D} trajectory analysis. From left to right: median object position, object orientation, finger joint angular errors, and median finger control jerk on the test set with 1k demonstrations. \ABBR{} approximately halves jerk while preserving object tracking comparable to DIAL-MPC.}
\label{fig:appendix-hot3d-trajectory-quality}
\end{figure}

\Cref{fig:appendix-hot3d-trajectory-quality} compares object and finger joint tracking errors and finger control jerk.

\section{\texttt{ShapeFilter}}
\label{sec:appendix_shapefilter}

\subsection{Settings}
\label{sec:shapefilter-settings}
\paragraph{Task.}
A Tesollo DG-5F hand on a Franka FR3 grasps a cross-shaped object and inserts it into a matching hole on the box lid. The 29-D action comprises end-effector translation, a continuous 6-D rotation, and 20 hand joints. Control runs at 20\,Hz and physics at 60\,Hz.

\paragraph{Dataset.}
Pose augmentation of 605 human demonstrations produces 3,037 demonstrations and 665,770 frames. Unless otherwise stated, we use the motion split with 2,448 training and 589 test demonstrations from 484 and 121 underlying human demonstrations, respectively.

\paragraph{Success criterion.}
The object centroid enters the insertion region $[-0.08,0.08]\times[0,0.16]\times[-0.08,0.08]$ m in the box coordinate frame, indicating insertion completion. The insertion tolerance is 1.96\,mm.

\paragraph{Domain randomization.}
\label{sec:shapefilter-domain-randomization}

Object and scene perturbations are summarized in App.~\ref{app:perturbations} and \Cref{tab:shapefilter-domain-randomization}.

\subsection{Spatial OOD Evaluation}
\label{sec:shapefilter-evaluation}
\begin{figure}[H]
\centering
\includegraphics[width=0.40\linewidth]{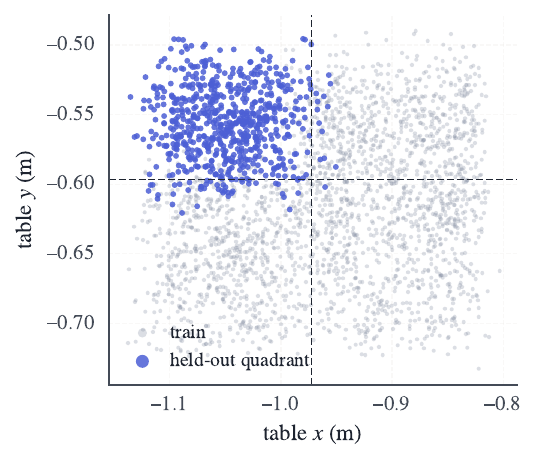}
\hfill
\includegraphics[width=0.40\linewidth]{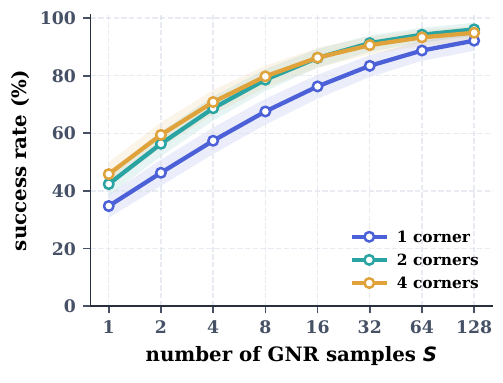}
\caption{\texttt{ShapeFilter} spatial generalization. The training (gray) and test (blue) quadrants are shown at left. We show the test-time scaling curves of \ABBR{} on the test split.}
\label{fig:appendix_spatial_split}
\end{figure}

We evaluate \ABBR{}'s generalization to spatial OOD demonstrations by creating three different spatial OOD training-test splits. The splits assign demonstrations by initial object position in the four quadrants of the table as shown in~\Cref{fig:appendix_spatial_split}. We create the splits using 1, 2, or 4 quadrants for training, and the remaining quadrants for testing.

\subsection{Baselines}
\label{app:shapefilter:baselines}
% SHAPEFILTER-GOLD589-TABLE-BEGIN
\begin{table}[H]
\centering
\footnotesize % v2 size: appendix tables use footnotesize (<=5 columns) or scriptsize (wider)
\setlength{\tabcolsep}{5pt}
\caption{\texttt{ShapeFilter} success rates. Optimizer budgets are samples per iteration $n_s$ $\times$ MPC iterations $M$.}
\label{tab:shapefilter-baselines}
\begin{tabular}{@{}lrr@{}}
\toprule
Method & $n_sM$ or $S$ & Success (\%) \\
\midrule
DIAL-MPC & $32\!\times\!6$ & 48.4 \\
DIAL-MPC & $500\!\times\!6$ & 92.9 \\
CEM & $32\!\times\!6$ & 50.0 \\
CEM & $500\!\times\!6$ & 82.8 \\
iCEM & $32\!\times\!6$ & 65.6 \\
CMA-ES & $32\!\times\!6$ & 40.6 \\
\midrule
\ABBR{} & $8$ & 87.23 \\
\ABBR{} & $16$ & 92.26 \\
\ABBR{} & $32$ & 95.44 \\
\ABBR{} & $64$ & 97.37 \\
\bottomrule
\end{tabular}
\end{table}

\subsection{Ablations}
\label{sec:shapefilter-ablations}

\paragraph{Conditioning.}
\label{sec:shapefilter-conditioning}

We compare three conditioning inputs. State uses the IK reference and simulator object state. Geometry uses a 1,024-point cloud as the condition. Geometry $+$ state provides both.

\begin{table}[H]
\centering
\footnotesize
\setlength{\tabcolsep}{5pt}
\caption{\texttt{ShapeFilter} conditioning ablation. We report success rates (\%) across different numbers of \ABBR{} samples $S$ using different conditioning.}
\label{tab:shapefilter-conditioning}
\begin{tabular}{@{}rrrr@{}}
\toprule
Samples $S$ & State & Geometry & Geometry $+$ state \\
\midrule
1 & 50.42 & 50.93 & \textbf{53.48} \\
2 & 67.06 & 66.38 & \textbf{69.10} \\
4 & 78.44 & \textbf{80.31} & 78.61 \\
8 & 87.10 & \textbf{88.62} & 86.59 \\
16 & 92.36 & 93.72 & \textbf{94.06} \\
32 & 95.93 & 96.77 & \textbf{97.11} \\
64 & 97.45 & 98.13 & \textbf{98.81} \\
128 & 98.47 & 98.81 & \textbf{99.15} \\
256 & 99.32 & \textbf{99.49} & 99.32 \\
\bottomrule
\end{tabular}
\end{table}

% v2 restructure: former subsection "Ablation of Residual Prediction"; text first, then the figure
\paragraph{Residual prediction.}
We compare absolute and residual prediction on \texttt{ShapeFilter} with one, two, or four training quadrants.

\begin{figure}[H]
\centering
\includegraphics[width=0.8\linewidth]{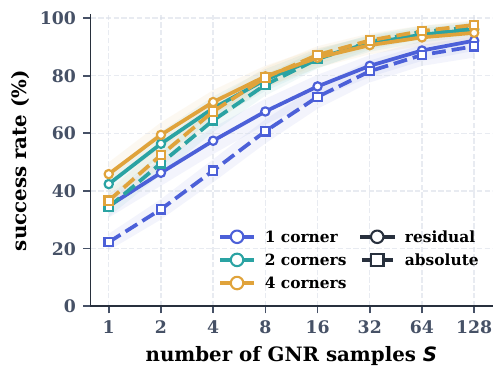}
\caption{Performance of \ABBR{} using either residual or absolute prediction with one, two, or four training quadrants. Shadows show 95\% confidence intervals.}
\label{fig:appendix-spatial-support-residual}
\end{figure}

With one, two, and four training quadrants, residual prediction reaches 34.72\%, 42.35\%, and 45.82\% success at $S{=}1$, and 92.19\%, 96.09\%, and 94.92\% at $S{=}128$ (\Cref{fig:appendix-spatial-support-residual}), showing that the residual prediction is beneficial especially when $S$ is small.

\subsection{Efficiency}
\label{sec:shapefilter-efficiency}
\begin{figure}[H]
\centering
\includegraphics[width=0.7\linewidth]{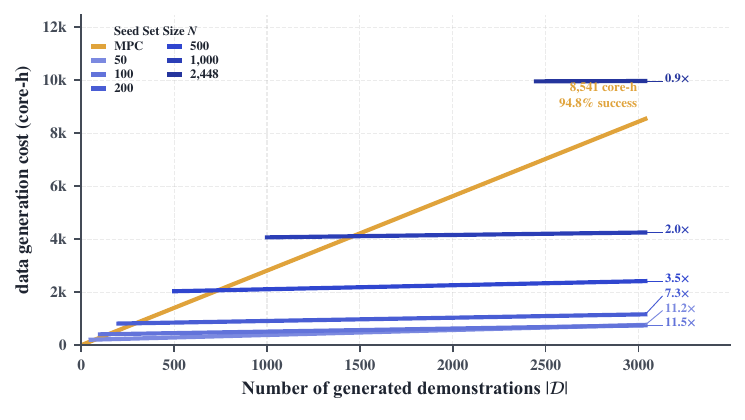} % v2 size: single-panel plots use 0.8\linewidth
\caption{\texttt{ShapeFilter} generation cost versus dataset size. MPC generates all $|\mathcal{D}|$ demonstrations; \ABBR{} uses MPC for $N$ successful seeds and generates the remaining $|\mathcal{D}|-N$ with configurations that meet or exceed the best MPC success rate. Numbers show the reduction in total cost.}
\label{fig:appendix-shapefilter-generation-cost}
\end{figure}

\Cref{tab:corpus-build-cost,tab:appendix-shapefilter-efficiency-details} report success and generation cost. \Cref{fig:appendix-shapefilter-generation-cost} compares total generation cost as dataset size grows; App.~\ref{sec:retargeting_demos_shapefilter} shows qualitative comparisons.
\begin{table}[H]
\centering
\caption{\texttt{ShapeFilter} dataset-generation efficiency. Dataset success rates include the successful seeds and apply the test-set success rate to the remaining demonstrations. Costs include MPC seed generation, \ABBR{} inference, and simulation, in thousands of core-h.}
\label{tab:appendix-shapefilter-efficiency-details}
\scriptsize
\setlength{\tabcolsep}{3.4pt}
\resizebox{\textwidth}{!}{%
\begin{tabular}{@{}rrrrrrrrrrr@{}}
\toprule
Seed set size $N$ & Quantity & \multicolumn{9}{c}{Samples $S$} \\
\cmidrule(lr){3-11}
 & & 1 & 2 & 4 & 8 & 16 & 32 & 64 & 128 & 256 \\
\midrule
50 & Success (\%) & 13.0 & 17.2 & 33.4 & 47.4 & 66.4 & 83.1 & 93.2 & 96.7 & 99.5 \\
% Exact core-h row:  & Cost (core-h) & 235 & 263 & 313 & 390 & 501 & 627 & 740 & 832 & 890 \\
 & Cost (k core-h) & 0.235 & 0.263 & 0.313 & 0.390 & 0.501 & 0.627 & 0.740 & 0.832 & 0.890 \\
\addlinespace[1.5pt]
100 & Success (\%) & 21.0 & 30.5 & 47.5 & 66.7 & 81.3 & 91.1 & 96.4 & 98.7 & 99.3 \\
% Exact core-h row:  & Cost (core-h) & 438 & 463 & 504 & 560 & 626 & 688 & 745 & 789 & 828 \\
 & Cost (k core-h) & 0.438 & 0.463 & 0.504 & 0.560 & 0.626 & 0.688 & 0.745 & 0.789 & 0.828 \\
\addlinespace[1.5pt]
200 & Success (\%) & 33.4 & 44.3 & 57.7 & 71.9 & 80.0 & 89.1 & 94.9 & 96.8 & 99.0 \\
% Exact core-h row:  & Cost (core-h) & 843 & 865 & 898 & 944 & 1,005 & 1,081 & 1,163 & 1,243 & 1,311 \\
 & Cost (k core-h) & 0.843 & 0.865 & 0.898 & 0.944 & 1.005 & 1.081 & 1.163 & 1.243 & 1.311 \\
\addlinespace[1.5pt]
500 & Success (\%) & 48.2 & 58.4 & 68.8 & 77.3 & 84.1 & 89.8 & 94.0 & 96.6 & 97.6 \\
% Exact core-h row:  & Cost (core-h) & 2,061 & 2,077 & 2,102 & 2,137 & 2,186 & 2,251 & 2,330 & 2,417 & 2,526 \\
 & Cost (k core-h) & 2.061 & 2.077 & 2.102 & 2.137 & 2.186 & 2.251 & 2.330 & 2.417 & 2.526 \\
\addlinespace[1.5pt]
1,000 & Success (\%) & 62.2 & 70.1 & 80.0 & 84.9 & 90.9 & 94.4 & 96.8 & 98.2 & 98.7 \\
% Exact core-h row:  & Cost (core-h) & 4,089 & 4,101 & 4,118 & 4,141 & 4,172 & 4,209 & 4,252 & 4,299 & 4,362 \\
 & Cost (k core-h) & 4.089 & 4.101 & 4.118 & 4.141 & 4.172 & 4.209 & 4.252 & 4.299 & 4.362 \\
\addlinespace[1.5pt]
2,448 & Success (\%) & 90.8 & 93.8 & 96.0 & 97.5 & 98.5 & 99.1 & 99.5 & 99.7 & 99.9 \\
% Exact core-h row:  & Cost (core-h) & 9,964 & 9,967 & 9,970 & 9,974 & 9,979 & 9,985 & 9,992 & 10,000 & 10,008 \\
 & Cost (k core-h) & 9.964 & 9.967 & 9.970 & 9.974 & 9.979 & 9.985 & 9.992 & 10.000 & 10.008 \\
\bottomrule
\end{tabular}%
}
\end{table}

\subsection{Trajectory Analysis}
\label{sec:shapefilter-trajectory-analysis}
\begin{figure}[H]
\centering
\includegraphics[width=0.9\textwidth]{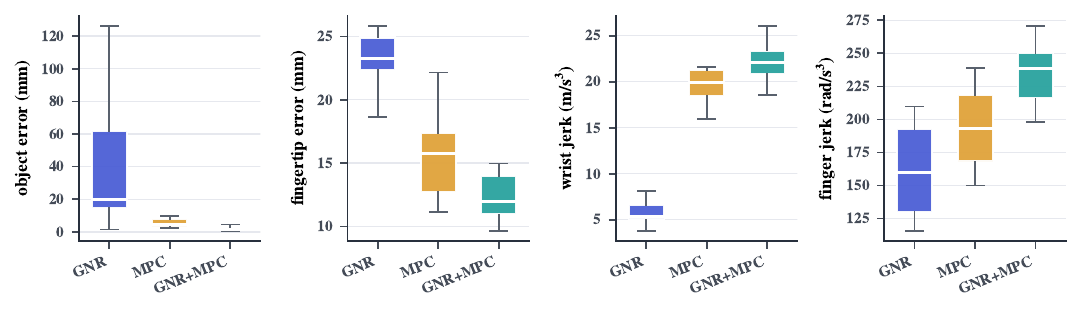}

\caption{\texttt{ShapeFilter} trajectory errors and jerk comparison. \ABBR{} has lower jerk; \ABBR{}+MPC has the lowest object and fingertip tracking errors.}
\label{fig:appendix-shapefilter-trajectory-quality}
\end{figure}

\Cref{fig:appendix-shapefilter-trajectory-quality} compares tracking error and smoothness.

In practice, we find that \ABBR{}$+$MPC data yields the best SRs for policy learning (\Cref{sec:exp:bc_s2r}) due to its improved precision and smoothness at the same time. In this case, \ABBR{} still speeds up BC data generation by reducing the number of MPC samples needed.

For example, \ABBR{}$+$MPC, which uses \ABBR{} trained with $N{=}50$ to initialize MPC with a single iteration ($M{=}1$, 500 samples), reaches over $95\%$ dataset success rate at 1,804 CPU core-h, more than $4\times$ lower than MPC (8,541 core-h). In comparison, \ABBR{} alone reduces data generation cost by $11.2\times$ (\Cref{sec:exp:difficulty}), while its data yields similar BC success to MPC data (\Cref{fig:bc_sr_curves}). This shows a trade-off between data generation cost and data precision. \ABBR{} alone is the most efficient, while \ABBR{}$+$MPC yields higher-precision data for policy learning at a moderately higher cost.

\section{\texttt{NIST}}
\label{sec:appendix_nist}

\subsection{Settings}
\label{sec:nist-settings}
\paragraph{Task.}
In a bimanual Franka/DG-5F system, one arm grasps and screws an M8 knob onto a threaded stud. The 29-D action contains translation, continuous 6-D rotation, and 20 hand joints.

\paragraph{Dataset.}
The dataset contains 358 successful demonstrations and 206,946 frames. At 20 Hz, the mean trajectory duration is 29\,s.

\paragraph{Success criterion.}
Success requires horizontal error below 1 mm, screwing down by at least 8 mm, and axis error below 0.003 rad.

\subsection{Optimization and Efficiency}
\label{sec:nist-cost-accounting}
Optimization has three phases. Wrist alignment uses 86 samples for 10 iterations over seven steps at stride two. Finger closure uses 86 samples for 10 iterations over six steps. The final stage of screwing is the hardest and uses 172 samples for 10 iterations over 13 steps at a stride of 4 and 200\,Hz physics, using approximately 83\% of the compute. \texttt{NIST} MPC sample budgets sum all three stages and have 3,440 samples in total.

\begin{figure}[H]
\centering
\includegraphics[width=0.8\linewidth]{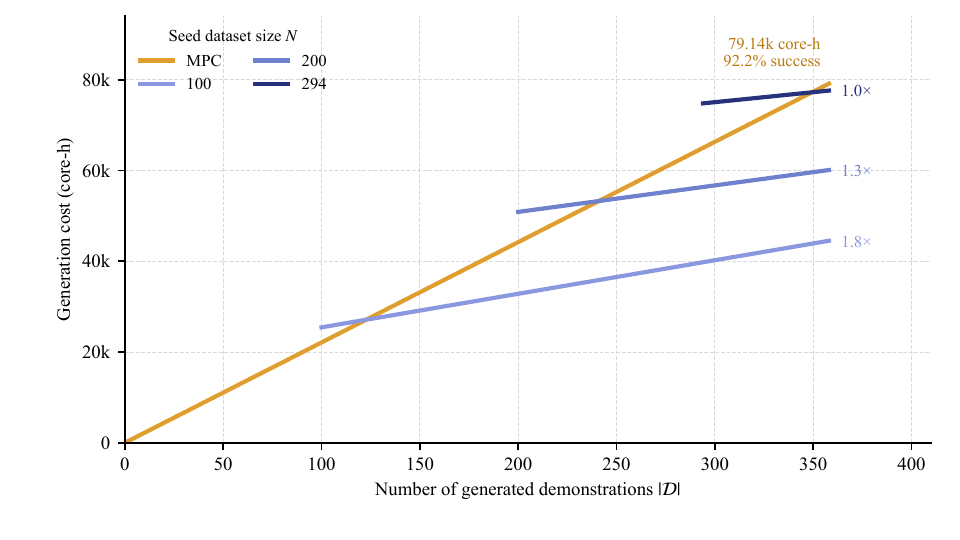} % v2 size: single-panel plots use 0.8\linewidth
\caption{\texttt{NIST} generation cost versus dataset size. MPC generates all $|\mathcal{D}|$ demonstrations; \ABBR{}+MPC uses MPC for $N$ successful seeds and generates the remaining $|\mathcal{D}|-N$ using evaluated configurations with similar dataset success rates. Numbers show the reduction in total cost.}
\label{fig:appendix-nist-generation-cost}
\end{figure}

The three-stage optimization costs 221.05 core-h (6.32 wall-hours) for MPC and 73.94 core-h (about 2.11 wall-hours) for \ABBR{}+MPC in \Cref{tab:difficulty-marginal-cost}, including \ABBR{} inference. \Cref{fig:appendix-nist-generation-cost} compares methods at similar dataset success rates. We show that the dataset generation can be $1.8\times$ faster using \ABBR{}+MPC by using 100 demonstrations as the MPC seed.

% \begin{table}[H]
% \centering
% \caption{\texttt{NIST} dataset-generation efficiency. Dataset success rates include the successful seeds and estimated successes on the remaining demonstrations. Each MPC budget uses the schedule with the highest success rate. \ABBR{} generation costs include MPC seed generation, \ABBR{} inference, and \ABBR{}-initialized MPC, in thousands of CPU core-h (App.~\ref{sec:nist-cost-accounting}).}
% \label{tab:appendix-nist-efficiency-details}
% \scriptsize
% \setlength{\tabcolsep}{4.0pt}
% \resizebox{\textwidth}{!}{%
% \begin{tabular}{@{}rrr@{\hspace{12pt}}rrr@{\hspace{12pt}}rrr@{}}
% \toprule
% \multicolumn{3}{c}{$N{=}100$} & \multicolumn{3}{c}{$N{=}200$} & \multicolumn{3}{c}{$N{=}294$} \\
% \cmidrule(lr){1-3}\cmidrule(lr){4-6}\cmidrule(lr){7-9}
% MPC samples & Success (\%) & Cost (k core-h) & MPC samples & Success (\%) & Cost (k core-h) & MPC samples & Success (\%) & Cost (k core-h) \\
% \midrule
% 1,736 & 64.0 & 35.90 & 1,736 & 79.3 & 57.34 & 1,736 & 91.9 & 77.41 \\
% 1,752 & 70.7 & 36.79 & 1,752 & 83.4 & 57.89 & 1,752 & 93.3 & 77.61 \\
% 1,768 & 74.1 & 37.42 & 1,784 & 89.0 & 58.91 & 1,784 & 95.5 & 77.97 \\
% 1,784 & 79.7 & 38.17 & 1,816 & 90.3 & 59.08 & 1,816 & 97.2 & 78.37 \\
% 1,816 & 80.9 & 39.81 & 1,848 & 93.8 & 60.10 & 1,912 & 97.8 & 78.95 \\
% 1,848 & 85.4 & 41.02 & 2,104 & 95.9 & 64.46 & -- & -- & -- \\
% 1,912 & 87.6 & 42.51 & -- & -- & -- & -- & -- & -- \\
% 1,976 & 92.1 & 44.52 & -- & -- & -- & -- & -- & -- \\
% 2,104 & 93.2 & 45.91 & -- & -- & -- & -- & -- & -- \\
% \bottomrule
% \end{tabular}%
% }
% \end{table}

% v2 restructure: renamed from "Efficiency" to distinguish it from the per-benchmark Efficiency subsections
\section{Dataset Generation Efficiency}
\label{sec:efficiency-accounting}

\subsection{Compute Cost}

We report both wall time and CPU compute time (core-h).

\begin{table}[H]
\centering
\caption{Measured per-episode generation cost.}
\label{tab:efficiency-per-episode}
\footnotesize % v2 size: appendix tables use footnotesize (<=5 columns) or scriptsize (wider)
\setlength{\tabcolsep}{5pt}
\begin{tabular}{@{}llrrr@{}}
\toprule
Task & Method & Sample budget & Core-h / episode & Wall h / episode \\
\midrule
\texttt{HOT3D} & DIAL-MPC & 3,000 & 1.0777 & 0.2427 \\
\texttt{HOT3D} & iCEM & 3,000 & 1.4213 & 0.3282 \\
\texttt{HOT3D} & CMA-ES & 3,000 & 1.2272 & 0.2877 \\
\texttt{HOT3D} & \ABBR{}, $S{=}256$ & 256 & 0.1094 & 0.0383 \\
\midrule
\texttt{ShapeFilter} & MPC & 3,000 & 3.2001 & 0.0472 \\
\bottomrule
\end{tabular}
\end{table}

\paragraph{GPU training time.}

On one H100 GPU, \texttt{ShapeFilter} training takes 0.0108 to 0.5022 GPU-h over 100 epochs as the seed set grows from 50 demonstrations to the full training set. \texttt{NIST} training takes 0.0202 to 0.0406 GPU-h over 200 epochs with 100 to 294 demonstrations. The training cost is overall marginal compared to physics-based sampling.

\subsection{Cost versus Dataset Coverage}

\Cref{tab:corpus-build-cost} sums CPU generation costs for MPC seed generation and \ABBR{} generation. \texttt{HOT3D} compares full MPC generation, dataset expansion, and generation from 28k MPC seeds at $S{=}256$; \texttt{ShapeFilter} and \texttt{NIST} compare generation cost and success. \texttt{NIST} costs follow App.~\ref{sec:nist-cost-accounting}.

\begin{table}[H]
\centering
\caption{Dataset generation cost (thousands of core-h) and coverage.}
\label{tab:corpus-build-cost}
\scriptsize
\setlength{\tabcolsep}{4.2pt}
\begin{tabular}{@{}llrrrr@{}}
\toprule
Task & Generation & \multicolumn{3}{c}{Generation compute ($10^3$ core-h)} & Coverage \\
\cmidrule(lr){3-5}
& & MPC data & \ABBR{} & Total & \\
\midrule
\texttt{HOT3D} & all MPC & 390.043 & -- & 390.043 & 26.47\% \\
\texttt{HOT3D} & all MPC data $+$ \ABBR{} $S{=}256$ & 390.043 & 26.047 & 416.090 & 61.77\% \\
\texttt{HOT3D} & 28k MPC seeds $+$ \ABBR{} $S{=}256$ & 115.939 & 32.623 & 148.562 & 59.71\% \\
\texttt{ShapeFilter} & all MPC & 8.541 & -- & 8.541 & 93.70\% \\
\texttt{ShapeFilter} & 50 MPC seeds $+$ \ABBR{} & 0.203 & 0.562 & 0.766 & 94.82\% \\
\texttt{NIST} & all MPC & 79.136 & -- & 79.136 & 92.19\% \\
\texttt{NIST} & 100 MPC seeds $+$ \ABBR{}$+$MPC & 25.439 & 19.078 & 44.517 & 92.12\% \\
\bottomrule
\end{tabular}
\end{table}

For \texttt{HOT3D}, the full MPC generation run costs 390k core-h. Applying \ABBR{} on the evaluated MPC failures expands the dataset to 223k trajectories (App.~\ref{sec:hot3d-dataset-expansion}). Together, the methods cover 61.77\% of the evaluation cohort. Using 354.1 core-s per demonstration at $S{=}256$ (\Cref{tab:baselines}), the additional cost is about $264{,}806\times354.1/3600\approx26{,}047$ core-h, for a total of 416.1k core-h. The dataset grows by $2.33\times$ for $6.68\%$ additional compute, reducing CPU cost per retained trajectory from 4.08 to 1.87 core-h ($2.18\times$).

A more practical way to deploy \ABBR{} is to use a smaller seed dataset to reduce the MPC generation cost. Obtaining 28k MPC successes requires about 108k attempts at the measured MPC success rate of 26.47\%. At 1.0777 core-h per demonstration (\Cref{tab:efficiency-per-episode}), seed generation costs about 115.9k core-h. Applying \ABBR{} to the remaining 331k demonstrations at 354.1 core-s each costs about 32.6k core-h, for a total of 148.6k core-h. Including the seeds, this strategy yields about 215k trajectories (59.71\% coverage), at $2.63\times$ lower CPU cost than the full MPC run.

For \texttt{ShapeFilter}, using 50 seed demonstrations reduces generation cost from 8.54k to 766 core-h while increasing dataset success from 93.7\% to 94.8\%. For \texttt{NIST}, using 100 seed demonstrations reduces generation cost from 79.14k to 44.52k core-h at about 92.2\% success. See \Cref{tab:appendix-shapefilter-efficiency-details} and \Cref{fig:appendix-nist-generation-cost} for the cost comparisons.

\section{Multi-Embodiment Evaluation}
\label{sec:multi-embodiment}

We evaluate whether a \ABBR{} model can learn from demonstrations across multiple dexterous robot hands and retarget trajectories across hands.

\new{We train on demonstrations released by SPIDER~\citep{spider} and evaluate retargeting performance using the GPU-accelerated MuJoCo Warp simulator~\citep{zakka2026mjlab} for both SPIDER and \ABBR{}.}

\subsection{Setup and Results}

\new{\paragraph{GPU Evaluation.} The experiments in the main paper use SuperDex~\citep{superdex2026}, a state-of-the-art simulator with high-precision physics and support for non-convex collisions. SuperDex is CPU-bound due to its precision requirements. Here, we use MuJoCo Warp to demonstrate that \ABBR{} is agnostic to the choice of compute.}

\paragraph{Training \& Evaluation.} We compare against the official SPIDER implementation~\citep{spider} on its dataset. We train one \ABBR{} model on 147 trajectories spanning five hands and evaluate it on 100 trajectories, with 20 per hand. We use a unified 52-dimensional representation for actions across embodiments. 

\begin{figure}[H]
\centering
\includegraphics[width=0.8\linewidth]{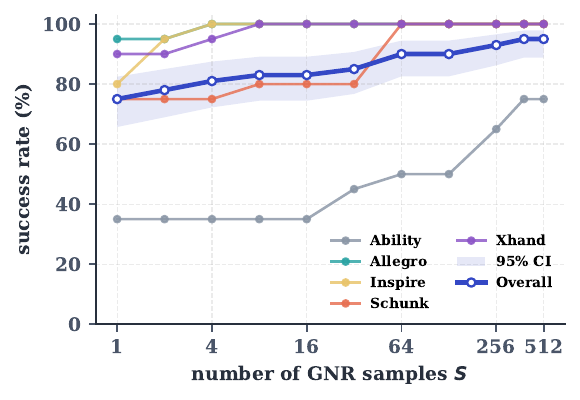} % v2 size: single-panel plots use 0.8\linewidth
\caption{\ABBR{} test-time scaling across five different dexterous hands.}
\label{fig:multiembodiment-scaling}
\end{figure}

\subsection{Efficiency Comparison}
\label{sec:spider-sample-accounting}

\new{\Cref{tab:spider-vs-gnr} shows that \ABBR{} reduces the sample budget while achieving comparable SRs to SPIDER in a GPU-accelerated simulator.}

% TODO: add the speedup of gpu time here

\begin{table}[H]
\centering
\footnotesize
\setlength{\tabcolsep}{5pt}
\caption{Multi-embodiment success on 100 SPIDER test trajectories.}
\label{tab:spider-vs-gnr}
\label{tab:multiembodiment-results}
\label{tab:multiembodiment-budget}
\begin{tabular}{@{}lrr@{}}
\toprule
Method & Samples & SR (\%) \\
\midrule
\ABBR{} & 128 & 90.0 \\
\ABBR{} & 256 & 93.0 \\
\ABBR{} & 384 & 95.0 \\
SPIDER & 16,384--32,768 & 96.0 \\
\bottomrule
\end{tabular}
\end{table}

\new{We further break down SR and ADD-AUC10 by hand embodiment in~\Cref{tab:spider_per_hand}, comparing \ABBR{} with SPIDER. Averaged over the five hands, \ABBR{} achieves $95\%$ SR and $61.4$ ADD-AUC10, compared with $96\%$ and $62.6$ for SPIDER, while using far fewer samples.}

\begin{table}[t]
\centering
\caption{Success rates and ADD-AUC10~\citep{spider} of each hand on the SPIDER benchmark~\citep{spider}.}
\label{tab:spider_per_hand}
\small
\begin{tabular}{@{}lrrrr@{}}
\toprule
& \multicolumn{2}{c}{SR (\%)} & \multicolumn{2}{c}{ADD-AUC10} \\
\cmidrule(lr){2-3}\cmidrule(lr){4-5}
Hand & \ABBR{} & SPIDER & \ABBR{} & SPIDER \\
\midrule
Inspire & 100 & 95 & 67.5 & 58.0 \\
Schunk  & 100 & 95 & 61.9 & 63.9 \\
XHand   & 100 & 100 & 63.2 & 66.4 \\
Ability & 75 & 95 & 55.9 & 62.1 \\
Allegro & 100 & 95 & 58.6 & 62.7 \\
\midrule
Mean    & 95 & 96 & 61.4 & 62.6 \\
\bottomrule
\end{tabular}
\end{table}

\section{Learning from Retargeted Demonstrations}
\label{sec:generated-data}

In this section, we study learning robot policies or driving a data flywheel using \ABBR{}-generated demonstrations.

\new{

\subsection{Behavioral Cloning}
\label{app:more-bc}

\Cref{sec:exp:bc_s2r} demonstrates that \ABBR{} can be applied as the data generator for BC policy learning. \ABBR{} speeds up dataset generation and even improves data quality by combining \ABBR{} with closed-loop MPC on a challenging millimeter-precision task \texttt{ShapeFilter}. The learned visuomotor policy uses images from four RGB cameras and achieves sim-to-real transfer without real-robot demonstrations. Now, we include more BC results and discussions.

\subsubsection{Contact Force for Behavioral Cloning}

Visuomotor control with force and tactile information has emerged as a promising direction for BC on contact-rich tasks. When and how force information helps BC remains an open question~\citep{liu2025factr}. Its usefulness depends on the task~\citep{chen2025dexforce} and on whether the policy learns to use force when visual signals are easier to learn from~\citep{liu2025factr}. Existing approaches incorporate contact information in several ways, including predicting future tactile signals as a training objective~\citep{zhou2026agile} and using force measurements as policy observations~\citep{chen2025dexforce,xue2025reactive}.

Although learning with force is not the primary focus of this paper, we test whether simulated contact-force labels can be used for BC. We train visuomotor diffusion policies (DPs) on 2k demonstrations and compare the DP without using force information to DPs using force either as a supervision objective or additional observations. As shown in~\Cref{tab:shapefilter_force_bc}, adding force supervision increases the observed closed-loop SR from $27.4\%$ to $31.1\%$, while using force as observations yields similar performance to the DP without force.

\begin{table}[H]
\centering
\caption{We report the mean and standard deviation across three seeds.}
\label{tab:shapefilter_force_bc}
\small
\begin{tabular}{@{}lr@{}}
\toprule
Policy & SR (\%) \\
\midrule
DP w/o force & 27.4 $\pm$ 1.9 \\
DP + force observations & 26.2 $\pm$ 2.2 \\
DP + auxiliary force prediction & \textbf{31.1} $\pm$ 2.0 \\
\bottomrule
\end{tabular}
\end{table}

% \subsubsection{Expanding MPC-Generated BC Data Using \ABBR{}}

% On \texttt{ShapeFilter}, adding 2k \ABBR{} demonstrations to a seed set of 131 MPC demonstrations improves closed-loop SR from $11.6\%$ to $26.2\%$. Adding 2,500 \ABBR{} demonstrations to a seed set of 339 MPC demonstrations improves SR from $16.1\%$ to $27.4\%$. In a separate experiment with 500 MPC seed demonstrations and the number of training epochs held fixed, adding 2,499 \ABBR{} demonstrations improves SR from $23.3\%$ to $28.2\%$. We evaluate on 589 references excluded from policy training.

\new{
\subsubsection{Behavioral Cloning on \texttt{HOT3D}}

\ABBR{} expands the dataset by retargeting demonstrations from the MPC-failure set, broadening coverage across geometries and motions. We test BC on \texttt{HOT3D} to see whether this expanded dataset can facilitate policy learning.

We randomly sample 5k demonstrations from the MPC-success set as the initial training data, and 3.7k demonstrations as the test set, which includes both MPC-failure and MPC-success demonstrations. The MPC-failure test set covers 3,060 unseen episodes from 269 source motions. The MPC-success test set covers 659 episodes across 223 motions. We then augment the training dataset with 5k or 15k new MPC or \ABBR{} demonstrations. With 5k additional demonstrations, MPC still adds data from the MPC-success set, while \ABBR{} is able to expand the dataset with new geometries and new motions from the MPC-failure set, thus improving the diversity and coverage of the training data. \Cref{tab:hot3d_bc_seed_expansion} shows that \ABBR{} achieves higher SR on both MPC-failure episodes (36.1\% vs.\ 33.1\%) and MPC-success episodes (53.5\% vs.\ 51.1\%) due to the expanded coverage. With 15k additional demonstrations, \ABBR{} further improves SR on MPC-failure episodes to 36.8\%, compared with 32.0\% for MPC augmentation.
}

\begin{table}[H]
\centering
\caption{Closed-loop success rates for DP on \texttt{HOT3D}. All training datasets include the same 5k MPC seed demonstrations.}
\label{tab:hot3d_bc_seed_expansion}
\small
\begin{tabular}{@{}lrrr@{}}
\toprule
Training data & Total demos & MPC-failure SR (\%) & MPC-success SR (\%) \\
\midrule
5k MPC (seed) & 5k & 33.8 & 50.5 \\
\midrule
$+$5k MPC & 10k & 33.1 & 51.1 \\
$+$5k \ABBR{} & 10k & \textbf{36.1} & \textbf{53.5} \\
\midrule
$+$15k MPC & 20k & 32.0 & 52.8 \\
$+$15k \ABBR{} & 20k & \textbf{36.8} & \textbf{53.7} \\
\bottomrule
\end{tabular}
\end{table}

}

\subsection{Data Flywheel}
\label{sec:data_flywheels}

Newly retargeted demonstrations can support subsequent rounds of \ABBR{} training, forming a data flywheel~\citep{dextrack,zhu2026dexflywheel}. We test this on \texttt{ShapeFilter}, starting with 200 MPC-generated seed trajectories. Each round uses MPC to solve demonstrations on which \ABBR{} fails and adds the successful trajectories to the training set. Each of the four rounds adds 220, 104, 120, and 105 trajectories, respectively, yielding 749 training trajectories. On the same 256-episode test set, success increases from $55.2\%$ to $74.4\%$ at $S=8$ and from $68.4\%$ to $84.8\%$ at $S=16$.

\begin{figure}[H]
\centering
\includegraphics[width=0.5\linewidth]{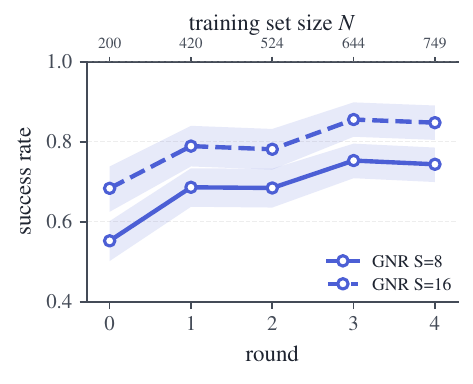} % v2 size: single-panel plots use 0.8\linewidth
\caption{\texttt{ShapeFilter} success across four data-flywheel rounds.}
\label{fig:appendix_flywheel}
\end{figure}

\clearpage
\section{Retargeting Demos}
\label{sec:retargeting_demos}

This section presents trajectories retargeted by different methods on \texttt{ShapeFilter}, \texttt{NIST}, and \texttt{HOT3D}.

\subsection{\texttt{ShapeFilter}}
\label{sec:retargeting_demos_shapefilter}

\begin{figure}[H]
\centering
\resizebox{\linewidth}{!}{%
\begin{tabular}{@{}r@{\hspace{5pt}}c@{}}
IK & \includegraphics[width=0.89\textwidth]{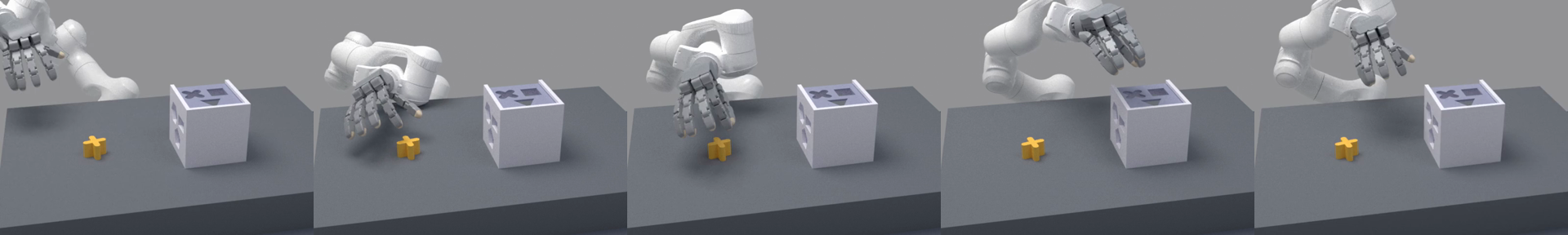} \\
MPC & \includegraphics[width=0.89\textwidth]{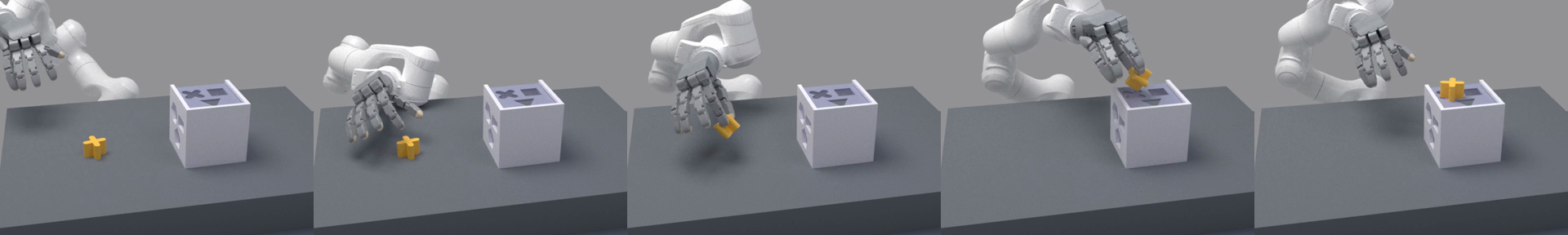} \\
\ABBR{} & \includegraphics[width=0.89\textwidth]{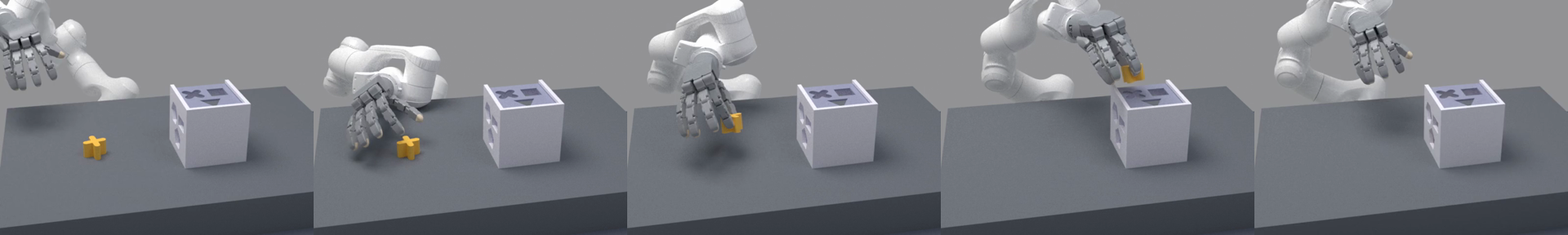} \\
\ABBR{}+MPC & \includegraphics[width=0.89\textwidth]{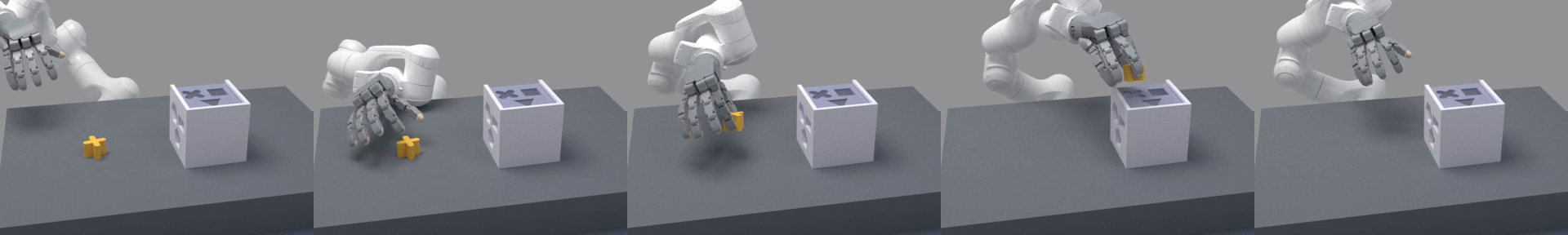}
\end{tabular}}
\caption{\texttt{ShapeFilter} example in which MPC fails while \ABBR{} and \ABBR{}+MPC succeed.}
\label{fig:appendix-shapefilter-demo}
\end{figure}

\begin{figure}[H]
\centering
\resizebox{\linewidth}{!}{%
\begin{tabular}{@{}r@{\hspace{5pt}}c@{}}
IK & \includegraphics[width=0.89\textwidth]{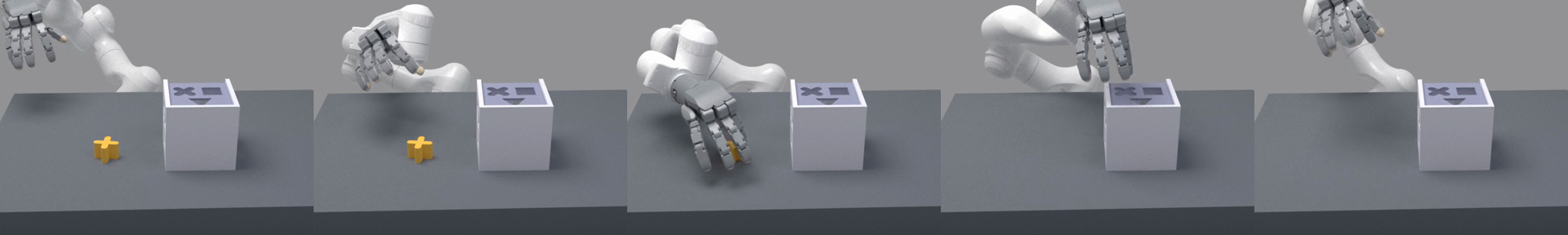} \\
MPC & \includegraphics[width=0.89\textwidth]{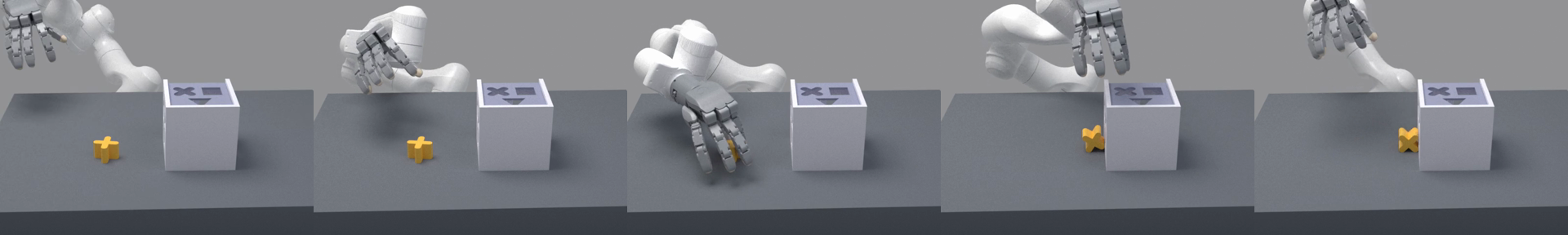} \\
\ABBR{} & \includegraphics[width=0.89\textwidth]{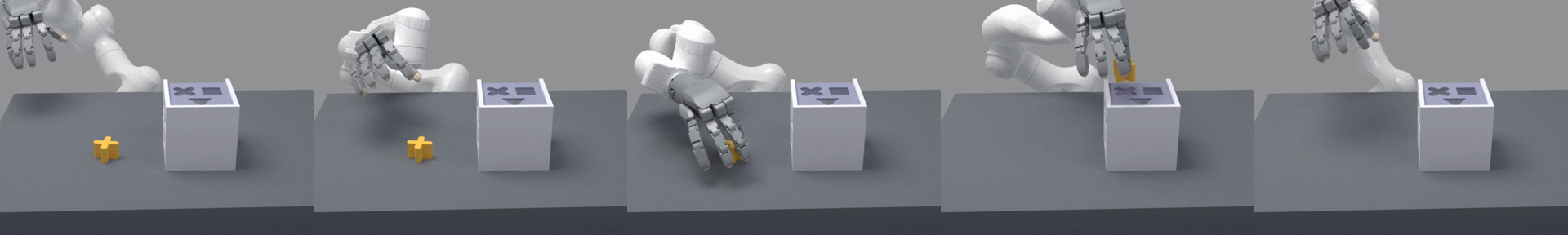} \\
\ABBR{}+MPC & \includegraphics[width=0.89\textwidth]{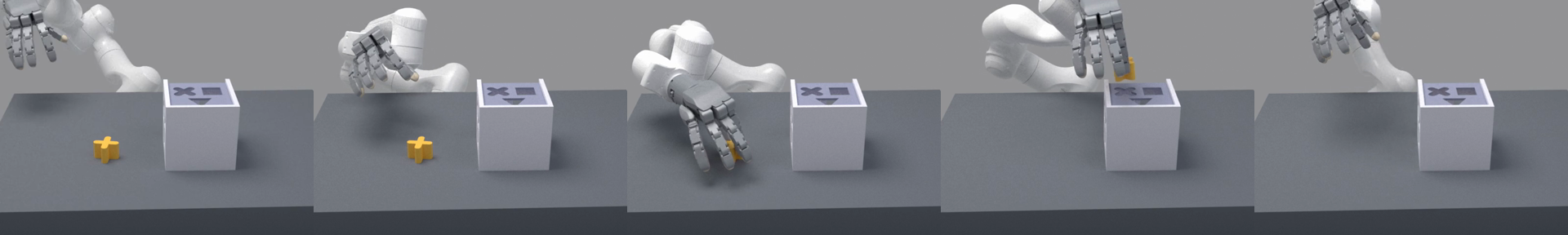}
\end{tabular}}
\caption{\texttt{ShapeFilter} example in which MPC fails while \ABBR{} and \ABBR{}+MPC succeed.}
\label{fig:appendix-shapefilter-demo-second}
\end{figure}

\clearpage
\subsection{\texttt{NIST}}
\label{sec:retargeting_demos_nist}

\begin{figure}[H]
\centering
\resizebox{\linewidth}{!}{%
\begin{tabular}{@{}r@{\hspace{5pt}}c@{}}
Human & \includegraphics[width=0.89\textwidth]{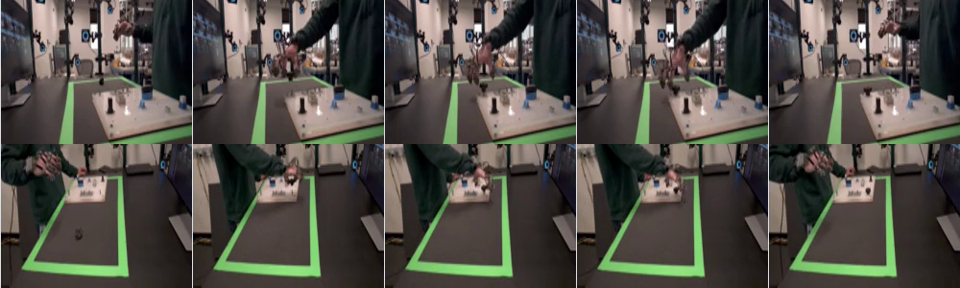} \\
IK & \includegraphics[width=0.89\textwidth]{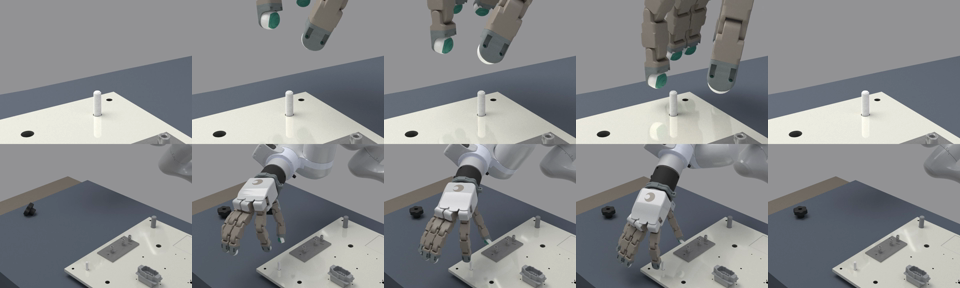} \\
MPC & \includegraphics[width=0.89\textwidth]{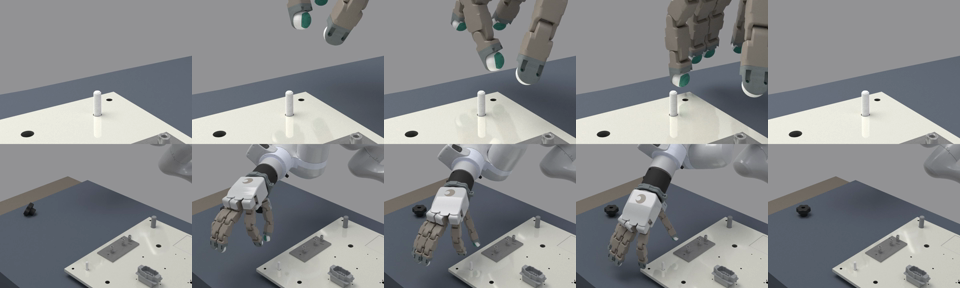} \\
\ABBR{} & \includegraphics[width=0.89\textwidth]{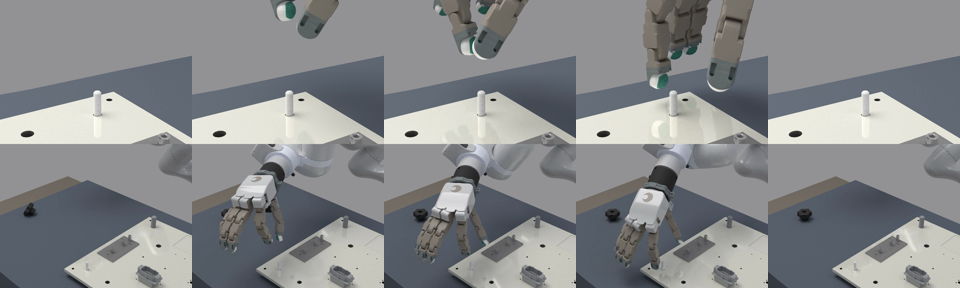} \\
\ABBR{}+MPC & \includegraphics[width=0.89\textwidth]{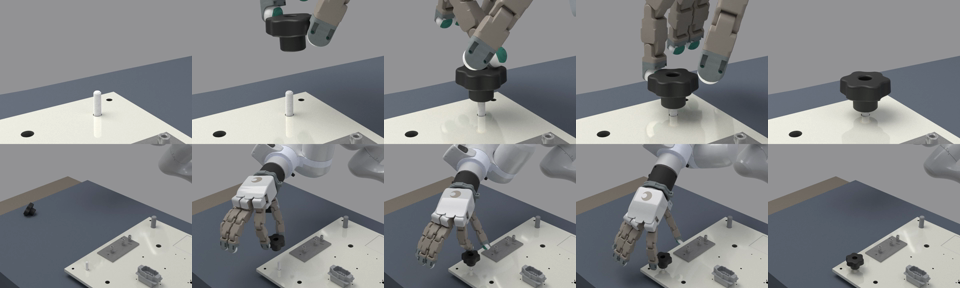}
\end{tabular}}
\caption{A \texttt{NIST} example. \ABBR{}+MPC succeeds where IK, MPC, and open-loop \ABBR{} fail.}
\label{fig:appendix-nist-demo}
\end{figure}

\begin{figure}[H]
\centering
\resizebox{\linewidth}{!}{%
\begin{tabular}{@{}r@{\hspace{5pt}}c@{}}
Human & \includegraphics[width=0.89\textwidth]{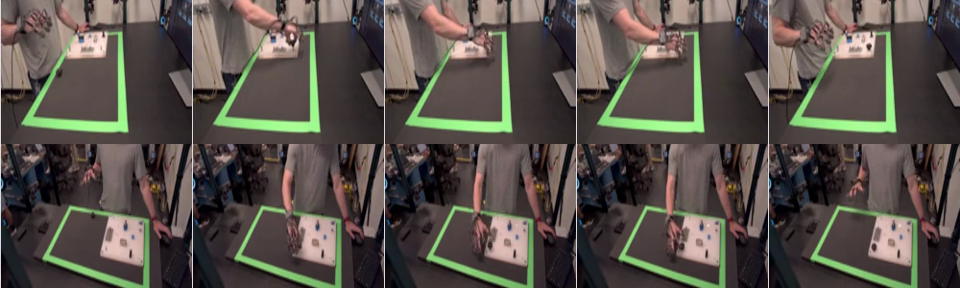} \\
IK & \includegraphics[width=0.89\textwidth]{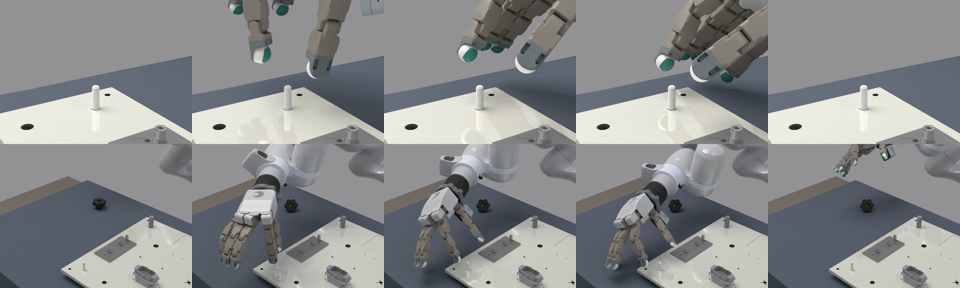} \\
MPC & \includegraphics[width=0.89\textwidth]{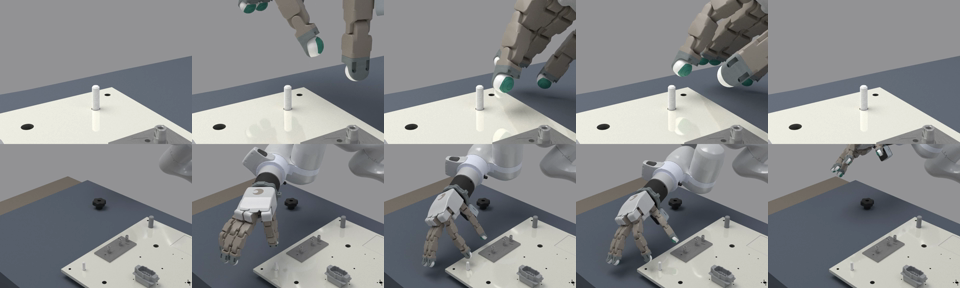} \\
\ABBR{} & \includegraphics[width=0.89\textwidth]{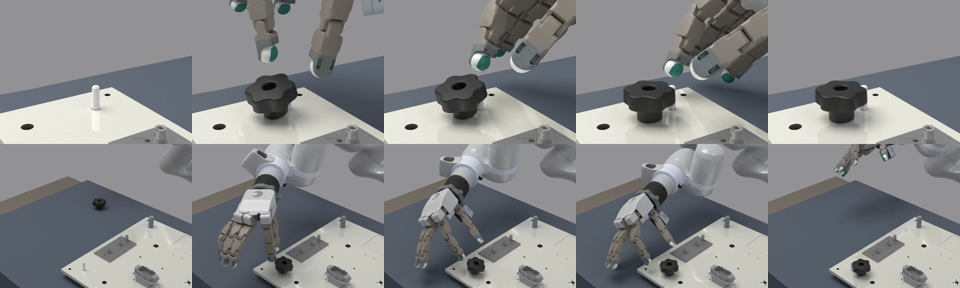} \\
\ABBR{}+MPC & \includegraphics[width=0.89\textwidth]{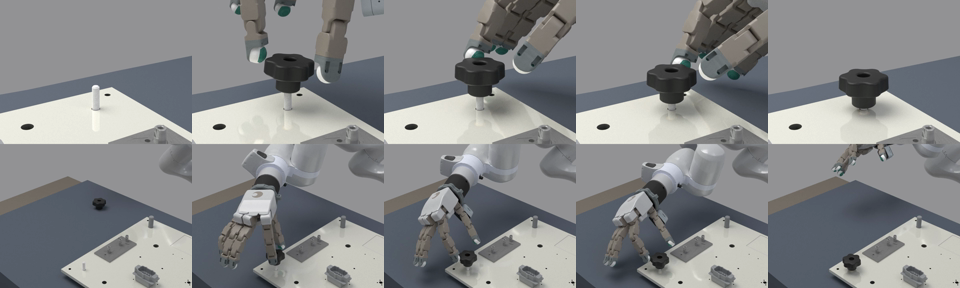}
\end{tabular}}
\caption{A \texttt{NIST} example. \ABBR{}+MPC succeeds where IK, MPC, and open-loop \ABBR{} fail.}
\label{fig:appendix-nist-demo-second}
\end{figure}

\clearpage
\subsection{\texttt{HOT3D}}
\label{sec:retargeting_demos_hot3d}

We present egocentric recordings and trajectories generated by MPC and \ABBR{} across different objects.

\begin{figure}[H]
\centering
\includegraphics[width=\linewidth]{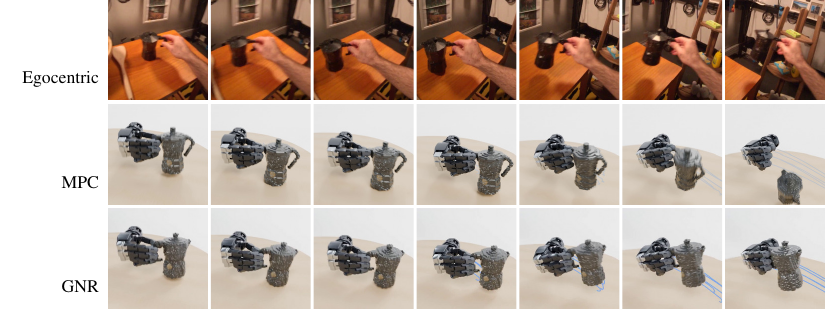}
\caption{\texttt{HOT3D} coffee pot.}
\label{fig:appendix-hot3d-coffee-pot}
\end{figure}

\begin{figure}[H]
\centering
\includegraphics[width=\linewidth]{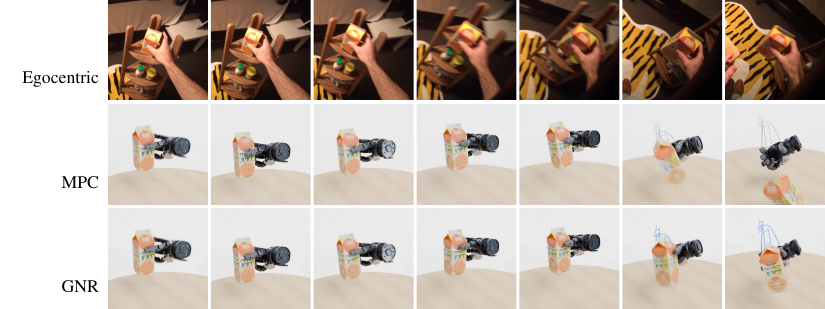}
\caption{\texttt{HOT3D} juice carton.}
\label{fig:appendix-hot3d-carton-oj}
\end{figure}

\begin{figure}[H]
\centering
\includegraphics[width=\linewidth]{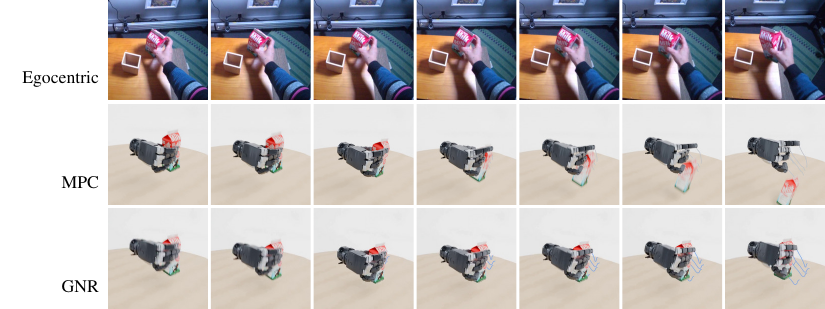}
\caption{\texttt{HOT3D} milk carton.}
\label{fig:appendix-hot3d-carton-milk}
\end{figure}

\begin{figure}[H]
\centering
\includegraphics[width=\linewidth]{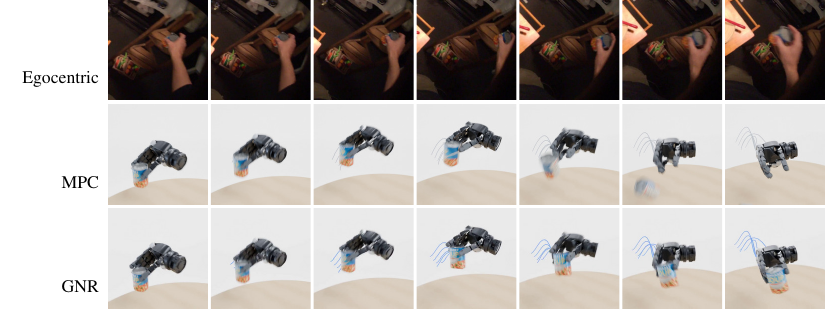}
\caption{\texttt{HOT3D} soup can.}
\label{fig:appendix-hot3d-can-soup}
\end{figure}

\begin{figure}[H]
\centering
\includegraphics[width=\linewidth]{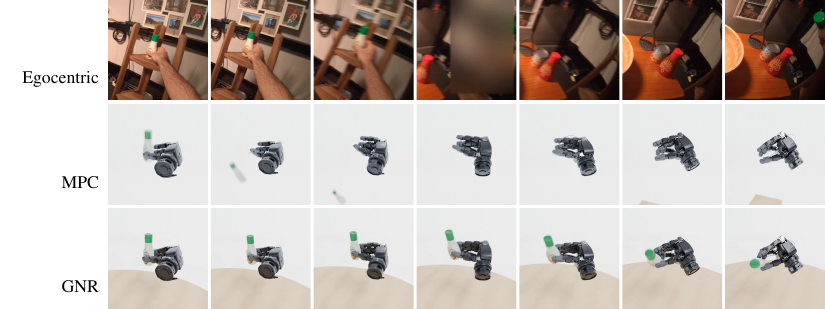}
\caption{\texttt{HOT3D} ranch bottle.}
\label{fig:appendix-hot3d-bottle-ranch}
\end{figure}

\begin{figure}[H]
\centering
\includegraphics[width=\linewidth]{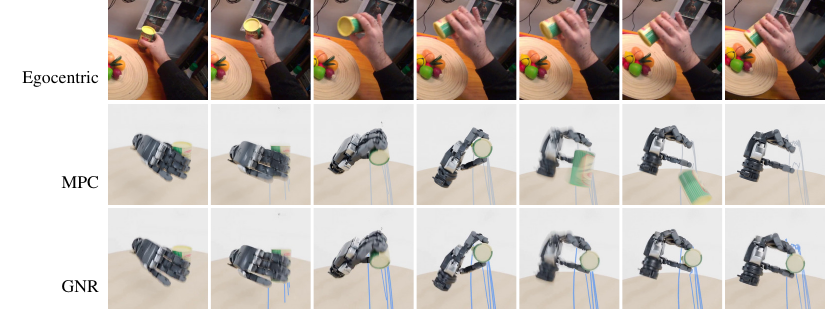}
\caption{\texttt{HOT3D} parmesan can.}
\label{fig:appendix-hot3d-can-parmesan}
\end{figure}

\begin{figure}[H]
\centering
\includegraphics[width=\linewidth]{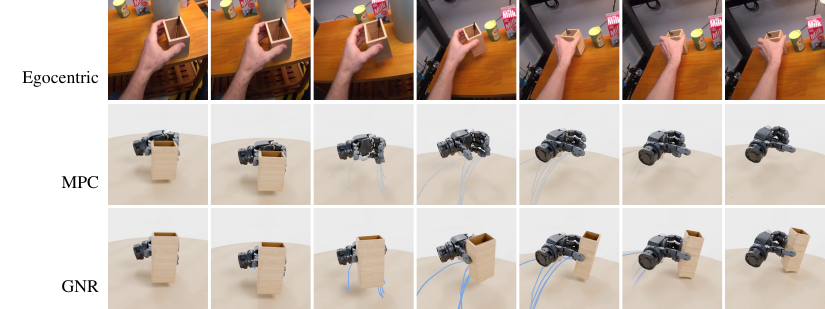}
\caption{\texttt{HOT3D} wooden pen holder.}
\label{fig:appendix-hot3d-holder-gray}
\end{figure}

\begin{figure}[H]
\centering
\includegraphics[width=\linewidth]{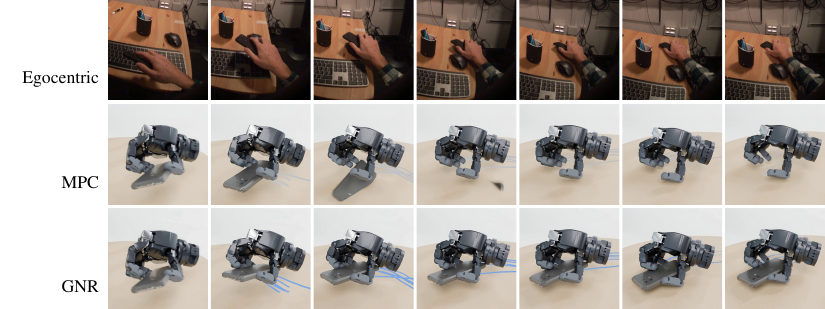}
\caption{\texttt{HOT3D} phone.}
\label{fig:appendix-hot3d-phone}
\end{figure}

\begin{figure}[H]
\centering
\includegraphics[width=\linewidth]{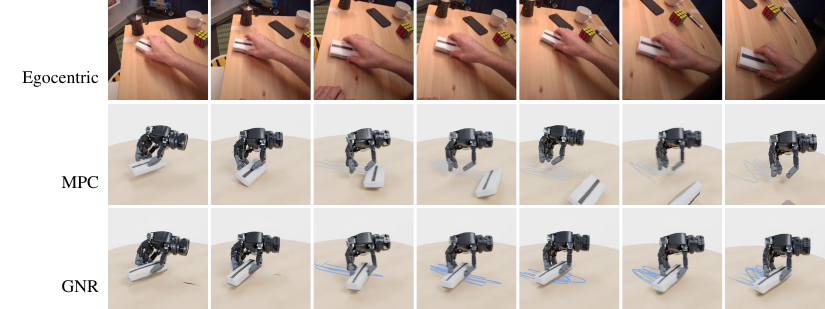}
\caption{\texttt{HOT3D} whiteboard eraser.}
\label{fig:appendix-hot3d-eraser}
\end{figure}

\begin{figure}[H]
\centering
\includegraphics[width=\linewidth]{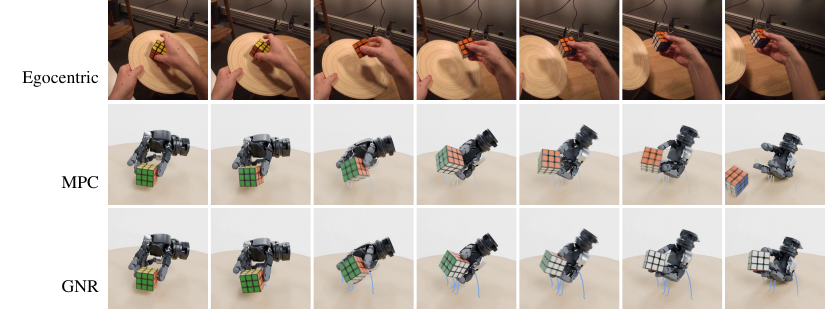}
\caption{\texttt{HOT3D} puzzle cube.}
\label{fig:appendix-hot3d-puzzle-cube}
\end{figure}

\paragraph{Human motion optimization.}

Aside from optimizing robot trajectories, we also generated 37,468 successful MPC-optimized human-motion demonstrations. The examples below show MPC making reconstructed human motion physically feasible.

\begin{figure}[H]
\centering
\resizebox{\linewidth}{!}{%
\begin{tabular}{@{}r@{\hspace{5pt}}c@{}}
Egocentric & \includegraphics[width=0.89\textwidth]{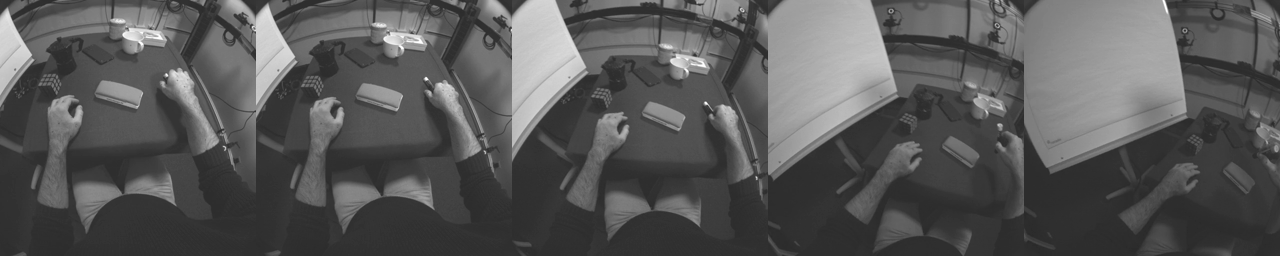} \\
\small real-to-sim & \includegraphics[width=0.89\textwidth]{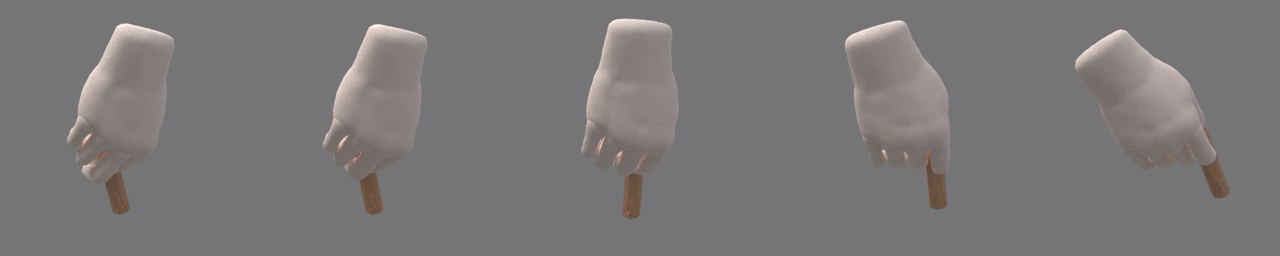} \\
\small real-to-sim + MPC & \includegraphics[width=0.89\textwidth]{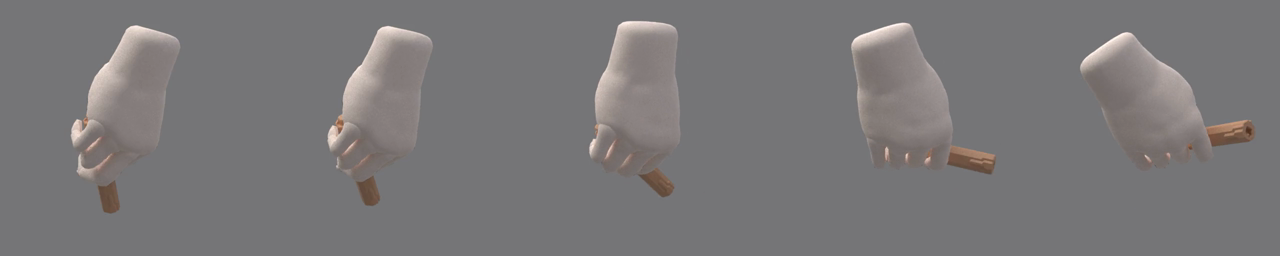}
\end{tabular}}
\caption{\texttt{HOT3D} whiteboard marker. MPC makes the reconstructed human motion physically feasible.}
\label{fig:appendix-hot3d-real2sim-marker}
\end{figure}

\begin{figure}[H]
\centering
\resizebox{\linewidth}{!}{%
\begin{tabular}{@{}r@{\hspace{5pt}}c@{}}
Egocentric & \includegraphics[width=0.89\textwidth]{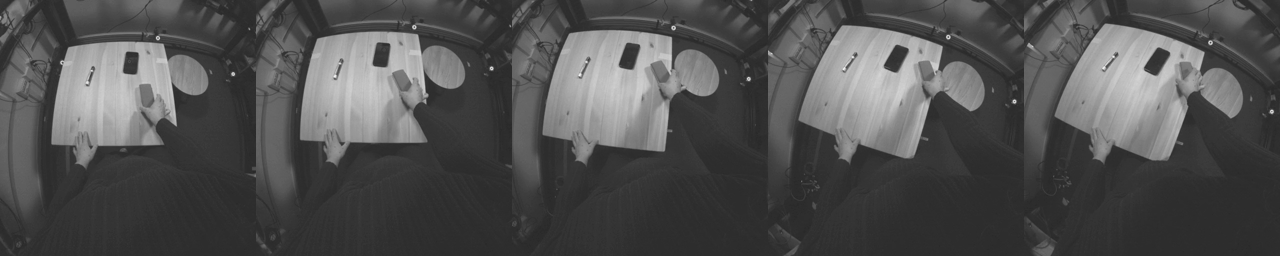} \\
\small real-to-sim & \includegraphics[width=0.89\textwidth]{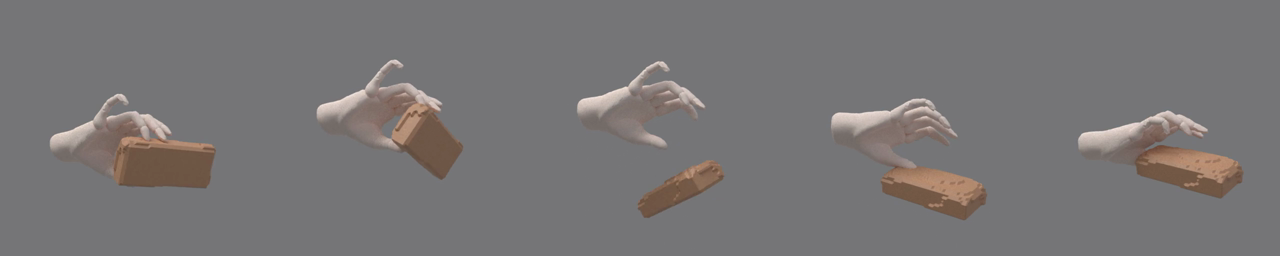} \\
\small real-to-sim + MPC & \includegraphics[width=0.89\textwidth]{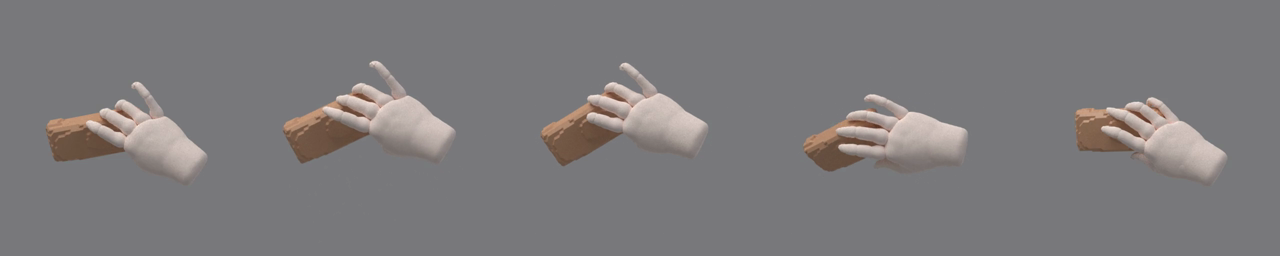}
\end{tabular}}
\caption{\texttt{HOT3D} whiteboard eraser. MPC makes the reconstructed human motion physically feasible.}
\label{fig:appendix-hot3d-real2sim-eraser}
\end{figure}

\begin{figure}[H]
\centering
\resizebox{\linewidth}{!}{%
\begin{tabular}{@{}r@{\hspace{5pt}}c@{}}
Egocentric & \includegraphics[width=0.89\textwidth]{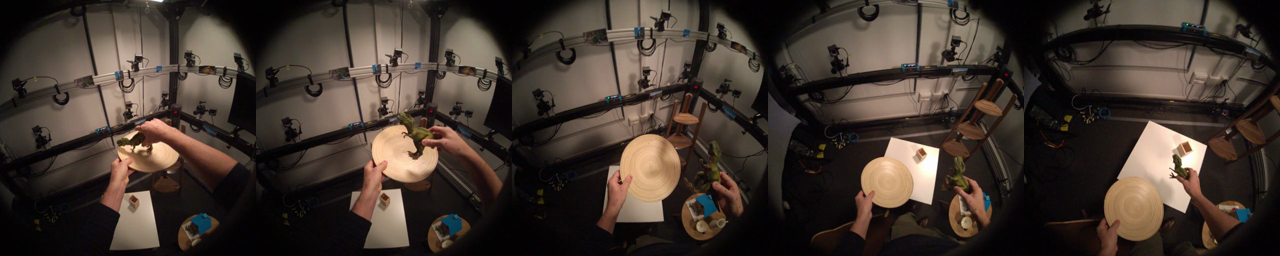} \\
\small real-to-sim & \includegraphics[width=0.89\textwidth]{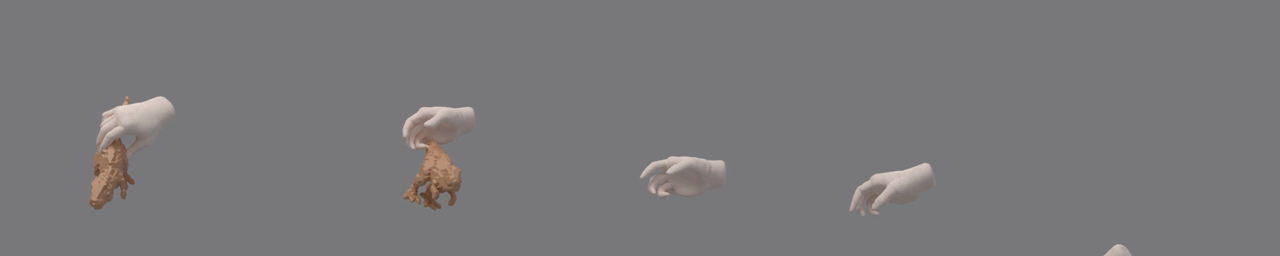} \\
\small real-to-sim + MPC & \includegraphics[width=0.89\textwidth]{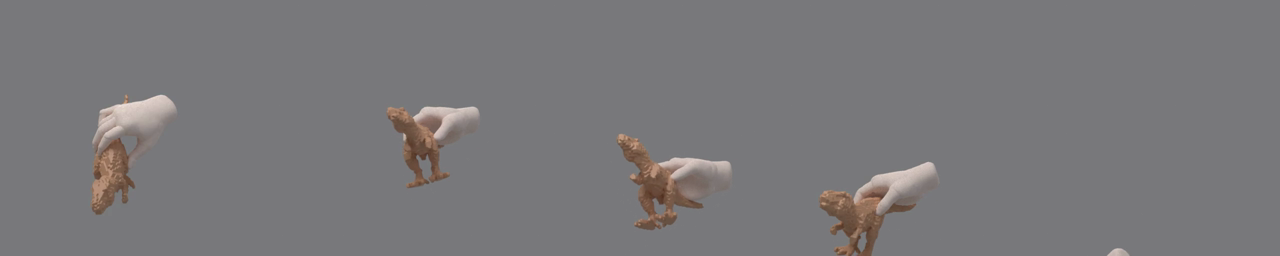}
\end{tabular}}
\caption{\texttt{HOT3D} dinosaur toy. MPC makes the reconstructed human motion physically feasible.}
\label{fig:appendix-hot3d-real2sim-dino}
\end{figure}

\end{document}